\documentclass[twoside,11pt]{article}

\usepackage[preprint]{jmlr2e}

\usepackage{amssymb,amsmath,amsfonts, amscd,array, algcompatible }
\usepackage{bbm, booktabs}
\usepackage{caption, color, chngcntr}
\usepackage{diagbox}
\usepackage{enumerate, enumitem}
\usepackage{floatrow}
\usepackage{graphicx}
\usepackage{hhline, hyphenat}
\usepackage{latexsym, lipsum, longtable }
\usepackage{makecell, mathrsfs, mathtools, multirow}
\usepackage{pifont}
\usepackage{relsize}
\usepackage{scrextend, soul, subcaption, siunitx}
\usepackage{tabularx, textgreek, verbatim}
\usepackage[ruled]{algorithm2e}
\usepackage[dvipsnames]{xcolor}

\definecolor{darkgreen}{rgb}{0.0,0.40,0.13}
\usepackage{tikz}
\usetikzlibrary{positioning}
\usepackage[titletoc]{appendix}
\usepackage[section]{placeins}
\usepackage{bm}

\newcommand{\bbE}{\mathbb{E}}

\newcommand{\bbN}{\mathbb{N}}

\newcommand{\bbP}{\mathbb{P}}

\newcommand{\bbR}{\mathbb{R}}
\newcommand{\bbS}{\mathbb{S}}

\newcommand{\calC}{\mathcal{C}}

\newcommand{\calH}{\mathcal{H}}

\newcommand{\calK}{\mathcal{K}}

\newcommand{\calM}{\mathcal{M}}
\newcommand{\calN}{\mathcal{N}}
\newcommand{\calO}{\mathcal{O}}

\newcommand{\spn}{\operatorname{span}}

\newcommand{\ind}{\mathbbm{1}} 
\newcommand{\id}{\mathbb{I}}

\newcommand{\rbracket}[1]{\left(#1\right)}
\newcommand{\sbracket}[1]{\left[#1\right]}
\newcommand{\cbracket}[1]{\left\{#1\right\}}
\newcommand{\norm}[1]{\left\|#1\right\|}
\newcommand{\normS}[2]{\left\|#1\right\|_{#2}}
\newcommand{\abs}[1]{\left|#1\right|}
\newcommand{\innerprod}[1]{\langle #1 \rangle}

\newcommand{\Var}{\mathrm{Var}}

\renewcommand{\leq}{\leqslant}
\renewcommand{\geq}{\geqslant}
\renewcommand{\subset}{\subseteq}

\graphicspath{{figures/}}

\newcommand{\II}{\mathbbm{1}}
\DeclareMathOperator*{\argmin}{{arg\,min}}

\DeclareMathOperator{\Cov}{{Cov}}

\DeclareMathOperator{\rmdiam}{\mathrm{diam}} 

\DeclareMathOperator{\MSE}{\mathrm{MSE}}

\newcommand{\hxp}{{h'_x}}

\newcommand{\Ilhjk}{{I^{(l,h)}_{j,k}}}
\newcommand{\nlh}{{n_{l,h}}}

\newcommand{\Slh}{{S_{l,h}}}

\newcommand{\Rlh}{{R_{l,h}}}

\newcommand{\vlh}{{v_{l,h}}}

\newcommand{\mulh}{{\mu_{l,h}}}

\newcommand{\Siglh}{\Sigma_{l,h}}

\newcommand{\hhx}{{\widehat{h}_x}}
\newcommand{\hvlh}{{\widehat{v}_{l,h}}}

\newcommand{\hvlhhx}{{\widehat{v}_{l,\widehat{h}_x}}}

\newcommand{\hSiglh}{{\widehat{\Sigma}_{l,h}}}

\newcommand{\hmulh}{{\widehat{\mu}_{l,h}}}

\newcommand{\lamlh}[1]{{\lambda_{#1}(\Sigma_{l,h})}}
\newcommand{\hlamlh}[1]{{\lambda_{#1}(\widehat{\Sigma}_{l,h})}}

\newcommand{\hv}{{\widehat{v}}}
\newcommand{\hR}{\widehat{R}}
\newcommand{\hS}{\widehat{S}}
\newcommand{\hG}{\widehat{G}}
\newcommand{\hmu}{{\hat{\mu}}}

\newcommand{\hdist}{{\widehat{\mathrm{dist}}}}
\newcommand{\dist}{{\mathrm{dist}}}
\newcommand{\hF}{{\widehat{F}}}

\newcommand{\hfj}[1]{{\widehat{f}_{j| #1}}}
\newcommand{\hfjk}[1]{{\widehat{f}_{j,k| #1}}}

\newcommand{\Xsub}{\text{X$\in\psi_2\cap\mathcal{C}^2$}}
\newcommand{\Ysub}{\text{Y$\in\psi_2$}}
\newcommand{\zetasub}{$\bm{\zeta}\in\psi_2$}

\newcommand{\reach}{\ensuremath{\mathrm{reach}}\xspace}

\newcommand{\len}{\ensuremath{\mathrm{len}}\xspace}
\newcommand{\Holder}[2]{\left[ #1\right]_{\mathcal{C}^{#2}}}
\newcommand{\Linfty}[1]{\left| #1 \right|_{L^\infty}}

\ShortHeadings{Conditional Regression for the NSVM}{Wu and Maggioni}
\firstpageno{1}

\newcommand{\ip}[2]{\innerprod{#1,#2}}
\newcommand{\cS}{\mathcal S}
\begin{document}

\title{Conditional Regression for the Nonlinear Single-Variable Model}

\author{\name{Yantao Wu} \email ywu212@jhu.edu \\
 \addr Department of Mathematics\\
 Johns Hopkins University\\
 Baltimore, MD 21218-2683, USA
 \AND
\name{Mauro Maggioni} \email mauromaggionijhu@icloud.com \\
 \addr Department of Mathematics and Department of Applied Mathematics \& Statistics\\
 Johns Hopkins University\\
 Baltimore, MD 21218-2683, USA}

\maketitle
\begin{abstract}
Regressing a function $F$ of a predictor $X\in\mathbb R^d$ from $n$ observations without incurring the statistical and computational curse of dimensionality requires exploitable structure.
We consider compositions $F=f\circ g$ in which the inner map $g$ takes values in a low-dimensional space and the outer function $f$ maps these values to the response. Examples include single- and multi-index models, in which $g$ extracts one or more linear features, as well as certain neural networks. Substantially less is known for nonlinear inner maps.
We study the Nonlinear Single-Variable Model (NSVM), in which $g=\Pi_\gamma$ returns the arclength coordinate of the closest point on an unknown regular curve $\gamma$, and $F(X)=f(\Pi_\gamma X)$ for an unknown one-dimensional link $f$.
The predictor may have full-dimensional variation throughout a tubular neighborhood of the curve.
We group observations into slices according to their response values, use local principal component analysis to estimate a tangent direction in each slice, and fit a one-dimensional piecewise constant regression along that direction.

We assume that at and above a fixed response scale, the length and diameter of the arclength preimage of a response interval are comparable to the length of its intersection with the link range. Under this coarse monotonicity condition, an upper bound on the arclength density, and sufficient variation normal to the curve, the estimation term attains the one-dimensional minimax mean squared rate up to logarithmic factors. The remaining terms measure curve approximation and the probability of regions of low arclength density.

In thin slices the variance along the tangent is small relative to the normal variances. For wide slices, where the tangential variance dominates, we prove a complementary bound using the largest principal component.
For a fixed noisy thin-slice model, the estimator can be constructed in time $\mathcal{O}(d^2n\log n)$. With the model parameters held fixed, the constants and sample-size thresholds depend at most polynomially on the ambient dimension $d$.
\end{abstract}


\section{Introduction}
We consider the standard regression problem of estimating a function $F:\bbR^d\to\bbR$ from $n$ samples $\{(X_i,Y_i)\}_{i=1}^n$, where $X_1,\dots,X_n$ are i.i.d. realizations of a predictor variable $X\in\bbR^d$ with distribution $\rho_X$, and $\zeta_1,\dots,\zeta_n$ are mutually independent realizations of a centered observational-noise variable $\zeta$, independent of $X_1,\dots,X_n$. With $Y$ denoting the response paired with $X$, our observations satisfy
\[ Y_i = F(X_i) + \zeta_i\,. \]
For a bounded function $u$ on a set $D\subseteq\bbR^m$ and $0<s\leq1$, write $\Linfty{u}=\sup_{x\in D}|u(x)|$ and define its H\"older seminorm by
\[
 \Holder{u}{s}:=\sup_{\substack{x,y\in D\\x\neq y}}\frac{|u(x)-u(y)|}{\norm{x-y}^s}.
\]
We write $u\in\calC^s(D)$ when this seminorm is finite. For nonnegative quantities $a,b$ we write $a\lesssim b$ (equivalently $b\gtrsim a$) if $a\leq Cb$ for a constant $C>0$, and $a\asymp b$ if both inequalities hold.
For the distribution-free benchmark, let $\rho_X$ range over probability measures on $[0,1]^d$ and let $F$ range over the H\"older ball $\Linfty{F}+\Holder{F}{s}\leq L$ on this cube, with $L>0$ fixed and $0<s\leq1$. The squared $L^2(\rho_X)$ prediction error integrates the squared difference between estimator and regression function against the predictor law. For independent centered noise with variance bounded by a fixed positive constant, the minimax expected squared $L^2(\rho_X)$ error over these functions, design measures and noise laws is of order $n^{-2s/(2s+d)}$, and its square root is of order $n^{-s/(2s+d)}$ \citep{Stone82,GKKW02}.
For example, averaging responses on a regular grid of width $h$, with estimate zero in empty cells, gives a squared risk bounded by a constant times $h^{2s}+1/(nh^d)$ uniformly in $\rho_X$. Choosing $h\asymp n^{-1/(2s+d)}$ gives the upper bound. For kernel and multiscale estimators see \citet{GKKW02,DeVore:UniversalAlgorithmsLearningTheoryI,DeVore:UniversalAlgorithmsLearningTheoryII}.
This rate deteriorates dramatically as the dimension $d$ of the ambient space increases: this is an instance of the {\em curse of dimensionality}.
Because no estimator can achieve a faster minimax rate over all such functions and $d$ is very large in many applications, it is natural to consider structured regression problems in which the curse of dimensionality can be avoided.

In this work, we study the Nonlinear Single-Variable Model (NSVM) $F(x)=f(\Pi_\gamma x)$, where $f$ is a one-dimensional link, $\gamma$ is an unknown arclength-parametrized curve in $\mathbb{R}^d$ and $\Pi_\gamma$ returns the parameter of the closest point on $\gamma$. The predictor $X$ may have full-dimensional variation in a neighborhood of the curve and is therefore not intrinsically low-dimensional. Grouping observations with responses in a common interval gives a response slice. Our estimator learns its local curve geometry by principal component analysis (PCA), assigns each point to a nearby slice, and performs one-dimensional local polynomial regression along the resulting coordinate. For the tuned estimator the estimation term attains the one-dimensional minimax rate up to logarithmic factors, with additional terms for curve approximation and low arclength density.

We call $f$ coarsely monotone when, at and above a fixed response scale, the length and diameter of the inverse image of a response interval are comparable to the length of its intersection with the link range. We assume an upper bound on the density of the arclength coordinate and control the risk separately on neighborhoods where that density is bounded below. Under a lower bound on normal variation, the tangent is approximated by the smallest principal component of a thin slice, where the tangential variance is small relative to the normal variances. We also give a guarantee for wide slices, where the tangential variance dominates, under a small-curvature condition, where we use the largest principal component.

We first state the model, its main guarantees, and our contributions, and then place them in the context of related work. We view the present model as primarily of theoretical interest because its inner function is currently restricted to projection onto a one-dimensional manifold. Extending the construction to higher-dimensional manifolds is natural and may help explain why some functions can be learned efficiently even from data that appear far from any low-dimensional structure. We also give a stylized application to learning committor functions, which give the probability of reaching one prescribed set before another in a stochastic system, where a reaction-path coordinate can induce an approximate NSVM structure; see Section \ref{s:ex_reaction_coordinates}.

\subsection{The Nonlinear Single-Variable Model and contributions}
Our contributions are as follows:
\begin{enumerate}[label=(\roman*)]
\item We study the {\em{Nonlinear Single-Variable Model}}, an intermediate model between the semiparametric single-index model and the nonparametric function-composition model: both the outer and inner functions are nonlinear, while the inner function is the closest-point coordinate associated with an unknown curve $\gamma\subset\mathbb{R}^d$, and the distribution of $X$ has a corresponding tubular geometric structure;
\item We give finite-sample bounds for the tuned estimator in thin and wide slices. In the thin-slice regime, the error is bounded by the sum of a one-dimensional estimation term, a geometric approximation term, and the contribution of regions of low arclength density (Theorem~\ref{Thm: NLSIM}). Theorem~\ref{Thm: NLSIM_wide} gives the corresponding saturation bound for wide slices.
\item We give an algorithm that constructs the estimator in near-linear time for a fixed noisy thin-slice model. We give the dependence of the risk, sample-size and computational bounds on $d$ and on the model parameters.
\end{enumerate}

This model generalizes the single-index model by allowing the inner map $g=\Pi_\gamma$ to be nonlinear, while remaining amenable to estimation with strong statistical guarantees whose rate exponents and smoothness requirements are independent of the ambient dimension.

\begin{definition}[Nonlinear Single-Variable Model]\label{def:NSVM}
The regression function $F$ admits the decomposition
$$\bbE[Y|X] = F(X) = f(\Pi_\gamma X) $$
where the \emph{underlying curve} $\gamma: [0, \len_\gamma ]\to\bbR^d$ is twice continuously differentiable, is parameterized by arclength without loss of generality, and has length $\len_\gamma$
, and the \emph{link function} $f\,:\,[0, \len_\gamma]\to\bbR^1$ depends only on one variable.
The random vector $X$ is supported in some domain $\Omega_\gamma\subseteq\bbR^d$ containing the image of $\gamma$, such that the closest-point projection $\Pi_\gamma:\Omega_\gamma\to [0, \len_\gamma]$, with
$$\Pi_\gamma(x):=\argmin_{t\in[0, \len_\gamma]}\norm{x-\gamma(t)}\,,$$
is well defined on $\Omega_\gamma$, i.e., the minimizer is unique.
\end{definition}
Let $t=\Pi_\gamma X$ be the random arclength coordinate. For each unit vector $v$ on the sphere $\bbS^{d-1}$, choose an orthogonal matrix $M_v$ that maps the last coordinate vector to $v$. We assume that the displacement from the closest point is normal to the curve also at its endpoints. With $Z_{d-1}\in\bbR^{d-1}$ denoting the centered displacement in normal coordinates, this gives
\begin{equation}
 X = \gamma(t) + M_{\gamma'(t)} \begin{pmatrix} Z_{d-1} \\ 0 \end{pmatrix}
 \label{e:Xsplit}
 \end{equation}
Conditional on $t=t_0$, the displacement $X-\gamma(t_0)$ belongs to $\spn\{\gamma'(t_0)\}^\perp$. When the conditional covariance of $Z_{d-1}$ is a multiple of the identity, write $\sigma_\gamma(t_0)^2$ for its variance in each normal direction. Let $\sigma_\gamma$ and $\bar\sigma_\gamma$ be the essential infimum and supremum of $\sigma_\gamma(t)$ under the arclength law. We denote its density, when it exists, by $\rho_t$.

\noindent{\bf{The regression problem for the Nonlinear Single-Variable Model}}: Given pairs $\{(X_i,Y_i)\}_{i=1}^n$, where $X_1,\dots,X_n$ are independent copies of $X$ as above and $Y_i=F(X_i)+\zeta_i$, with $F$ as in Definition \ref{def:NSVM} and observational noise $\zeta_i$, the goal is to estimate the regression function $F$ on the support of $X$. The variables $\zeta_1,\dots,\zeta_n$ are assumed to be sub-Gaussian, mutually independent, and independent of $X_1,\dots,X_n$. Write $\sigma_\zeta^2$ for the sub-Gaussian variance proxy of the noise, so $\bbE e^{u\zeta}\leq e^{u^2\sigma_\zeta^2/2}$ for all $u\in\bbR$.

\begin{figure}[t]\centering
\includegraphics[height=0.4\textwidth]{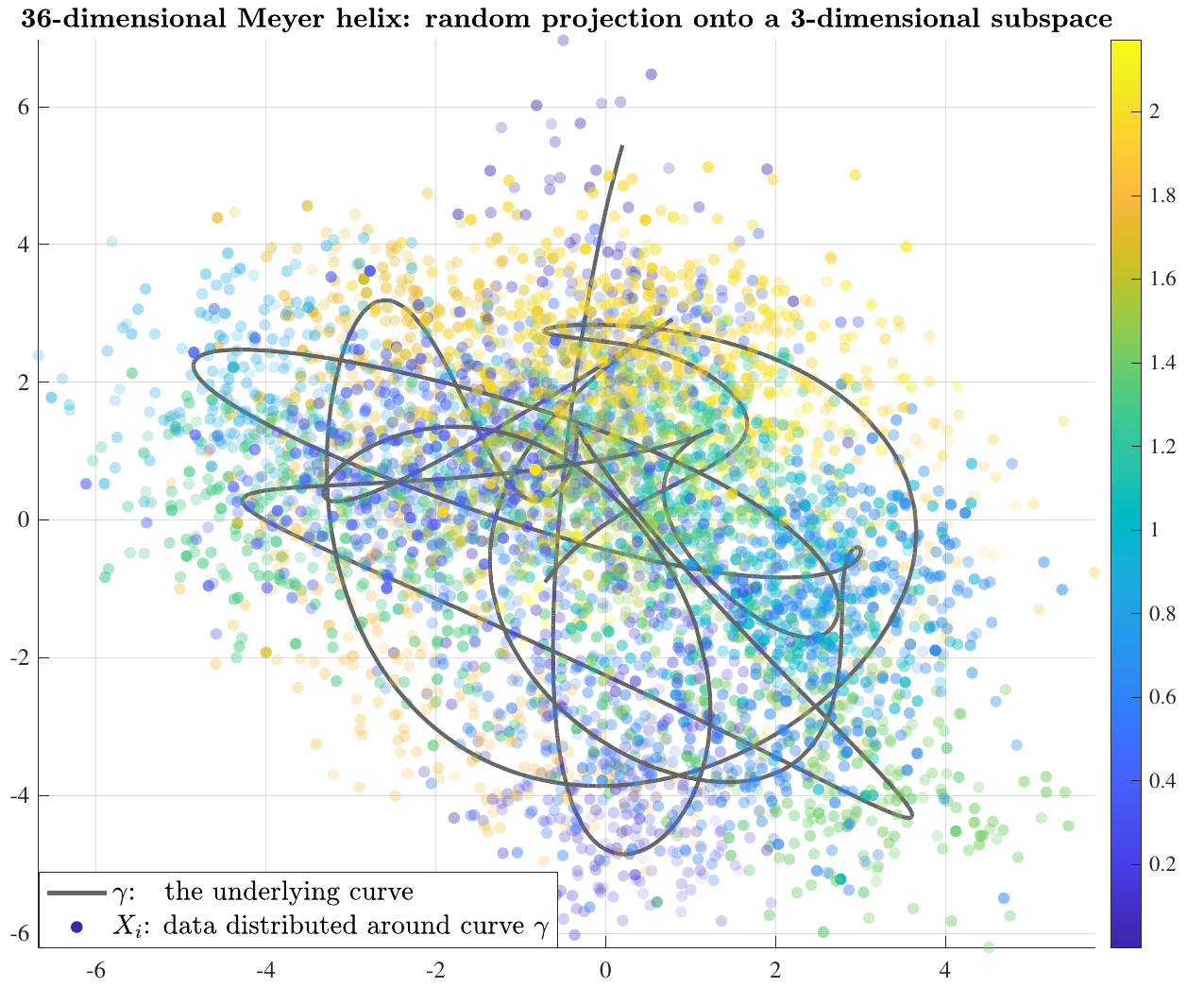}
\includegraphics[height=0.4\textwidth]{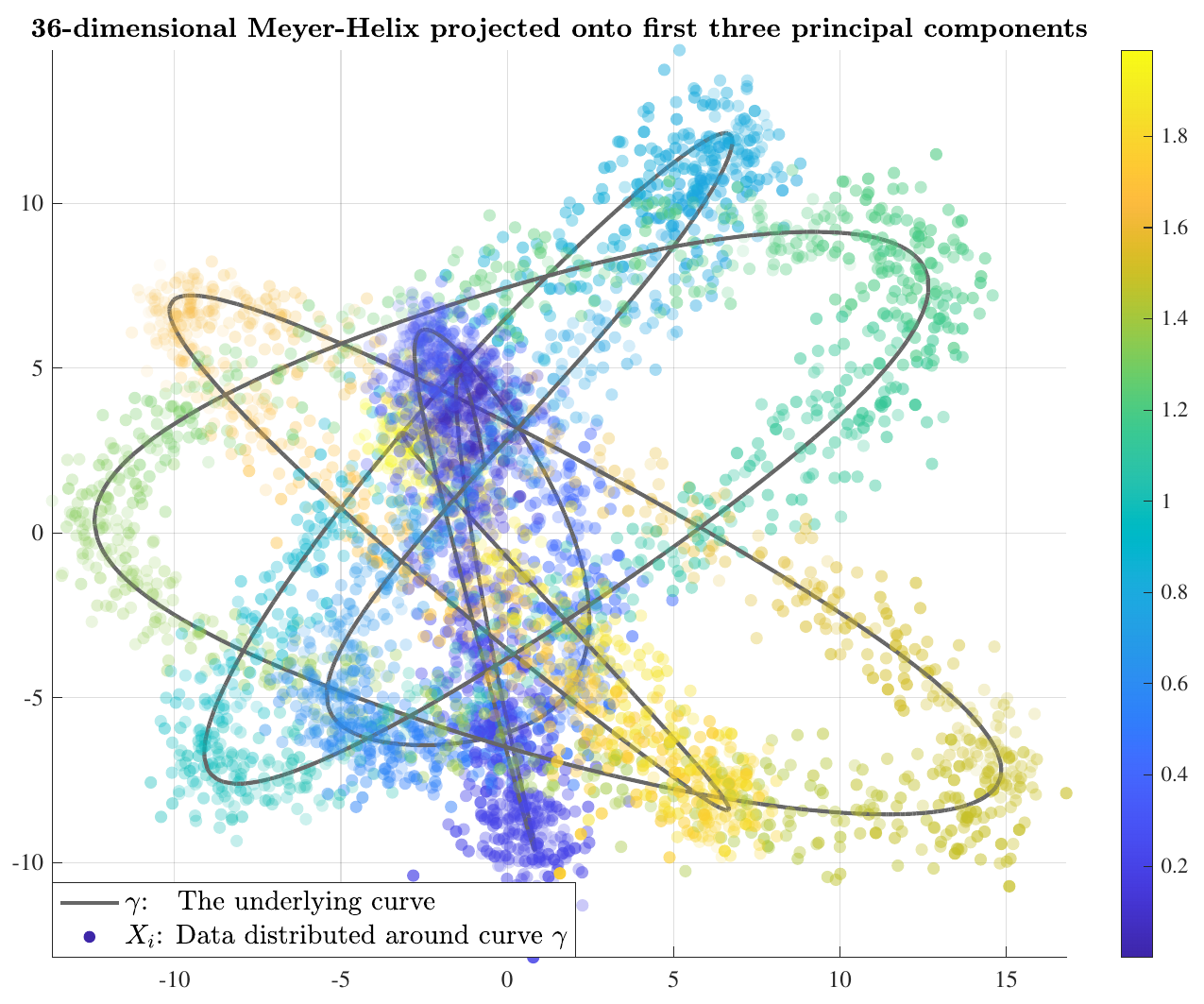}
\caption{An example of the Nonlinear Single-Variable Model in Definition \ref{def:NSVM}. The curve $\gamma$, shown in black, is a Meyer helix in $\bbR^{36}$ (see Appendix \ref{Appendix: Meyer-Helix}) with normal-noise scale $0.5$; the link $f\in\calC^{0.7}(\bbR)$ is strictly monotone, and $\zeta\equiv0$. We generate $n=5000$ samples $X_i$ in a tube of radius 6 around the curve and color them by $Y_i=F(X_i)=f(\Pi_\gamma X_i)$. Left: a random projection onto $\mathbb{R}^3$. Right: the orthogonal projection onto the first three principal components. The distribution does not appear to admit a low-dimensional linear embedding without increased geometric complexity; see Appendix \ref{Appendix: Meyer-Helix}.}
\label{Figure: 36 dimensional Meyer-Helix-L1-C1-variant1}
\end{figure}

The distribution of $X$, the link function $f$, and the underlying curve $\gamma$ are all unknown. Figure~\ref{Figure: 36 dimensional Meyer-Helix-L1-C1-variant1} illustrates the model.

The domain $\Omega_\gamma\subseteq\bbR^d$ must be chosen so that the closest-point projection $\Pi_\gamma$ is well defined; this condition is related to the reach of $\gamma$ \citep{federer1959curvature}.
Define the distance function $\dist_\gamma(x) : \bbR^d \to [0,\infty)$, $ \dist_\gamma(x) = \inf \{ \norm{ x -\gamma(t_0)} :\ t_0\in [0,\len_\gamma] \} $,
the domain $Unp_\gamma$ is defined as the set of all points $x\in\bbR^d$ for which there is a unique point $\gamma(t_0)$ closest to $x$.
The map $\Pi_\gamma : Unp_\gamma \to [0,\len_\gamma]$ maps $x\in Unp_\gamma$ to the unique $t \in[0,\len_\gamma]$ such that $\dist_\gamma(x) = \norm{ x - \gamma(t) }$.
Let $B(x,r)$ denote the open Euclidean ball of radius $r$ about $x$. At arclength $t_0\in[0,\len_\gamma]$, the local reach is
$ \reach_\gamma(t_0) := \sup \{ r : \ B(\gamma(t_0), r) \subseteq Unp_\gamma \}$,
and the global reach is
$ \reach_\gamma := \inf \{ \reach_\gamma(t_0) :\ t_0\in[0,\len_\gamma] \}$.
Both $\reach_\gamma(t_0)$ and $\reach_\gamma$ take values in $[0,\infty]$. With these definitions, $\Omega_\gamma$ could be as large as $\bigcup_{t_0\in[0,\len_\gamma]} B(\gamma(t_0), \reach_\gamma(t_0))$.
Thus, $\rho_X$ need not be supported on, or highly concentrated near, $\gamma$. As in single-index models, the data may have substantial variance in directions normal to the index; here we allow this variance to be as large as possible while keeping $\Pi_\gamma$, the nonlinear analogue of orthogonal projection onto the index line, well defined.

For the formal guarantees we take $s\in[\frac12,1]$ and write $\Linfty{f}$ for the supremum of $|f|$ on $[0,\len_\gamma]$. To specify coarse monotonicity, let $\mathcal R_f:=f([0,\len_\gamma])$ be the link range. Write $|E|$ for the Lebesgue measure of a subset $E$ of the line and $\rmdiam(E)$ for its diameter, with $\rmdiam(\emptyset)=0$. We call $f$ coarsely monotone at scale $\omega_f\geq0$ if there are constants $C_f>C_f'>0$, depending only on $f$ and with ratio bounded by an absolute constant, such that every interval $T$ with $|T\cap\mathcal R_f|\geq\omega_f$ satisfies
\begin{equation*}\tag{CM}\label{eq:coarse-link}
 C_f'|T\cap\mathcal R_f|\leq|f^{-1}(T)|\leq \rmdiam(f^{-1}(T))\leq C_f|T\cap\mathcal R_f|.
\end{equation*}
In each prediction risk, $X$ is a test predictor drawn independently of the training sample. We denote the mean squared prediction error by MSE. In the informal bound below, $C_1$ is the coefficient of the sampling term and $C_2$ bounds the contributions of curve approximation and low arclength density at the chosen resolution. Theorem~\ref{Thm: NLSIM} gives their formulas.
 \begin{theorem}[Informal]\label{Thm: Simplified Version}
Suppose that $\sigma_\zeta>0$, that the arclength density is bounded above, that $f\in\calC^s(\bbR^1)$ for some $s\in[\frac12,1]$, that $f$ is coarsely monotone, and that for a sufficiently large constant $C_{\mathrm{LCV}}>0$ the lower conditional variance condition $\sigma_\gamma \geq C_{\mathrm{LCV}}C_f\max(\sigma_\zeta,\omega_f)$ holds.
Under the remaining model and scale conditions of Theorem~\ref{Thm: NLSIM}, for sufficiently large $n$, the regression estimator $\hF$ constructed by Algorithm \ref{Alg: NVM} satisfies
\begingroup
\[
\bbE\sbracket{\abs{\hF(X)- F(X)}^2} \lesssim C_1(f,\gamma,\rho_X,\sigma_\zeta,d) n^{-\frac{2s}{2s+1}}\log n + C_2(f,\gamma,\rho_X,\sigma_\zeta,d)\,.
\]
\endgroup
With the other model parameters held fixed across dimensions, $C_1$ and $C_2$ grow at most polynomially in $d$.
The estimator $\hF$ can be constructed by Algorithm~\ref{Alg: NVM}, which for a fixed noisy model runs in time $\mathcal{O}(d^2 n\log n)$.
\end{theorem}
The bound on the expected squared $L^2(\rho_X)$ error contains two terms. The first has the one-dimensional minimax exponent for $s$-H\"older regression, up to a logarithmic factor. The second accounts for the geometric error at the chosen slice width and for the probability of neighborhoods with low arclength density. Under a uniform positive lower bound on the arclength density, we may choose the density threshold below that lower bound so that the second term contains only the geometric contribution.
The slice centers and principal directions describe the curve locally, and the fitted regressions estimate $F$ in these local coordinates. Reversing the parametrization replaces $\Pi_\gamma$ by $\len_\gamma-\Pi_\gamma$ and $f(t)$ by $f(\len_\gamma-t)$, with the same observations. Thus the data do not determine a global orientation for the coordinate or link.
Further remarks may be found after the formal statements of the main results in Section \ref{s:estimator}.

Table \ref{tab:notation} collects the notation specific to the NSVM and its estimator.
\begin{table}[h]
\caption{\normalfont \textbf{Notation}}
\label{tab:notation}
\centering
\renewcommand{\arraystretch}{1.18}
\resizebox{\columnwidth}{!}{
\begingroup
\begin{tabular}{| l l | l l |}
\hline
\textbf{symbol} & \textbf{definition} & \textbf{symbol} & \textbf{definition} \\
\hline
$\Pi_\gamma$ & closest-point coordinate on $\gamma$ & $\len_\gamma,\reach_\gamma$ & length and reach of $\gamma$ \\
$F=f\circ\Pi_\gamma$ & regression function & $f$ & one-dimensional link function \\
$l,j$ & slice count and bins per slice & $\Rlh$ & $h$th of the $l$ response intervals \\
$\Slh,\hS_{l,h}$ & slice law and sample & $I^{(l,h)},A_{l,h}$ & projected interval and accepted region \\
$\mulh,\hmulh$ & population and empirical slice centers & $\Siglh,\hSiglh$ & population and empirical slice covariances \\
$G_{l,h},\widehat G_{l,h}$ & population and empirical slice-shape scores & $\vlh,\hvlh$ & selected principal directions \\
$\calH_l$ & indices of retained slices & $n_{loc}$ & minimum count for a retained slice \\
$\dist(x,h),\hdist(x,h)$ & population and empirical slice distances & $\hxp,\hhx$ & response and assigned slice indices \\
\hline
\end{tabular}
\endgroup}
\end{table}

\subsection{Related work}
\noindent{\bf{Intrinsically low-dimensional models}}. One approach is to impose geometric assumptions on the distribution $\rho_X$ of the predictor $X$, for example, that $\rho_X$ is supported on a low-dimensional manifold $\mathcal{M}$ of dimension $r\ll d$, while $F$ is a generic $s$-H\"older function on $\mathcal{M}$.
In these settings, some estimators attain the optimal rate with respect to the intrinsic dimension $r$ and admit fast algorithms; see \citep{BL07, Kpotufe11, KG13, LMV16, LMV22ME,liu2024deepneuralnetworksadaptive}. For example, the estimators in \citep{LMV16,LMV22ME} can be constructed in time $\mathcal{O}(C^r d^s n\log n)$ and adaptively achieve a minimax-optimal rate, up to logarithmic factors, over a large family of function spaces with unknown regularity that may vary across locations and scales.
The model considered here extends some of the works above by allowing the data not to lie exactly on a low-dimensional manifold, for example because of factors irrelevant to predicting the values of $F$ or because of noise.

\noindent{\bf{Highly regular function classes}}. In a different direction, one may consider function classes with smaller complexity than $s$-H\"older functions on $\mathbb{R}^d$.
Here smoothness of order $s>1$ refers to regularity of derivatives. One straightforward but highly restrictive assumption is that the smoothness order $s$ scales proportionally with $d$, making the minimax rate above independent of $d$.
Larger function classes can be obtained by imposing either strong mixed smoothness \citep{stromberg}, such as membership in highly anisotropic Besov spaces (with $\mathcal{O}(1)$ axis-aligned directions of $\mathcal{O}(1)$ smoothness and $\mathcal{O}(d)$ smoothness in all remaining directions), or integrability of the Fourier transform \citep{Barron93}.
In the context of deep neural networks, \cite{suzuki2021deep} demonstrate that functions in this class can be approximated well, with suitable architectures, in a way that avoids the curse of dimensionality if the degree of anisotropic smoothness is sufficiently large (``$\mathcal{O}(d)$''), while leaving open the aspects of learning and optimization.
The condition of belonging to a Barron's space requires sufficiently fast decay of the Fourier transform of $\nabla F$, which is a condition that gets stronger with the dimension, except for very special cases, among which, notably the single- and multi-index models that we discuss momentarily, where the Fourier transform of $\nabla F$ (and $F$ itself) is a singular measure.
This requirement excludes intrinsically low-dimensional nonlinear models from the Barron's space since the Fourier transform of a measure supported on (or near) a general nonlinear curve has slow decay \citep{ACK87, BGGIST07}. Our main model here is therefore typically not included in the Barron's space, though both the Barron space and our main model include the single-index model as a special case.

\noindent{\bf{Compositional models}}. Other function classes arise from structural assumptions, often expressed through compositional models.
For example, suppose that
$$
F=f\circ g\quad, \text{ with } g:\bbR^d\to\bbR^r \text{ and } f:\bbR^r\to\bbR^1
$$
where $f$ and $g$ have suitable H\"older regularity and $r\leq d$.
For general compositions, \citet{JLT09} prove that sufficiently smooth inner functions can improve the minimax nonparametric rate for estimating $F=f\circ g$, although the resulting rate may still suffer from the curse of dimensionality.
Recent works combine anisotropic smoothness with composability, especially in the context of approximators or estimators constructed with deep neural networks. In this direction, \citep{10.1214/19-AOS1875,shen2021deepquantileregressionmitigating} consider spaces of functions that are compositions of low-dimensional functions with anisotropic smoothness conditions, studying the dependence of approximability and learning risk on the dimensionality of the spaces involved and the smoothness of the corresponding functions. Unfortunately the innermost such function is defined on the original high-dimensional space $\mathbb{R}^d$, so in general these results do not circumvent the curse of dimensionality, unless again the (anisotropic, in general) regularity of that function scales with $d$. Another direction is to design different types of neural networks that exploit nonlinear compositions, see \citep{Lai2021TheKS, liu2024kankolmogorovarnoldnetworks}.
Most of these results address only whether a function can be approximated well by a neural network; they neither analyze the learning problem nor address computational questions, such as whether an optimization algorithm can efficiently find the parameters of the desired estimator.
Another aspect that is often left unaddressed is the dependence of constants on the ambient dimension, which is unfortunately often exponential: see \cite{Shamir_discussion} for a discussion of some of these aspects, and \cite{shen2021deepquantileregressionmitigating} for bounds in particular cases.
In the parametric setting, models in which $g$ is a polynomial and $f$ is a function in a known finite-dimensional space have been considered; for example \cite{wang2024learning,lee2024neuralnetworklearnslowdimensional} (and related work referenced therein) consider special classes of polynomials $g$ for which a customized layer-by-layer training of a neural network leads to estimators of $F$ that are provably not cursed by the dimensionality.

\noindent{\bf{Single- and multi-index models}}. Classical examples of structural assumptions based on function composition for which the curse of dimensionality can provably be avoided include single- and multi-index models, as well as generalized linear models \citep{Stone82,Hastie:GeneralizedAdditiveModels,doi:10.1080/01621459.1985.10478157,Horowitz_generalclassnonparametric}.
When $g=G:\bbR^d\to\bbR^r$ is linear, the resulting model is called the multi-index model: $F\in\calC^s(\bbR^d)$ depends on only $r$ linear features, $F(x)=f(G x)$ for some {\em link} function $f\in\calC^s(\bbR^r)$, matrix $G:\mathbb{R}^d\rightarrow\mathbb{R}^r$, and the projection of $X$ onto the row space of $G$, that is, the range of $G^\intercal$, is sufficient for regression.
The special case $r=1$ is the single-index model, in which $F(x)=f(\langle v,x\rangle)$ for an unknown index vector $v\in\bbS^{d-1}$ and an unknown link function $f\in\calC^s(\bbR)$.
\citet[Question~2]{Stone82} asked whether the one-dimensional rate $n^{-s/(2s+1)}$ is attainable in the single-index model. For a known rank-$r$ index subspace, the corresponding $r$-dimensional benchmark is $n^{-s/(2s+r)}$.
For a twice differentiable link, \citet[Theorem~3.3]{HS89} obtain a pointwise kernel estimation error of order $n^{-2/5}$ in probability under their regularity conditions. Their average-derivative estimator also yields a root-$n$ estimate of the index direction when the average derivative is nonzero. See also \citet[Section~2.5]{Horowitz98}.
\citet[Theorems~1 and~2]{GL07} prove matching upper and lower bounds of order $n^{-2s/(2s+1)}$ for expected squared $L^2(\rho_X)$ error under their conditions on the projected design and noise. For related dimension-reduction results see \citet[Chapter~22]{GKKW02}.
Unfortunately, these estimators often lack implementations that provably avoid the curse of dimensionality; for example, their constants may be exponential in the dimension. \cite{liu2024learningcentralsubspace} is a recent approach, based on contour regression analysis, for multi-index models; however, the dependence of its constants on $d$ is not made explicit, and the available analysis does not establish polynomial dependence on $d$.
These models have also attracted recent attention through their connection to ``feature learning'' in neural networks. See, for example, \citet{lee2024neuralnetworklearnslowdimensional,alberti2021gradient} and the references therein on learning single-index models with suitably trained neural networks or gradient descent, respectively. Related work suggests that deep neural networks exploit low-dimensional linear subspaces for classification and prediction \citep{doi:10.1126/science.adi5639}.

Functions in our nonlinear single-variable model can be approximated efficiently by wavelets, local Fourier expansions, or neural networks adapted to the geometry of $\gamma$, for example through local tensor approximations whose first component is tangent to the curve. What remains unclear is how to estimate and compute such representations from data without incurring the curse of dimensionality. Related work includes kernel methods composed with a linear projection \citep{PoggioRosascoHyperkernels}, which provide some algorithmic guarantees, and neural-network approximations \citep{liao2024taskmanifoldneuralnet}, for which no corresponding algorithmic guarantee is given.

Many methods have been developed for jointly estimating the index vector $v$ (or matrix $G$) and the link $f$, with different tradeoffs between statistical assumptions and computational cost; see \citet{LMV22B} for an extended discussion.
One category includes semiparametric methods based on maximum likelihood estimation \citep{Ichimura93, HHI93, DHH97, DH99, DHP06, CFGW97} and M-estimators that produce $\sqrt{n}$-consistent index estimates under general assumptions, but with computationally demanding implementations, relying on sensitive bandwidth selections for kernel smoothing and on high-dimensional joint optimization of $f$ and $v$.
Another category includes direct methods. For example, Average Derivative Estimation \citep{Stoker86, HS89} estimates the index vector $v$ by exploiting its proportionality to $\nabla F$.
Early implementations of this idea suffer from the curse of dimensionality due to the use of kernel estimation for the gradient.
\citet{Xia_TwoEstimatorsSIM} uses local kernel estimators for estimating the direction of the gradient, and the idea could in principle be applied to our model. However, our theoretical analysis would not apply, and it is unclear whether the resulting method would avoid the curse of dimensionality.
The method of \citet{HJS01} uses an iterative scheme to adapt an elongated neighborhood window gradually, but it requires a sufficiently accurate initialization.
Gradient-based methods (stochastic and non-stochastic) have been studied in the context of neural networks \citep{lee2024neuralnetworklearnslowdimensional,bietti2023learninggaussianmultiindexmodels,bietti2022learning,pmlr-v247-damian24a}.

A category of techniques particularly relevant to the present work includes {\em{conditional (or inverse-regression) methods}}, which derive their estimators from statistics of the conditional distribution of the explanatory variable $X$ given observations of the (noisy) response variable $Y$.
Prominent examples include sliced inverse regression \citep{DL91, Li91}, sliced average variance estimation \citep{Cook00}, simple contour regression \citep{LZC05,CLS13}, and its multi-index analysis in \citet{LW07}, which yields a near-optimal estimator for the multi-index model.
Conditional methods partition the range of the response variable $Y$ into small intervals and consider their preimages, which, for a monotone single-index link, can be thin along the index direction $v$ if the transverse variance is large enough relative to the within-slice index variance.
They are often straightforward to implement, requiring the computation of conditional empirical moments or other statistics related to the level sets of $f$, and may have only one tuning parameter: the slice width.
Unfortunately, several of the works above, including \citep{HJS01,LW07}, while obtaining asymptotic learning rates that avoid the curse of dimensionality and are in some cases minimax optimal or near-optimal, require a minimum sample size that is exponential in the ambient dimension $d$ (either explicitly, or through constants in the bounds).

Beyond the statistical fact that single- and multi-index models need not incur a cost cursed by the ambient dimension, another critical problem is to design algorithms with reasonable computational cost for implementing statistically optimal estimators.
For example, even in the case of single-index models, $M$-estimators typically require high-dimensional nonconvex optimization in $v$ and $f$; methods based on optimizing over $v$, even when combined with random sampling, may scale exponentially in $d$ because of the need to obtain points inside a narrow cone around the unknown $v$.
Conditional methods can often be implemented efficiently, with running time $\calO(C_d n)$ up to logarithmic factors (except for contour-regression methods, which scale quadratically in $n$), where $C_d$ is a low-order polynomial in $d$.
\citet{LMV22B} discuss these techniques in detail for the single-index model and introduce Smallest Vector Regression, which attains optimal statistical guarantees up to logarithmic factors, provides a theoretically optimal rule for selecting the slice width (a crucial parameter whose choice is often not discussed; see \citealp[p. 75]{CLS13}), and admits an implementation with cost $\mathcal{O}(d^2n\log n)$. Neither the rate exponents nor the constants in the statistical, sample-size, and computational bounds incur the curse of dimensionality.

At the level of statistical models, the manuscript of \citet{manuscriptNSIM2018} introduced a version of the nonlinear single-index model and attempted a first study of its properties using the ideas in \citet{LMV22B}, which had put the focus on the development of inverse regression techniques under assumptions on the geometry of sufficient statistics for regression, and on using low-order moments to estimate the geometry of response slices.
The contribution here lies in the slice-geometry estimator, its geometric adaptivity, and the accompanying finite-sample and computational guarantees rather than in the curve-projection model alone.
The ideas in \citet{manuscriptNSIM2018} were further developed, by a subset of the authors \citep{KeretaKlockNaumova2021}, to estimate local index vectors by linear regression within response slices and predict with a $k$-nearest-neighbor estimator based on a localized proxy for the geodesic metric on the curve.
The main results in \citet{KeretaKlockNaumova2021} for bi-Lipschitz link functions, under the stated assumptions, are not proven therein, due to gaps in the proof\footnote{The positivity of their normalized covariance $\sigma_{j,Y}$, computed in response slice $j$ between the response and the predictor projected on the local tangent, used pervasively in all the main results (Corollary 3.1; the main Theorem 4.3, Remark 4.1 (the headline rates (4.4) and the noise-free (4.5)), and Proposition 4.5), requires an upper bound on the Hessian, and therefore, in particular, $f\in\mathcal{C}^2$; once that assumption is added, one further needs to assume a sufficiently small upper bound on the noise level, the supremum norm of the link Hessian and the Lipschitz constant of the link to guarantee that a suitable scale for which $\sigma_{j,Y}$ is positive in fact exists. Other errors in the proofs appear either repairable or of lesser consequence.}. The proofs as given only work by restricting the results to the case of monotone $\mathcal{C}^2$ functions with sufficiently small Lipschitz constant and Hessian, as well as small observation noise. Under these stronger assumptions the results do not yield optimal learning rates.
On the computational side, the construction therein proposed is straightforward to implement when the curve is nearly linear, but in the general case, it requires prior knowledge of a critical scale parameter lying in a narrow interval determined by the reach, which is itself difficult to estimate. The numerical experiments reported in \citet{KeretaKlockNaumova2021} do not establish that all assumptions required by the theory hold in those settings.

Our estimator uses response-slice PCA to estimate local tangent directions. In the thin-slice regime a common weight depending on the reach and the arclength scale enters the empirical distance. The wide-slice guarantee uses the largest principal component on each slice. In both cases we regress the response on the estimated one-dimensional coordinate.

In conclusion, while the situation is rather well-understood for compositional models $F=f\circ g$ with $g$ linear, much remains open when $g$ is nonlinear: one needs to identify which geometric and statistical assumptions allow the compositional structure to circumvent the curse of dimensionality, beyond settings that require very high regularity of $g$, regularity increasing with $d$, or membership in special polynomial families. In the curve-projection setting treated here, the remaining challenge is not only to specify a nonlinear index, but to learn from data the relevant local geometry, the nonlinear coordinate, and the one-dimensional link with explicit risk and computational guarantees.

\section{An estimator for the Nonlinear Single-Variable Model}
\label{s:estimator}

We propose an inverse-regression estimator of $F$ for the Nonlinear Single-Variable Model, together with an efficient algorithm that also produces a sketch of the curve $\gamma$ and local coordinates along the curve.

\paragraph{Model assumptions.} Each result below specifies which of the following conditions it uses. We write $\psi_2$ for the sub-Gaussian class. A vector has variance proxy $v^2$ if every centered unit projection has moment generating function bounded by $e^{u^2v^2/2}$ at each $u\in\bbR$. When the predictor law has a density we also denote that density by $\rho_X$.

\begin{enumerate}[label = \textcolor{black}{\rm\textbf{(\Xsub)}}]
\item \label{Xsub} $X$ is sub-Gaussian with variance proxy $R_0^2$ and has a Lebesgue density supported in a closed normal tube $\mathcal T_b:=\{\gamma(t)+u:0\leq t\leq\len_\gamma,\ u\perp\gamma'(t),\ \norm{u}\leq b\}$ for some finite $0<b<\reach_\gamma$. On $\mathcal T_b$ the density $\rho_X$ is $C^2$ up to the boundary from the interior, meaning that its restriction admits a $C^2$ extension to a neighborhood of $\mathcal T_b$. The density is zero outside $\mathcal T_b$ and may have a positive interior trace on either end cap. For some $C_X\geq1$, set $B_X=C_X\sqrt d\,R_0$. The support lies in $B(0,B_X)$.
\end{enumerate}

\begin{enumerate}[label = \textcolor{black}{\rm\textbf{(\Ysub)}}]
\item \label{Ysub} For some $C_Y>0$, the response $Y$ is sub-Gaussian with variance proxy $C_Y^2R_0^2$.
\end{enumerate}

\begin{enumerate}[label = \textcolor{black}{\rm\textbf{(\zetasub)}}]
\item \label{zetasub} $\zeta$ is centered and sub-Gaussian with variance proxy $\sigma_\zeta^2$.
\end{enumerate}

In normal coordinates let $\id_k$ denote the $k\times k$ identity matrix.

\begin{enumerate}[label = \textcolor{black}{\rm\textbf{(}$\bm{\gamma_1}$\textbf{)}}]
\item \label{gamma1} $\gamma$ has a Lipschitz derivative $\gamma' : [0,\len_\gamma] \to\bbR^d$.
For each $t_0\in [0,\len_\gamma]$, the conditional random vector $Z_{d-1} | t=t_0$ is mean zero, isotropic with variance $\sigma_\gamma(t_0)^2 \id_{d-1}$, and supported in a Euclidean ball $B(0, c\,\reach_\gamma)\subseteq\bbR^{d-1}$ for some $c<1$.
\end{enumerate}

Choose a fixed response interval $R=[-C_YR_0,C_YR_0]$ and a positive integer $l$. For $1\leq h\leq l$, let $\Rlh$ denote the $h$th of its $l$ equal bins, of width $2C_YR_0/l$, assigning each shared endpoint to one bin.

\begin{enumerate}[label=\textbf{($\bm{\omega_f}$)}]
\item \label{Omega} The link is coarsely monotone in the sense of \eqref{eq:coarse-link}. We choose the response scale and slice width so that
\[
 \mathcal R_f+[-2\sigma_\zeta,2\sigma_\zeta]\subseteq R,\qquad
 C_YR_0\lesssim|\mathcal R_f|,\qquad
 \max(\sigma_\zeta,\omega_f)\leq|\Rlh|\leq|\mathcal R_f|/3 .
\]
\end{enumerate}

Write $w=C_f|\Rlh|$ for the corresponding arclength scale. The next condition identifies the region where the arclength density is large enough to populate nearby slices.

\begin{enumerate}[label=\textbf{($\bm{\rho}$)}]
\item \label{rho} For the arclength density $\rho_t$ use its interior traces under \ref{Xsub} and \ref{gamma1} at $0$ and $\len_\gamma$. There is $C_\rho<\infty$ such that $\rho_t\leq C_\rho$ on $[0,\len_\gamma]$. For $c_\rho\in(0,1/\len_\gamma]$ and a number of slices $l$, we let
\[ \Omega_0:=\Bigl\{x\in\Omega_\gamma:\ \rho_t\geq c_\rho\ \text{ on }\ [\Pi_\gamma x-2w,\ \Pi_\gamma x+2w]\cap[0,\len_\gamma]\Bigr\}. \]
\end{enumerate}

The following two conditions describe the thin- and wide-slice regimes, respectively.

\begin{enumerate}[label = \textcolor{black}{\rm\textbf{(LCV)}}]
\item \label{LCV} We assume that $\sigma_\gamma \geq C_{\mathrm{LCV}}C_f\max(\sigma_\zeta,\omega_f)$, where $C_{\mathrm{LCV}}>0$ is a sufficiently large constant.
\end{enumerate}

\begin{enumerate}[label = \textcolor{black}{\rm\textbf{(SC)}}]
\item \label{SC} There exist $c_1,c_2>0$ such that
\[ \bar\sigma_\gamma<c_1C_f\max(\sigma_\zeta,\omega_f),\qquad C_f\max(\sigma_\zeta,\omega_f)\leq c_2\reach_\gamma. \]
The essential upper bound $\bar\sigma_\gamma$ controls the normal covariance in each wide slice.
\end{enumerate}

In the statistical bounds $C$ may vary from line to line and depends only on $C_\rho/c_\rho$, $C_f/C_f'$, $c$ and $C_X$. In the thin-slice bounds $C$ may also depend on $\bar\sigma_\gamma/\sigma_\gamma$. This convention applies to implicit constants and to the fixed constants in the assumptions and tuning parameters.

\noindent{\bf Step 1:} extract geometric features of the underlying curve $\gamma$.
Taking $n$ divisible by four, randomly split and relabel the sample into two halves, and split the first half into two quarters. Use the fixed partition of $R$ specified above. We use the first quarter to estimate the slice geometry and the second half for regression. The second quarter is unused.
We form the empirical slice pairs $\{(\hS_{l,h}, \hR_{l,h})\}_{h=1}^l$ from the first quarter, where $\hR_{l,h}:=\{Y_i:i\leq n/4,\ Y_i\in\Rlh\}$ and $\hS_{l,h}:=\{X_i:i\leq n/4,\ Y_i\in\Rlh\}$. Each empirical slice $\hS_{l,h}$ is the empirical pre-image of the interval $R_{l,h}$ in the output variable $Y$. For each slice with $\bbP(Y\in\Rlh)>0$, let $\Slh$ denote the conditional distribution $X\mid Y\in\Rlh$. Then $\hS_{l,h}$ is the empirical version of $\Slh$, and its moments approximate those of $\Slh$ when the slice contains sufficiently many observations.
For each slice retained for regression, compute its empirical mean $\hmulh$ and use PCA to select a unit eigenvector $\hvlh$ of its covariance, called its significant vector. The rule below selects the smallest or largest principal component according to the shape of the slice. The mean and this direction yield a local first-order approximation to the curve.

\noindent{\bf Step 2: } design a distance function $\dist(x,h)$ (and its empirical version $\hdist(x,h)$), based on the geometric shape of $\Slh$ (and $\hS_{l,h}$ respectively),
that measures the distance from $x$ to the slice $\Slh$ ($\hS_{l,h}$, respectively).
By assigning each $x\in\Omega_\gamma$ to the ``nearest'' slice according to this distance function, we partition the domain $\Omega_\gamma$ into several local neighborhoods. The same rule assigns each input in the second half to a local neighborhood using its $X$-value only. Because the significant vector $\hvlh$ is approximately tangential to the curve $\gamma$, the centered linear projection provides a local coordinate along the curve, up to the sign of $\hvlh$.

\noindent{\bf Step 3.} Fit a piecewise constant regression on the projected coordinate in each local neighborhood using the second half of the sample.

\subsection[Extracting the geometric features of the underlying curve]{Extracting the geometric features of the underlying curve $\gamma$}\label{Subsection: extracting geometric features}

Let $t=\Pi_\gamma x\in(0,\len_\gamma)$ and $\norm{x-\gamma(t)}<\reach_\gamma$. The closest-point condition gives $\ip{x-\gamma(t)}{\gamma'(t)}=0$.
The derivative of this equation with respect to $t$ is $-1+\ip{x-\gamma(t)}{\gamma''(t)}<0$. Implicit differentiation and the chain rule give, wherever $f$ is differentiable at $t$,
\begin{equation}\label{e:gradientF}
 \nabla F(x)=\frac{f'(t)}{1-\ip{x-\gamma(t)}{\gamma''(t)}}\,\gamma'(t)\,.
 \end{equation}
Each interior fiber $\Pi_\gamma^{-1}(t)$ lies in the affine hyperplane $\gamma(t)+\gamma'(t)^\perp$. At regular points on this fiber the gradient, when it exists, is parallel to $\gamma'(t)$. At the two endpoints the normal-displacement model \eqref{e:Xsplit} places sampled points in the corresponding hyperplanes.
Thus $\gamma'(t)$ is a null direction of the centered data on an interior fiber, and it is the unique such direction up to sign when the data span the normal hyperplane affinely.

The geometry of the level sets plays a central role in our estimator, as it does in earlier work on the single-index model, including \citep{Stoker86, HS89}, and in recent refinements such as the Smallest Vector Regression estimator of \cite{LMV22B}.
For a segment of direction $v$, the interior fibers of $\Pi_\gamma$ lie in parallel hyperplanes perpendicular to $v$. If $f$ is injective, each interior index value determines a distinct response level.
For each interval $\Rlh$ of the fixed partition of $R$, the first quarter of the sample gives the empirical slice \begingroup
$$\hS_{l,h}:=\{X_i:i\leq n/4,\ Y_i\in\Rlh\}\,,$$
\endgroup which is a sample from the conditional distribution $X\mid Y\in\Rlh$.
Each empirical slice $\hS_{l,h}$ is utilized in \cite{LMV22B} as an approximation to a fiber $\Pi_\gamma^{-1}(s_0)$ at an arclength value $s_0$, and $\hS_{l,h}$ should be ``thin'' along the direction of $v$ (and therefore $\nabla F$) and ``wide'' on directions orthogonal to $v$, at least under suitable assumptions on $F$ and the noise level $\sigma_\zeta$.
\citet{LMV22B} then perform local PCA on each $\hS_{l,h}$ and use the smallest principal component to approximate the direction of $v$: since $\gamma$ is a line with direction $v$, the per-slice estimates target the same line $\spn\{v\}$ and can be combined after accounting for the arbitrary signs of the eigenvectors.
Once $v$ is estimated, the input data is projected onto this line, and one-dimensional nonparametric regression yields an estimator for $f$.

\begin{figure}[t]\centering
\floatbox[{\capbeside\thisfloatsetup{capbesideposition={right,top},capbesidewidth=0.5\textwidth}}]{figure}[\FBwidth]
{\caption{In the same setup as in Figure~\ref{Figure: 36 dimensional Meyer-Helix-L1-C1-variant1}, we partition the range uniformly into $l=800$ intervals, and consider two slices.
Top: a visualization of the two empirical slices, where we show only 2000 samples per slice (in green and blue), with $\gamma$ in black. The red circles and vectors are the sample means and smallest principal components of the two empirical slices. Bottom: bar plots with the largest, second largest, average, second smallest, and smallest singular value of the empirical slices. The smallest singular value is \emph{significantly} smaller than the others.}\label{Figure: thin slices of 36 dimensional Meyer-Helix-L1-C1-variant1}}
{\includegraphics[width=0.5\textwidth]{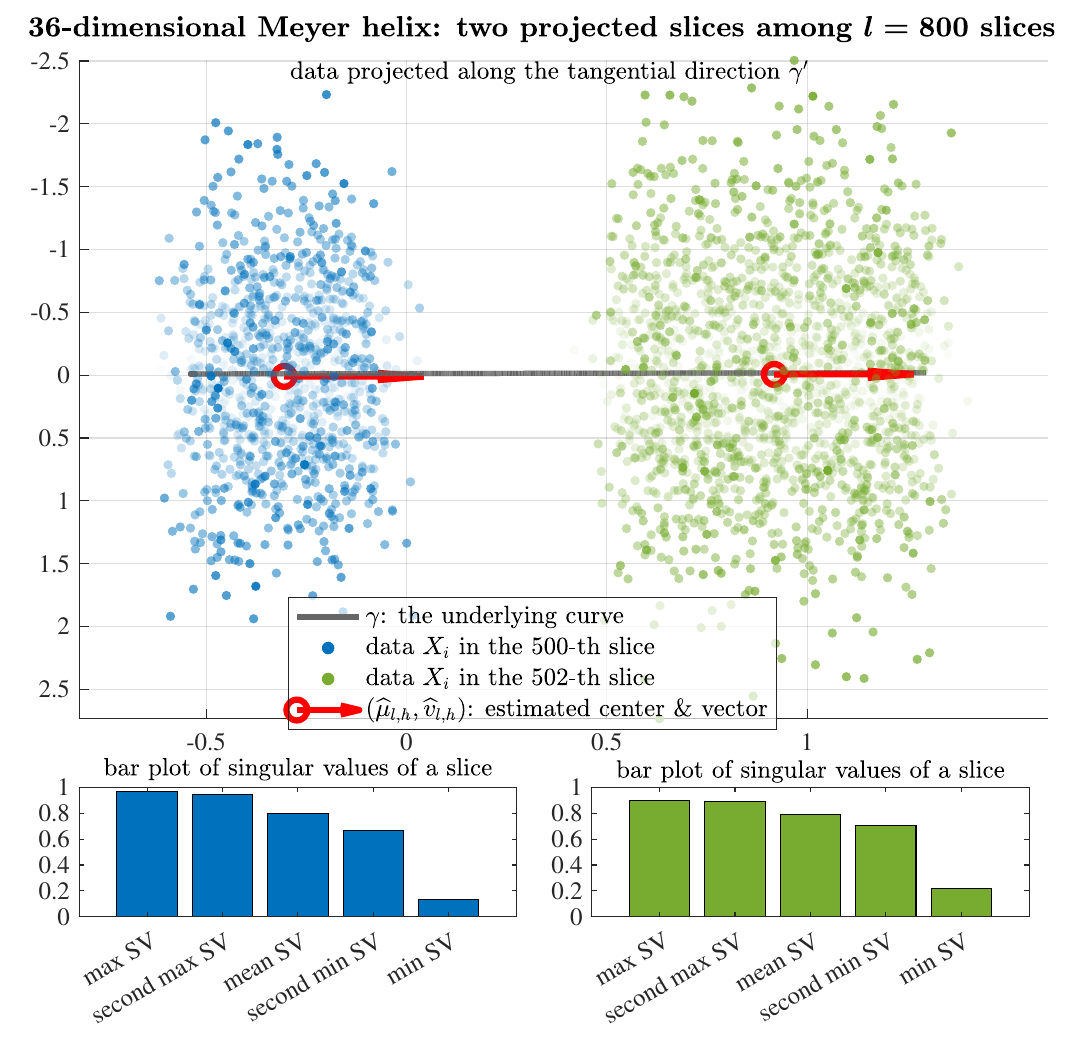}}
\end{figure}

In the NSVM, we still use empirical slices, but here we cannot perform an aggregation of estimated vectors because the tangent vectors to $\gamma$ are not constant due to the curvature of $\gamma$. Instead, we can only rely on local information to estimate the closest-point projection onto $\gamma$, which is now a nonlinear function.
Under Assumption \ref{LCV}, each $\hS_{l,h}$ still approximates a level set $\Pi_\gamma^{-1}(t_0)$: roughly speaking, when $\Rlh$ has sufficiently small diameter, $\hS_{l,h}$ is ``thin'' along the tangential direction $\gamma'(t_0)$ and ``wide'' along directions orthogonal to $\gamma'(t_0)$.
Consequently, local PCA on each empirical slice $\hS_{l,h}$ yields:
\begin{enumerate}[label=(\roman*)]
\item the empirical mean $\hmulh$, which approximates the conditional expectation $\mulh:=\bbE[X\mid Y\in\Rlh]$ and is expected to lie near $\gamma$;
\item the empirical covariance matrix $\hSiglh$, which approximates the conditional covariance matrix $\Siglh:=\bbE[(X-\mulh)(X-\mulh)^\intercal\mid Y\in\Rlh]$;
\item the smallest principal component $\hvlh$ of $\hSiglh$, which approximates the smallest principal component $\vlh$ of $\Siglh$ and is expected to be tangent to the curve.
\end{enumerate}
Together, these quantities yield a local first-order approximation of $\gamma$ near $\hS_{l,h}$. Figure \ref{Figure: thin slices of 36 dimensional Meyer-Helix-L1-C1-variant1} depicts an example where the sample mean is approximately on the curve, and the smallest principal component vector is approximately tangential to the curve.

\begin{figure}[t]\centering
\floatbox[{\capbeside\thisfloatsetup{capbesideposition={right,top},capbesidewidth=0.5\textwidth}}]{figure}[\FBwidth]
{\caption{ In the same setup and visual conventions of Figure~\ref{Figure: 36 dimensional Meyer-Helix-L1-C1-variant1}, but with the range $R$ split uniformly into 80 intervals, and a different pair of slices.
Top: the slices are now elongated along the curve, rather than perpendicularly to it.
Bottom: the {\em{largest}} singular value is now significantly larger than the remaining ones. }\label{Figure: wide slices of 36 dimensional Meyer-Helix-L1-C1-variant1}}
{\includegraphics[width=0.5\textwidth]{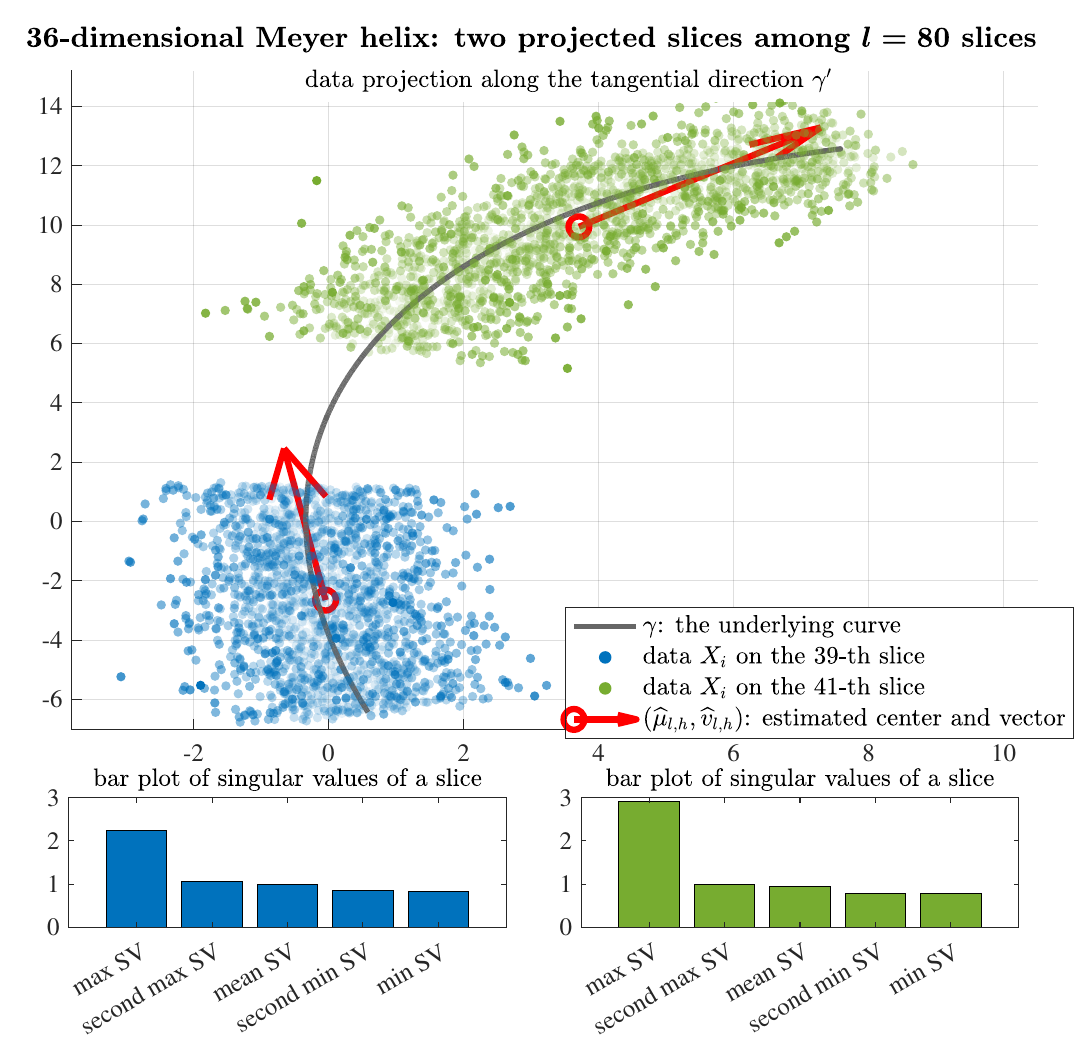}}
\end{figure}

This argument relies on Assumption \ref{LCV}, which constrains the geometry of $\gamma$, the radius of $\Omega_\gamma$ around the curve, the regression function $F$, and the noise level $\sigma_\zeta$.
Without such an assumption, it might not be the case that slices are ``thin'' in the direction tangent to the curve for various reasons:
(i) with an insufficient number of samples, choosing very fine partitions of the range $R$ is not optimal: smaller intervals will contain fewer samples, leading to high variance in estimating the slice means and principal components;
(ii) observational noise contributes the term $C_f^2\sigma_\zeta^2$ to the upper bound on within-slice arclength variance in Proposition~\ref{Prop: Pi_Gamma_X}(b). The proposition applies at widths $|\Rlh|\geq\max(\sigma_\zeta,\omega_f)$.
(iii) the reach $\reach_\gamma$ may be small because of the complexity of the underlying curve $\gamma$;
(iv) the distribution of $X$ might strongly concentrate around $\gamma$, i.e., $\sigma_\gamma$, the lower conditional standard deviation in a normal direction, is small.
In these situations, we might encounter slices that are ``wide'', i.e., much wider in the curve's local tangent direction than in all other directions—essentially the opposite of being ``thin''.
In this case, the largest principal component can be significantly larger than the remaining components and is the one that aligns with the tangent direction of $ \gamma $.
Figure \ref{Figure: wide slices of 36 dimensional Meyer-Helix-L1-C1-variant1} illustrates how, in these situations, the sample mean and the {\em{largest}} principal component of a slice may provide an approximation of the underlying curve $\gamma$. In this setting, the largest singular value is significantly larger than the others.

Consider a slice $\Slh$ with conditional mean $\mulh$ and conditional covariance matrix $\Siglh$, together with their empirical counterparts $\hS_{l,h}$, $\hmulh$, and $\hSiglh$.
For a symmetric positive semidefinite $d\times d$ matrix $M$, write $\lambda_1(M)\geq\cdots\geq\lambda_d(M)$ for its eigenvalues and $v_i(M)$ for a unit eigenvector associated with $\lambda_i(M)$. For $d\geq3$ let $\lambda_{\mathrm{mid}}(M)=(\lambda_2(M)\cdots\lambda_{d-1}(M))^{1/(d-2)}$ be the geometric mean of the intermediate eigenvalues. When the eigenvalues are positive, the population and empirical shape scores are
$$ G_{l,h} := \log\rbracket{\frac{ \lamlh{\mathrm{mid}}^2 }{\lamlh{1} \lamlh{d} }}, \ \ \hG_{l,h} := \log\rbracket{\frac{ \hlamlh{\mathrm{mid}}^2 }{\hlamlh{1} \hlamlh{d} }}\,,$$
For the branch conditions, $G_{l,h}>0$ denotes $\lamlh{\mathrm{mid}}^2>\lamlh{1}\lamlh{d}$ and $G_{l,h}\leq0$ denotes the complementary inequality. We use these comparisons also at zero eigenvalues, with the same convention for $\hG_{l,h}$.
A slice $\Slh$ is ``thin'' when $\lamlh{d}\ll\lamlh{1} \approx \lamlh{2 }\approx \cdots \approx \lamlh{d-1}$, and hence $G_{l,h}>0$; the larger $G_{l,h}$ is, the ``thinner'' $\Slh$ is.
A slice $\Slh$ is ``wide'' when $\lamlh{1}\gg\lamlh{2 }\approx \cdots \approx \lamlh{d}$, and hence $G_{l,h}<0$; the more negative $G_{l,h}$ is, the ``wider'' $\Slh$ is.
If $G_{l,h}$ is close to zero, then the geometric shape of the slice $\Slh$ is undetermined: it may be roughly isotropic or may have both very large $\lamlh{1}$ and very small $\lamlh{d}$.
We define the {\em{significant vector $\vlh$}} and the \emph{empirical significant vector} $\hvlh$ as
$$ \vlh := \begin{cases} v_d(\Siglh) & \text{ if } G_{l,h} > 0 \\ v_1(\Siglh) & \text{ otherwise} \end{cases}, \ \ \ \ \hvlh := \begin{cases} v_d(\hSiglh) & \text{ if } \hG_{l,h} > 0 \\ v_1(\hSiglh) & \text{ if } \hG_{l,h}\leq0 \end{cases}\,.$$
The significant vector $\vlh$ is used to estimate the tangent vector $\gamma'$ in both the ``thin'' and ``wide'' slice scenarios.
When $G_{l,h}\approx0$, the score alone gives little information about which principal direction follows the curve. We use the sample mean $\hmulh$ as the slice center.
Crucially, while in the ``thin'' scenario we expect $\hvlh$ to be a good approximation of the tangential vector, in the ``wide'' scenario the curvature of $\gamma$ can have a significant effect in our estimation of the local direction of the curve.

\subsection{Estimating the nonlinear projection $\Pi_\gamma$ by assigning points to the ``nearest'' slice}\label{Subsection: classifying points to nearest slice}

Before regressing $f$, we construct an estimator for $\Pi_\gamma(x)$, which is the value of the parameter in $[0,\len_\gamma]$ of the curve such that $\gamma(\Pi_\gamma(x))$ is the closest point on the curve to a point $x\in\Omega_\gamma$.
We retain slices with enough observations to estimate their geometry. Fix the count threshold $n_{loc}=\lceil c_\rho C_f'|\Rlh|n/400\rceil$, write $\nlh=\#\hS_{l,h}$ for the first-quarter count, and set $\calH_l=\{h:1\leq h\leq l,\ \nlh\geq n_{loc}\}$. For each retained slice we define a distance $\dist(x,h)$ and its empirical counterpart $\hdist(x,h)$. Under the hypotheses of Lemma~\ref{lem:assign}, the mean arclength of the nearest retained slice lies within $Cw$ of $\Pi_\gamma x$ throughout $\Omega_0$.
After this assignment, we use $\mulh$ and $\vlh$ to measure displacement along the tangent estimate relative to the slice center.

Our choice of distance function $\dist(x,h)$ is dictated by two purposes. First, it should be ``local'', i.e., the distance between $x$ and the center of a slice should play a role. Second, it should be anisotropic: for model points with $\Pi_\gamma x=t_i$ we have $\innerprod{x-\gamma(t_i), \gamma'(t_i)} = 0$, so the distance to the hyperplane normal to $\gamma'(t_i)$ should play a prominent role, but cannot be too dominant, as there may be multiple slices, even far away from each other, with $\innerprod{x-\gamma(t_i), \gamma'(t_i)}\approx 0$. We therefore consider a distance function of the form $$\abs{\innerprod{x-\mulh,\vlh}} + c \norm{x-\mulh}$$
for some $c>0$.
The value of $c$ cannot be too small, because the distance would then fail for highly self-entangled curves, or too large, because $\dist(x,h)$ should remain small when $x$ is close to the slice $\Slh$.
The optimal choice of $c$ depends on the curve and the distribution of data around it.
Using the arclength scale $w$, set $q_l=C(w/\reach_\gamma)^2$. We use this same weight on all thin slices. The population distances underlying the two empirical rules are
\begin{align} \label{Equation: distance function} \begin{split}
\dist(x, h) & = \begin{cases}
 |\innerprod{x-\mulh,\vlh}|^2 + q_l \norm{x-\mulh}^2 \ \text{ if } G_{l,h}>0\\
 \norm{x-\mulh}^2 + \frac{\lamlh{d}}{\lamlh{1}} |\innerprod{x-\mulh,\vlh}|^2 \ \ \text{ otherwise}
 \end{cases}\,,
\end{split}
\end{align}
and similarly for its empirical counterpart $\hdist(x, h)$. We set the eigenvalue ratio to zero when the largest eigenvalue is zero.
 In the ``thin'' slice scenario, this distance function focuses on measuring the displacement between $x$ and $\Slh$ along the direction $\vlh$, and is less sensitive to the displacement orthogonal to $\vlh$.
In the second branch the squared Euclidean distance has coefficient one, with an additional projected term weighted by the eigenvalue ratio.

In Theorem~\ref{Thm: NLSIM_wide}, we prescribe the second branch and take $\vlh$ and $\hvlh$ to be the largest population and empirical principal components for every retained slice, irrespective of the shape score.

We use \eqref{Equation: distance function} in the convergence analysis. This distance is a simplification of the Mahalanobis distance from $x$ to a slice $\Slh$
$\dist(x, h) := ||\Siglh^{-{1/2}}(x-\mulh)||$,
which downweights the direction of largest variance. In a wide slice the second branch of \eqref{Equation: distance function} instead approaches squared Euclidean distance as $\lamlh{d}/\lamlh{1}$ tends to zero. As it is a bit harder to analyze mathematically than the distance in \eqref{Equation: distance function}, we use it only in numerical simulations.

Define $\hxp$ by $F(x)\in R_{l,\hxp}$, assigning a shared endpoint to a single interval. Given the first-quarter observations, select the retained slice minimizing the empirical distance,
\[
\widehat h(x)=\hhx:=\argmin_{h\in\calH_l}\hdist(x,h),
\]
breaking ties by the smallest index. If $\calH_l$ is empty, return $\hF\equiv0$. Lemma~\ref{lem:assign} bounds $|\hhx-\hxp|$ by a constant throughout the region of sufficient arclength density.

\subsection{Local regression and the global estimator}\label{Subsection: conditional partitioning estimators}
For each $h\in\calH_l$, center the projected interval $I^{(l,h)}$ at $\innerprod{\hvlh,\hmulh}$ and give it half-width $r_l=Cw$. Partition it uniformly into $j$ bins. For an integer window radius $a$, let $A_{l,h}:=\{x:|\widehat h(x)-h|\leq a\}$ be the accepted region. Use observations from the second half in this region, taking $a=1$ except in the noiseless theorem. In each nonempty bin fit the constant equal to its mean response. To evaluate at $x$, project onto the line of the selected slice, clip to its interval, use the nearest nonempty bin, and truncate the resulting value to $[-M,M]$, with truncation level $\Linfty{f}\leq M\leq2\Linfty{f}$.

\begin{algorithm}[H]
{\small
\caption{Significant Vector Regression} \label{Alg: NVM}
 \SetKwInOut{Input}{Input}
 \SetKwInOut{Output}{Output}

 \Input{
Samples $\{(X_i,Y_i)\}_{i=1}^n$, resolutions $l,j$, response scale $C_YR_0$, $w=C_f|\Rlh|$, weight $q_l=C(w/\reach_\gamma)^2$, half-width $r_l=Cw$, threshold $n_{loc}=\lceil c_\rho C_f'|\Rlh|n/400\rceil$, and $\Linfty{f}\leq M\leq2\Linfty{f}$. For $d\geq3$ use local polynomial degree $m=0$ and index window $a=1$ unless a theorem specifies a wider window.
 }

 \Output{
$\hF$ estimate of $F$.
 }

 \begin{enumerate}\itemsep0em
\item[\bf{0.}] Randomly split and relabel the sample, taking $n$ divisible by four. Use $i\leq n/4$ for the geometry and intervals, and $i>n/2$ for regression.

\item[\bf{1.a}] Partition $R=[-C_YR_0,C_YR_0]$ into $l$ equal intervals. Set $\hS_{l,h}=\{X_i:i\leq n/4,\ Y_i\in\Rlh\}$.
 \item[\bf{1.b}] Let $\nlh=\#\{i\leq n/4:Y_i\in\Rlh\}$ and let $\calH_l:=\{h\in\{1,\dots,l\}:\nlh\geq n_{loc}\}$. For each $h\in\calH_l$,
compute the sample mean $\hmulh$, covariance $\hSiglh$ (denominator $\nlh$), and its eigenpairs $(\hlamlh{r},\hv_{l,h,r})$ in descending order. Use the first branch when $\hlamlh{\mathrm{mid}}^2>\hlamlh{1}\hlamlh{d}$, and the second otherwise. Set $\hvlh$ to $\hv_{l,h,d}$ in the first branch and $\hv_{l,h,1}$ in the second. In Theorem~\ref{Thm: NLSIM_wide} select the second branch for every slice. If $\calH_l=\emptyset$, return $\hF\equiv0$.

\item[\bf{2.a}] Compute the empirical distance in \eqref{Equation: distance function} with $\hmulh,\hSiglh,\hvlh$ in place of $\mulh,\Siglh,\vlh$, using the branch selected in Step~1.b. Set the eigenvalue ratio to zero if $\hlamlh{1}=0$.
\item[\bf{2.b}] Let $\hhx$ be the smallest index minimizing $\hdist(x,h)$ over $h\in\calH_l$.

\item[\bf{3.a}] Partition $I^{(l,h)}=[\innerprod{\hvlh,\hmulh}-r_l,\innerprod{\hvlh,\hmulh}+r_l]$ uniformly into $j$ bins $\Ilhjk$, assigning shared endpoints to one bin.

\item[\bf{3.b}] For $h\in\calH_l$ and $k\leq j$, let $n^{(l,h)}_{j,k}$ count the observations $i>n/2$ with $|\widehat h(X_i)-h|\leq a$ and $\innerprod{\hvlh,X_i}\in\Ilhjk$. Retain $\calK_j^{(l,h)}=\{k:n^{(l,h)}_{j,k}\geq1\}$. In each retained bin set $\hfjk{\hvlh}$ equal to the mean of these responses.
\item[\bf{3.c}] For $h\in\calH_l$, compute $ \hfj{\hvlh} (r) = \hfjk{\hvlh}(r)$ at a projected coordinate $r$ using $k=k^{(l,h)}(r):=\argmin_{k\in\calK_j^{(l,h)}}\dist(r,\Ilhjk)$, the index of the retained bin nearest to $r$ (prefer the bin containing $r$ in an endpoint tie, then the smallest index), and $\hfj{\hvlh}\equiv0$ if $\calK_j^{(l,h)}=\emptyset$;
let $\Pi_{I}$ be the metric projection onto the closed interval $I$;
then return the estimator truncated at level $M$,
\begingroup
\[ \hF(x)=\bigl(\hfj{\hvlhhx}(\Pi_{I^{(l,\hhx)}}\innerprod{\hvlhhx,x})\vee(-M)\bigr)\wedge M \ \, , \]
\endgroup
 \end{enumerate}
 \vspace{-10pt}
}
\end{algorithm}
\paragraph{Tuning.} The guarantees use the model constants $C_YR_0$ for the response partition, $w$ and $\reach_\gamma$ for $r_l$ and $q_l$, and $c_\rho C_f'$ for $n_{loc}$. The constants in $r_l$ and $q_l$ are sufficiently large. The truncation level $M$ bounds the link in absolute value. In the noiseless theorem the window $a$ is a sufficiently large fixed integer. The wide-slice theorem prescribes the largest component independently of the empirical score.

\subsection{Risk bounds for the estimator}\label{Subsection: main guarantees}
Throughout the three theorems $d\geq3$. In each bound we evaluate the region $\Omega_0$ at the chosen slice resolution $l$. The random split makes the estimated geometry independent of the regression sample. The risk bound \eqref{eq:master} separates approximation within a bin, curvature, error in the estimated direction, the sampling cost of regression (including empty bins), and queries outside $\Omega_0$.

For the thin-slice bounds we set
\begin{align}
 C_{\gamma,f}&:=\Bigl(\frac{C_f}{C_f'}\Bigr)^4 d^4\Bigl(\frac{R_0}{\sigma_\gamma}\Bigr)^8, \label{e:constantsinThm}\\
 M^*&:=\left(\sigma_\zeta^{-1}(C_f C_YR_0)^s\,\Holder{f}{s}\right)^{2/(2s+1)},\qquad
 l_{\max}:=\frac{C C_YR_0}{\max(\sigma_\zeta,\omega_f)}. \notag
\end{align}
Here $M^*$ is used only for positive noise scale $\sigma_\zeta$, and we set $l_{\max}=\infty$ when $\max(\sigma_\zeta,\omega_f)=0$. Fixed-model running-time estimates may also depend on $M^*$ and $l_{\max}$.

\begin{theorem}[Thin slices with observational noise]\label{Thm: NLSIM}
Assume that $\sigma_\zeta>0$, that the conditions \ref{Xsub}, \ref{Ysub}, \ref{zetasub}, \ref{gamma1}, \ref{LCV}, \ref{Omega} and \ref{rho} hold, and that $f\in\mathcal{C}^s(\bbR^1)$ with $s\in[\frac12,1]$. Choose the resolutions, up to integer rounding, by
\[ l^*:=\min\Bigl\{C^{-1}C_{\gamma,f}^{-1}n\log^{-2}n,\ l_{\max},\ M^*n^{\frac{1}{2s+1}}\Bigr\},\qquad j^*:=\Bigl\lceil C\frac{M^*n^{\frac{1}{2s+1}}}{l^*}\Bigr\rceil . \]
Suppose $n$ is such that $l^*\gtrsim C_f\len_\gamma/(C_f'\sigma_\gamma)$. With $\Omega_0$ evaluated at $l^*$, set
\begin{align*}
C_1&:=\Bigl(1+\frac{\Linfty{f}^2}{\sigma_\zeta^2}\Bigr)\Bigl(\frac{C_f}{C_f'}\len_\gamma\sigma_\zeta^2\Bigr)^{\frac{2s}{2s+1}}\Holder{f}{s}^{\frac{2}{2s+1}}\\
&\quad+(\log ed)^2\Holder{f}{s}^2\Bigl(\frac{C_XR_0\,C_fC_YR_0\,d}{\sigma_\gamma}\Bigr)^{2s},\\
C_2&:=\Holder{f}{s}^2\Bigl(\frac{\bar\sigma_\gamma C_f}{\reach_\gamma}\,\frac{C_YR_0}{l^*}\Bigr)^{2s}+\Linfty{f}^2\,\bbP(X\notin\Omega_0).
\end{align*}
Then the estimator constructed by Algorithm~\ref{Alg: NVM} satisfies
\[ \bbE\bigl[|\hF(X)-F(X)|^2\bigr]\lesssim C_1\,n^{-\frac{2s}{2s+1}}\log n+C_2\,. \]
Writing $H=|\calH_{l^*}|$, direct construction takes $O(d^2n+n\log n+(dn+j^*)H)$ operations and evaluation at a new point takes $O(1+dH)$. For a fixed noisy model the construction cost is $O(d^2n\log n)$.
\end{theorem}

The factor $1+\Linfty{f}^2/\sigma_\zeta^2$ accounts for empty bins. The second term of $C_1$ bounds the error in the estimated principal directions. The two terms of $C_2$ measure the geometric error at width $2C_YR_0/l^*$ and the probability of low arclength density.

For a Lipschitz monotone link, taking $s=1$ and $l^*=l_{\max}$ makes the geometric term proportional to $\max(\sigma_\zeta,\omega_f)^2$ at fixed geometry and link constants. This is $O(\sigma_\zeta^2)$ if $\omega_f\lesssim\sigma_\zeta$, in particular for a bi-Lipschitz link with $\omega_f=0$. Under a uniform positive lower bound on the arclength density we may choose $\Omega_0=\Omega_\gamma$, and only this geometric term accompanies the one-dimensional estimation rate.

The rate exponent is independent of the ambient dimension, while the sample-size thresholds and computational constants grow at most polynomially in $d$ for fixed model parameters. The regularity assumptions on $f$ and $\gamma$ are also dimension independent. Moreover, the slice centers and significant vectors provide an interpretable piecewise-linear sketch of $\gamma$ and local estimates of the nonlinear coordinate $\Pi_\gamma$, in addition to the final estimate of $F$.

 \begin{remark}\label{Remark: scales}
The within-bin approximation term in \eqref{eq:master} depends on $l$ and $j$ through their product $lj$. The geometric term depends on the slice width and hence on $l$. We therefore take $l^*$ as large as permitted by the sample-size and noise-scale constraints and allocate the remaining optimal product budget to $j^*$, so that
\begingroup
\[
l^*j^*\asymp M^*n^{1/(2s+1)}.
\]
\endgroup
The sample-size hypothesis ensures that $l^*$ also meets the lower bound required to resolve thin slices.
\end{remark}

 There are three important scope restrictions in our results:
\begin{enumerate}[label=(\roman*)]
\item Coarse monotonicity controls both the length and the diameter of inverse images of response intervals. The density assumption converts the length bound into a lower bound on the probability of nearby slices.
\item Assumption \ref{LCV} requires enough variation normal to the curve relative to the observational noise and the coarse-monotonicity scale. It is central to the thin-slice theorem because it makes the smallest principal component of a slice align with the tangent. Theorem~\ref{Thm: NLSIM_wide} gives a wide-slice saturation bound under its small-curvature and finite-reach conditions.
\item The curvature contribution is proportional to $(\bar\sigma_\gamma C_f C_YR_0/(\reach_\gamma l^*))^{2s}$. In addition, the risk bound contains $\Linfty{f}^2\bbP(X\notin\Omega_0)$.
\end{enumerate}
For a fixed noisy model, the construction cost in Theorem~\ref{Thm: NLSIM} is near-linear in $n$. When many slices are retained, it would be of interest to determine whether a multiscale data structure can reduce the $O(d|\calH_{l^*}|)$ cost of scanning them at each query.

\begin{theorem}[MSE of Algorithm~\ref{Alg: NVM} in the noiseless case]\label{Thm: noiselessNLSIM}
Suppose $\sigma_\gamma>0$ and that \ref{Xsub}, \ref{Ysub}, \ref{zetasub}, \ref{gamma1}, \ref{LCV}, \ref{Omega} and \ref{rho} hold. Assume that there is no observational noise, i.e., $\zeta\equiv0$ almost surely, that the link function $f$ is perfectly monotone, i.e., $\omega_f=0$, and that $f\in\mathcal{C}^s$ with $s\in[\frac12,1]$. Let $c_\rho{}\in(0,1/\len_\gamma]$ be arbitrary, and let $\Omega_0$ be the set defined in \ref{rho} with $l=l^*$. With $C_{\gamma,f}$ in \eqref{e:constantsinThm}, when $\frac{n}{\log^{3/2}n}\gtrsim\frac{C_fC_{\gamma,f}}{c_\rho C_f'\sigma_\gamma}$, if we choose $l^*=\Bigl\lceil C\frac{C_f'c_\rho\len_\gamma}{C_fC_{\gamma,f}}\frac{n}{\log^{3/2}n}\Bigr\rceil$ and $j^*=1$, then the estimation error of Algorithm~\ref{Alg: NVM}, run in Step~3.b with an integer upper bound $K$ on the index error in Lemma~\ref{lem:assign}(ii) and the enlarged window $a=2K+1$, satisfies
\[ \bbE\bigl[|\hF(X)-F(X)|^2\bigr]\lesssim\Holder{f}{s}^2\Bigl(\frac{C_f^2C_{\gamma,f}^2}{c_\rho^2C_f'^2}\frac{\log^3n}{n^2}\Bigr)^{s}+\bbP(X\notin\Omega_0)\Linfty{f}^2\,. \]
For every link in this class, the same estimator also satisfies the bound with first term $\bigl(C_fC_{\gamma,f}/(c_\rho(C_f')^2)\bigr)^2\log^3(n)/n^2$. If $\rho_t$ has a uniform positive lower bound, choose $c_\rho$ below it. Then $\Omega_0=\Omega_\gamma$ and the mean-squared error is $\mathcal O(n^{-2}\log^3n)$ for fixed model parameters throughout the stated range of $s$. Here $\sigma_\zeta=\omega_f=0$ gives $l_{\max}=\infty$.
\end{theorem}

For noiseless regression of an $s$-H\"older function on $[0,1]^k$, \citet[Theorems~1 and~2]{BAUER201793} obtain a sup-norm error of order $(\log n/n)^{s/k}$ in probability when each ball of radius $0<r\leq1$ centered in the cube has design probability at least $c r^k$ for some $c>0$. They prove a matching minimax lower bound over the H\"older class for uniform design. For a one-dimensional Lipschitz link, their sup-norm rate is $\mathcal O_{\mathbb P}(\log n/n)$. Under a positive lower bound on arclength density, our root mean-squared error has the same power of $n$ up to logarithmic factors, even though the arclength coordinate is unknown.
The proof uses a first-order local approximation of the curve and its closest-point coordinate.
In Theorem~\ref{Thm: noiselessNLSIM}, the absence of observational noise and the condition $\omega_f=0$ allow the slice width to decrease with $n$, subject to the sample-size requirement for estimating the geometry within each slice.

\begin{theorem}[NSVM without \ref{LCV}]\label{Thm: NLSIM_wide}
Assume that \ref{Xsub}, \ref{Ysub}, \ref{zetasub}, \ref{gamma1}, \ref{Omega}, \ref{rho} and \ref{SC} hold, with $c_1,c_2$ sufficiently small. Assume also that $f\in\mathcal{C}^s$ with $s\in[\frac12,1]$. Suppose $0<\reach_\gamma<\infty$ and $\max(\sigma_\zeta,\omega_f)>0$. When $\frac{n}{\log(dn)}\gtrsim\frac{C_YR_0^7d^{4}}{C_f^6\max(\sigma_\zeta,\omega_f)^7}$, if we choose $l^*=\Bigl\lceil\frac{C_YR_0}{\max(\sigma_\zeta,\omega_f)}\Bigr\rceil$ and $j^*$ the smallest integer for which
\begin{equation}\label{eqn: jwide} \Holder{f}{s}\Bigl(\frac{C_f\max(\sigma_\zeta,\omega_f)}{j^*}\Bigr)^{s}\leq\Holder{f}{s}\Bigl(\frac{\bar\sigma_\gamma C_f}{\reach_\gamma}\max(\sigma_\zeta,\omega_f)\Bigr)^{s}, \end{equation}
and if, in addition,
\begin{align*}
&\bigl(\sigma_\zeta^2+\Linfty{f}^2\bigr)\frac{l^*j^*}{n}+\Holder{f}{s}^2\Bigl(\frac{\bar\sigma_\gamma R_0^2d}{C_f^2\max(\sigma_\zeta,\omega_f)^2}\Bigr)^{2s}\Bigl(\frac{d\,l^*\log(dn)}{n}\Bigr)^{s}
\lesssim\Holder{f}{s}^2\Bigl(\frac{\bar\sigma_\gamma C_f}{\reach_\gamma}\max(\sigma_\zeta,\omega_f)\Bigr)^{2s},
\end{align*}
then the estimation error of Algorithm~\ref{Alg: NVM}, run with $\hvlh:=\widehat v_{l,h,1}$ and the second line of \eqref{Equation: distance function} for every $h\in\calH_l$, satisfies
\[ \bbE\bigl[|\hF(X)-F(X)|^2\bigr]\lesssim\Holder{f}{s}^2\Bigl(\frac{\bar\sigma_\gamma C_f}{\reach_\gamma}\max(\sigma_\zeta,\omega_f)\Bigr)^{2s}+\Linfty{f}^2\,\bbP(X\notin\Omega_0)\,. \]
\end{theorem}
The choice of $j^*$ controls approximation within a bin. The additional inequality requires the sampling error of the regression and the error in the estimated principal directions to be at most the geometric saturation term. For a straight segment, $\reach_\gamma=\infty$ and the curvature term vanishes. In this case \eqref{eq:master} gives the wide-slice bound for each finite $j\geq1$ under the same width and sample-size conditions, before the choice \eqref{eqn: jwide}.

\section{Analysis of the Estimator}\label{Section: Analysis of the Estimator}
The support ball in Assumption~\ref{gamma1} is centered at the origin of the normal space $\bbR^{d-1}$ and bounds the deviation from the curve, so that $\norm{X-\gamma(\Pi_\gamma X)}\leq c\,\reach_\gamma$ almost surely. Combined with isotropy, it also forces the tube to be thin relative to the reach in every direction: $\sigma_\gamma\lesssim\reach_\gamma/\sqrt{d-1}$.
For example, draw $t$ on $[0,\len_\gamma]$ and, independently, draw $Z_{d-1}$ uniformly in a ball of finite radius $0<b<c\,\reach_\gamma$. In \eqref{e:Xsplit}, the normal displacement is then centered and has covariance $b^2\id_{d-1}/(d+1)$ in normal coordinates. More generally, under \ref{gamma1}, $\Cov(Z_{d-1})=\bbE[\sigma_\gamma(t)^2]\id_{d-1}$ whenever $\bbE[\sigma_\gamma(t)^2]<\infty$. The ambient displacement $X-\gamma(t)$ can have anisotropic covariance, since its conditional covariance acts on the normal space at $t$.

Assumption \ref{gamma1} implies that for any $t_0\in[0,\len_\gamma]$, the conditional mean is on the curve
\[ \mu_{t_0} :=\bbE [X \mid t = t_0] = \gamma(t_0)\,, \]
and the conditional covariance matrix annihilates the tangent line $\spn\{\gamma'(t_0)\}$ and acts by multiplication by $\sigma_\gamma(t_0)^2$ on its orthogonal complement, since
\begin{align*}
\Sigma_{t_0} & := \bbE[ (X-\mu_{t_0})(X-\mu_{t_0})^{\intercal} \mid t=t_0 ] = \sigma_\gamma(t_0)^2 \id_{d} - \sigma_\gamma(t_0)^2 \gamma'(t_0)\gamma'(t_0)^{\intercal}\,.
\end{align*}
If $\sigma_\gamma(t_0)=0$, the zero eigenspace is all of $\bbR^d$.

By Assumption~\ref{rho}, if $\rho_t\geq c_\rho$ throughout $[0,\len_\gamma]$, then $\Omega_0=\Omega_\gamma$. For example, take $t$ uniform on a circular arc and draw the normal displacement independently in normal coordinates with a smooth radial density, written $g(u)$ for a displacement $u\perp\gamma'(t)$ with $\norm{u}<b$ and zero outside this ball. In the tube the ambient density is $g(u)/\bigl(\len_\gamma(1-\ip{u}{\gamma''(t)})\bigr)$ by the normal-coordinate Jacobian. It satisfies the one-sided smoothness condition in \ref{Xsub}, and $\rho_t=1/\len_\gamma$ up to both endpoints.
For any interval $T\subseteq[0,\len_\gamma]$ with $\bbP(t\in T)>0$ and $\bbE[\norm{X}^2\mid t\in T]<\infty$, the conditional moment identities in \ref{gamma1} give the conditional mean
\[ \mu_T :=\bbE [X \mid t\in T] = \bbE [ \gamma(t) \mid t\in T] \]
and conditional covariance matrix
\begin{align*}
\Sigma_T & := \bbE[ (X-\mu_T)(X-\mu_T)^{\intercal} \mid t\in T ] \\
 & = \bbE \sbracket{ (\gamma(t)-\mu_T)(\gamma(t)-\mu_T)^{\intercal} \mid t\in T} + \bbE [\sigma_\gamma(t)^2 \mid t\in T] \id_{d} - \bbE[ \sigma_\gamma(t)^2 \gamma'(t)\gamma'(t)^{\intercal} \mid t\in T] \ \, .
\end{align*}
This identity explains why Significant Vector Regression estimates the tangent direction when the slices are sufficiently thin. If $T$ is small enough that the first term has negligible spectral norm relative to the remaining terms, then the second term is a multiple of the identity and the third is negative semidefinite. Consequently, the smallest principal component of $\Sigma_T$ is approximately the largest principal component of $\bbE[\sigma_\gamma(t)^2\gamma'(t)\gamma'(t)^\intercal\mid t\in T]$, and hence estimates the local tangent direction.

With $\sigma_\gamma>0$, Assumption~\ref{LCV} allows $C_f|\Rlh|$ to be small relative to $\sigma_\gamma$ at widths $|\Rlh|\geq\max(\sigma_\zeta,\omega_f)$. Corollary~\ref{Coro: Thin2} gives an eigengap at such widths for slices of sufficient probability. Theorem~\ref{Thm: NLSIM_wide} treats the separate regime specified by \ref{SC}.

By Assumption~\ref{Omega}, $C_f'|\mathcal R_f|\leq\len_\gamma\leq C_f|\mathcal R_f|$ and $|\mathcal R_f|\asymp C_YR_0$.

Combining two of these gives a fact that is used twice below and is worth isolating. Assumption \ref{gamma1} places the conditional law of the normal component in a ball of radius $c\,\reach_\gamma$ of the normal space $\bbR^{d-1}$ and requires it to be isotropic, which forces $\sqrt{d-1}\,\sigma_\gamma\leq\sqrt{d-1}\,\bar\sigma_\gamma\leq c\,\reach_\gamma$; combining this with Assumption \ref{LCV} gives
\begingroup
\begin{equation}\label{eqn: SCfree}
C_f\max(\sigma_\zeta,\omega_f)\ \leq\ \frac{\sigma_\gamma}{C_{\mathrm{LCV}}}\ \leq\ \frac{c\,\reach_\gamma}{C_{\mathrm{LCV}}\sqrt{d-1}}\,.
\end{equation}
\endgroup
When $l^*=l_{\max}$, \eqref{eqn: SCfree} bounds the geometric part of $C_2$ by a constant times $\Holder{f}{s}^2(\bar\sigma_\gamma/\sqrt{d-1})^{2s}$.

\subsection{Localization and population eigengaps}
Write $t=\Pi_\gamma X$, $W=X-\gamma(t)$, $\tau(u)=\gamma'(u)$, and $\kappa=1/\reach_\gamma\geq\norm{\gamma''}_\infty$. For a slice of positive probability write $\bar t_h=\bbE[t\mid Y\in\Rlh]$ and $\bar\tau_h=\tau(\bar t_h)$. For pointwise statements we consider query points satisfying the support bounds in \ref{Xsub} and \ref{gamma1}, as holds almost surely for a test draw. Orient each population principal component so that its inner product with $\bar\tau_h$ is nonnegative.

\begin{proposition}\label{Prop: Pi_Gamma_X}
Suppose \ref{zetasub}, \ref{Omega} and \ref{rho} hold. Let $T\subseteq R$ be a bounded interval of positive length with $|T|\geq\max(\sigma_\zeta,\omega_f)$ and $\bbP(Y\in T)\geq c_\rho C_f'|T|/600$. Then:
\begin{enumerate}[label=\textup{(\alph*)}]
\item For every $i=1,\dots,n$ and every $\tau\geq1$,
\[ \bbP\bigl\{|\Pi_\gamma X_i-\bbE[\Pi_\gamma X\mid Y\in T]|\gtrsim C_f(|T|+\sqrt{\tau\log n}\,\sigma_\zeta)\bigm|Y_i\in T\bigr\}\leq Cn^{-\tau}. \]
\item $\Var[\Pi_\gamma X\mid Y\in T]\lesssim C_f^2(|T|^2+\sigma_\zeta^2)$.
\item Given $Y\in T$, $|\Pi_\gamma X-\bbE[\Pi_\gamma X\mid Y\in T]|\lesssim C_f(|T|+|\zeta|)$ almost surely, and $\bbP(|\zeta|>r\mid Y\in T)\lesssim e^{-r^2/(2\sigma_\zeta^2)}$ for every $r\geq0$ when $\sigma_\zeta>0$.
\end{enumerate}
\end{proposition}

\begin{proposition}\label{Prop: Thin1}
Suppose \ref{zetasub}, \ref{gamma1}, \ref{Omega} and \ref{rho} hold. Let $T\subseteq R$ be an interval satisfying the width and probability hypotheses of Proposition~\ref{Prop: Pi_Gamma_X}, and assume $\bbE[\norm{X}^2\mid Y\in T]<\infty$. Write $\mu_T$ and $\Sigma_T$ for the conditional mean and covariance of $X$ given $Y\in T$, and $v_T$ for a unit eigenvector of $\Sigma_T$ associated with its smallest eigenvalue.
\begin{enumerate}[label=\textup{(\alph*)}]
\item Under the additional assumptions of Corollary~\ref{Coro: Thin2} with $C_f|T|\leq\sigma_\gamma/C$, for $\tau\geq1$,
\[
\bbP\!\left(\left|\ip{v_T}{X_i-\mu_T}\right|\gtrsim C_f(|T|+\sqrt{\tau\log n}\,\sigma_\zeta)\mid Y_i\in T\right)\leq Cn^{-\tau}.
\]
\item $\lambda_d(\Sigma_T)\lesssim C_f^2(|T|^2+\sigma_\zeta^2)$.
\end{enumerate}
\end{proposition}

\begin{corollary}\label{Coro: Thin2}
Suppose \ref{Xsub}, \ref{Ysub}, \ref{zetasub}, \ref{gamma1}, \ref{LCV}, \ref{Omega} and \ref{rho} hold. Assume $\sigma_\gamma>0$ and $d\geq2$, and let $\vlh$ denote the oriented eigenvector for $\lambda_d(\Siglh)$. Then, for every $l$ such that $l\gtrsim C_fC_YR_0/\sigma_\gamma$ and $|\Rlh|\geq\max(\sigma_\zeta,\omega_f)$, and every $h$ with $\bbP(Y\in\Rlh)\geq c_\rho C_f'|\Rlh|/600$, we have
\[ \lambda_{d-1}(\Siglh)-\lambda_d(\Siglh)\gtrsim\sigma_\gamma^2\qquad\text{and}\qquad\norm{v_{l,h}-\gamma'(\bar t_h)}\lesssim\norm{\gamma''}_\infty C_f|\Rlh|. \]
For $d\geq3$, the population product test selects this smallest principal component.
\end{corollary}
The corresponding wide-slice eigengap is stated in Proposition~\ref{Proposition: wide slice}. We prove the population estimates in Appendix~\ref{app:geometry}.

\subsection{From slice geometry to prediction error}
For each query in $\Omega_0$, Lemma~\ref{lem:near} supplies a nearby slice of sufficient probability. On the event of Lemma~\ref{lem:conc} this slice is retained and the empirical means and covariances are accurate simultaneously on all retained slices. Lemma~\ref{lem:assign} then gives both $|\hhx-\hxp|\leq C$ and coverage of the projected query by its fitted interval. The conditional risk estimate of Lemma~\ref{lem:risk} accounts for variation within each bin, regression noise, empty bins and queries outside the fitted intervals. Combining these estimates yields \eqref{eq:master}.
\section{Numerical Experiments}

The experiments use the computational implementation reported with the original manuscript. It uses the empirical response range, pools training observations for geometry and regression, assigns points by Mahalanobis distance, and uses practical bin-retention and rounding conventions. We use 90\% of each sample for training and 10\% for testing. The theoretical bounds above concern the split-sample estimator with fixed intervals, a common thin-slice weight and degree-zero fits. The experiments compare observed errors with the corresponding rate and saturation scales. Implementation details and reproduction scripts are available in the accompanying repository.\footnote{\url{https://github.com/williamwuyantao/Nonlinear-Single-Variable-Model-scripts}}

All numerical experiments reported below use smooth link functions and degree-one local polynomial regression ($m=1$). Since the links are smooth on their compact domains, they are Lipschitz; consistently with the theoretical range $s\in[\frac12,1]$, we use the rate benchmark at the endpoint $s=1$. We vary $n$ from $10^4$ to $10^6$. The resolutions $l$ and $j$ follow the original experimental protocol. In practice, $l$ and $j$ may instead be selected by ten-fold cross-validation, which gave similar results in the experiments where it was used. Writing $\hF_n$ for the fit obtained from the sample of size $n$, we study its mean squared error $\bbE[|\hF_n(X)-F(X)|^2]$, the tangential error of the estimated slice center, and the difference between the significant vector and the tangential direction. In the Gaussian simulations $\sigma_\gamma$ denotes the normal-coordinate standard deviation used to generate the Gaussian before truncation. Each experiment is repeated independently five times. Section \ref{s:ex_reaction_coordinates} gives a stylized application to learning a reaction coordinate in a high-dimensional stochastic system.

In each example in this section, we randomly generate $n_{\max}:=2\times10^6$ points from the underlying curve $\gamma$, for which we will have a complete parametrization.
We use all these $2\times10^6$ data to compute an approximation to the center $\mulh := \bbE\sbracket{ X \mid Y\in\Rlh }$ and the average tangential vector $\gamma'_{l,h} := \frac{\bbE\sbracket{ \gamma'(\Pi_\gamma X) \mid Y\in\Rlh}}{\norm{\bbE\sbracket{ \gamma'(\Pi_\gamma X) \mid Y\in\Rlh}}}$ on each slice.
Here, the unit vector $\gamma'_{l,h}$ is parallel to the average tangential direction $\bbE\sbracket{ \gamma'(\Pi_\gamma X) \ \mid \ Y\in\Rlh} $ on the slice $\Slh$.

To estimate accurately the nonlinear curve-approximation term $\MSE_{(NCA)}$, which measures the geometric contribution in Theorem \ref{Thm: NLSIM}, we replace the estimated parameters $\{(\hmulh,\hvlh)\}_{h=1}^l$ by the ``oracle'' parameters $\{(\mulh,\gamma'_{l,h})\}_{h=1}^l$ (computed on the $n_{\max}$ points as described above) when performing the local linear projection and the local polynomial regression on each sample slice.
For the oracle nonlinear-curve-approximation diagnostics described above, we retain the same values $(l^*,j^*)$ as above and report the corresponding MSE at $n=n_{\max}$, denoted by ``MSE at $n = 2\times10^6$'' in Figs. \ref{Figure: LearningInfo_arccircle_D20}, \ref{Figure: LearningInfo_MeyerHelixL1V1_D7}, \ref{Figure: LearningInfo_MeyerHelix_D3-9}, \ref{Figure: LearningInfo_MeyerHelix_ZeroNoise}, \ref{Figure: LearningInfo01_wide_Meyer-Helix_D21}, and \ref{Figure: LearningInfo02_wide_Meyer-Helix_D21}.
In this way, we reduce errors arising from the estimation of the parameters $(\hmulh,\hvlh)_{h=1}^l$, so the reported ``MSE at $n = 2\times10^6$'' primarily reflects the nonlinear curve-approximation term $\MSE_{(NCA)}$.

Throughout this section, we use log--log plots to study how the following quantities decay with the number of samples $n$: the mean squared error $\bbE[|\hF_n(X) - F(X)|^2]$, the estimation error of the center along the tangential direction $\bbE[|\innerprod{\hmu_{l^*,h}-\mu_{l^*,h}, \gamma'_{l^*,h} }|]$, and the difference between the significant vector and the tangential direction $\bbE[||\hv_{l^*,h}-\gamma'_{l^*,h}||]$.
On the interval $n\in[10^5,10^6]$, we use linear least-squares regression to estimate the decay rates. If the measured error levels off in this range, the fitted slope has smaller magnitude.
We add a dashed vertical line $n=10^5$ for those figures to emphasize that the learning rate only corresponds to $n\in[10^5,10^6]$.
At $s=1$ the one-dimensional benchmark for the decaying term in the MSE bound is $n^{-2/3}$ up to a logarithmic factor. At fixed slice width the geometric term in the upper bound can remain positive as $n$ grows. The theorem gives no positive lower bound on the MSE.

\subsection{Circular arcs of varying curvature}\label{Example 1}
We begin with circular arcs, the simplest nonlinear curves: they have constant curvature and admit a linear embedding in $\bbR^2$. We embed arcs of fixed length $\len_\gamma=1$ in ambient dimension $d=20$. Again, we fix $\sigma_\gamma=0.5$ and the random vector $Z_{d-1}$ follows the normal distribution $\calN(0,\sigma_\gamma^2 \id_{d-1})$ with truncation at $\norm{Z_{d-1}} < 0.9 \, \reach_\gamma$. The curvature varies from $0.04$ to $0.4$.
We set our upper bound for the curvature to be 0.4 because otherwise $\reach_\gamma$ becomes too small and the variance of $Z_{d-1}$ will no longer be approximately $\sigma_\gamma^2$.

\begin{figure}[!htb]\centering
\includegraphics[width=0.85\textwidth]{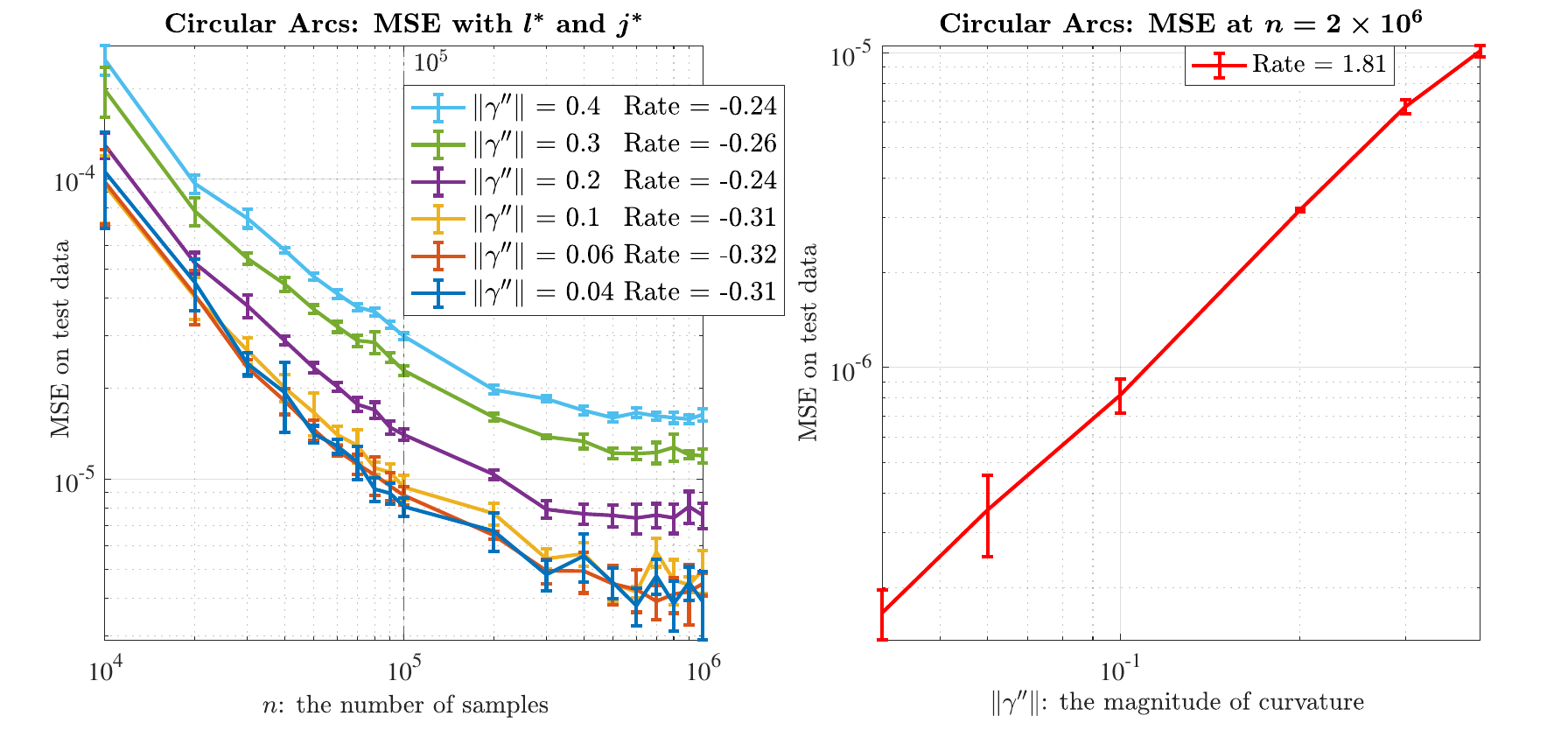} \\
\includegraphics[width=0.85\textwidth]{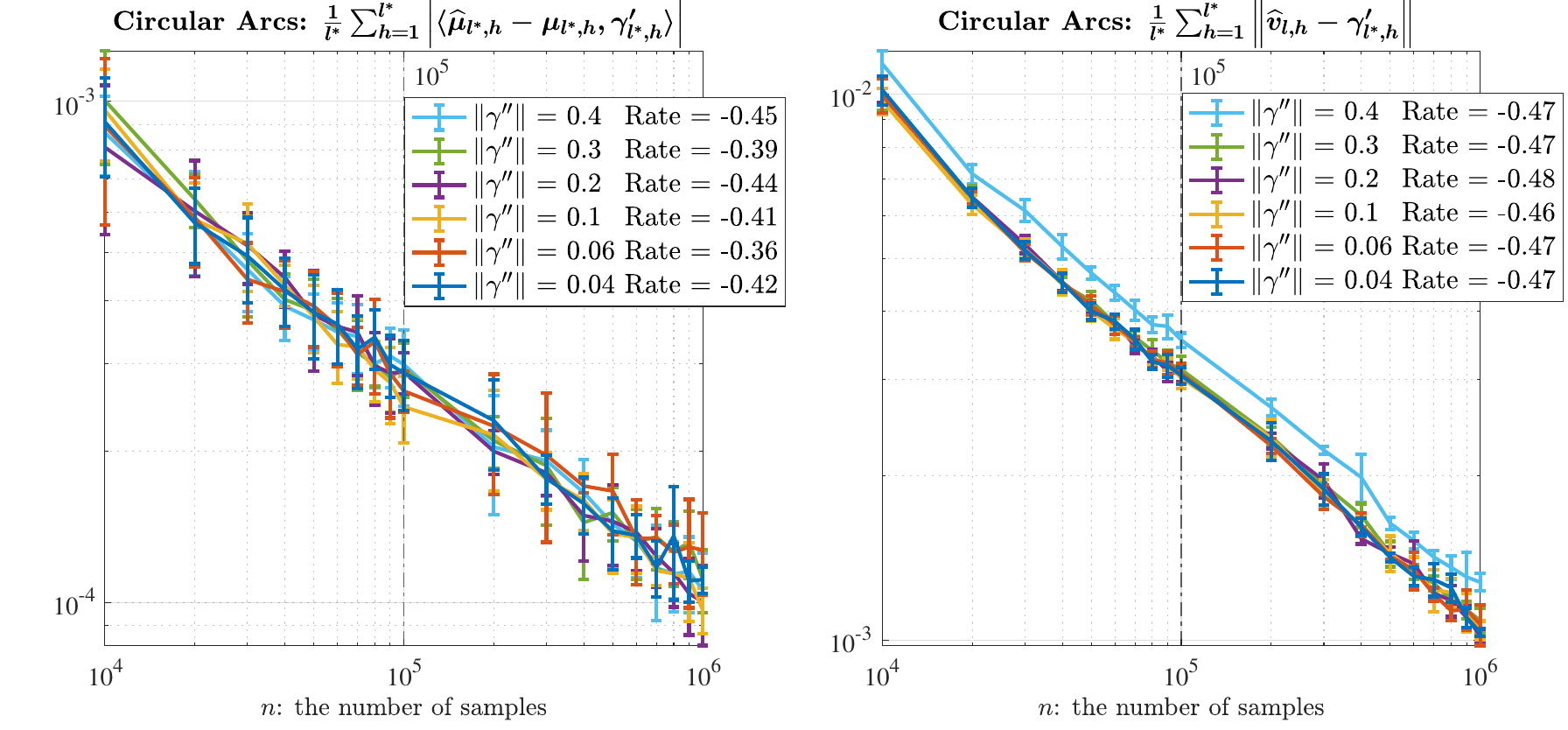}
\caption{Numerical tests in Section \ref{Example 1}: Circular arcs embedded in $\mathbb{R}^d$, $d=20$, with unit length and curvature varying in $[0.04,0.4]$. We fix the noise level $\sigma_\zeta=0.03$.
Upper row: MSE for $\hF$ (left) and MSE at $n=2\times10^6$ as a function of the curvature (right); bottom row: estimation error for the center along tangential direction (left) and difference between estimated significant vector and the tangential direction (right) over $5$ runs.}
\label{Figure: LearningInfo_arccircle_D20}
\end{figure}
At the Lipschitz endpoint $s=1$, the geometric term in Theorem~\ref{Thm: NLSIM} is proportional to $\reach_\gamma^{-2}$ at fixed slice width, link and normal variance. For the circular arcs tested here, the opening angle is less than $\pi$, so the reach equals the radius and $\reach_\gamma^{-2}=\norm{\gamma''}_\infty^2$. We examine this prediction numerically.
We choose the smooth link $f(t)=\exp(t)$ on $[0,1]$. Since it is Lipschitz on this compact interval, we use the benchmark of Theorem~\ref{Thm: NLSIM} at $s=1$.
The observational noise follows the normal distribution $\calN(0,\sigma_\zeta^2)$ with $\sigma_\zeta=0.03$.
Figure \ref{Figure: LearningInfo_arccircle_D20} supports the following conclusions, consistent with Theorem \ref{Thm: NLSIM} and our analysis:
(i) When the observational noise is nontrivial, $\sigma_\zeta>0$, the geometric term of the upper bound can persist at large $n$. In the upper-left panel, the MSE decays more slowly as $n$ increases and changes very little near $n\approx10^6$, which is consistent with the upper bound. The average decay rate over $n\in[10^5,10^6]$ is estimated by least squares.
(ii) The curvature comparison also depends on the post-truncation normal variance, since the truncation radius varies with the reach. The upper-right plot shows that the MSE at $n=2\times10^6$ is larger for curves with higher curvature and smaller for curves of lower curvature. A least-squares fit of the slope is consistent with the theoretical prediction.
(iii) The lower panels report the tangential center error and the error in the significant direction. For the theoretical estimator, Lemma~\ref{lem:conc} and \eqref{eq:thetathin} separate sampling error, which scales as $n_{loc}^{-1/2}$ up to logarithms at fixed slice width, from the geometric error in the direction.

\subsection{Meyer helix}
We perform numerical tests for a collection of curves called Meyer helices in $\mathbb{R}^d$, which we construct so that the curve complexity grows with the ambient dimension $d$.
With $\mathbf 1_E$ denoting the indicator of a set $E$, the Meyer staircase (named after Y.~Meyer) is the map $[0,1]\rightarrow L^p(\mathbb{R})$, for some $p\ge 1$, given by $t\mapsto \mathbf{1}_{[0,\frac12]}(\cdot-t)$.
We smooth out this original example by considering translations of a Gaussian, and we add oscillatory twists to increase the curve's complexity, thereby obtaining the Meyer helix.
Over the dimensions tested, the measured lengths, diameters, maximum curvatures and reaches are consistent with powers $d^{3/2}$, $d^{1/2}$, $d^{-1/2}$ and $d^{1/2}$, respectively. Two measures of effective linear dimension, the sum of the singular values relative to the largest and the number exceeding $5\%$ of the largest, grow approximately linearly with $d$ over this range (Appendix~\ref{Appendix: Meyer-Helix}).
This collection of curves allows us to verify the performance of our algorithm when varying $d$: we fix $\sigma_\gamma=0.5$ and let the random vector $Z_{d-1}$ follow the normal distribution $\calN(0,\sigma_\gamma^2 \id_{d-1})$ with truncation at $|Z_{d-1}| < 0.9 \, \sqrt{d}$.
\subsubsection{Dependence on observational noise}\label{Example 2.1}

 We let the underlying curve $\gamma$ be the Meyer helix in dimension $d=7$, with $\len_\gamma=53.78$ and $\reach_\gamma=2.65$. The link function $f(t)=\len_\gamma\exp(t/\len_\gamma)$ on $[0,\len_\gamma]$ is smooth and hence Lipschitz, so we use the benchmark $s=1$. The observational noise is distributed as $\zeta\sim\calN(0,\sigma_\zeta^2)$, with $\sigma_\zeta$ varying from $0.05$ to $0.2$.
 \begin{figure}[tb]\centering
\includegraphics[width=0.85\textwidth]{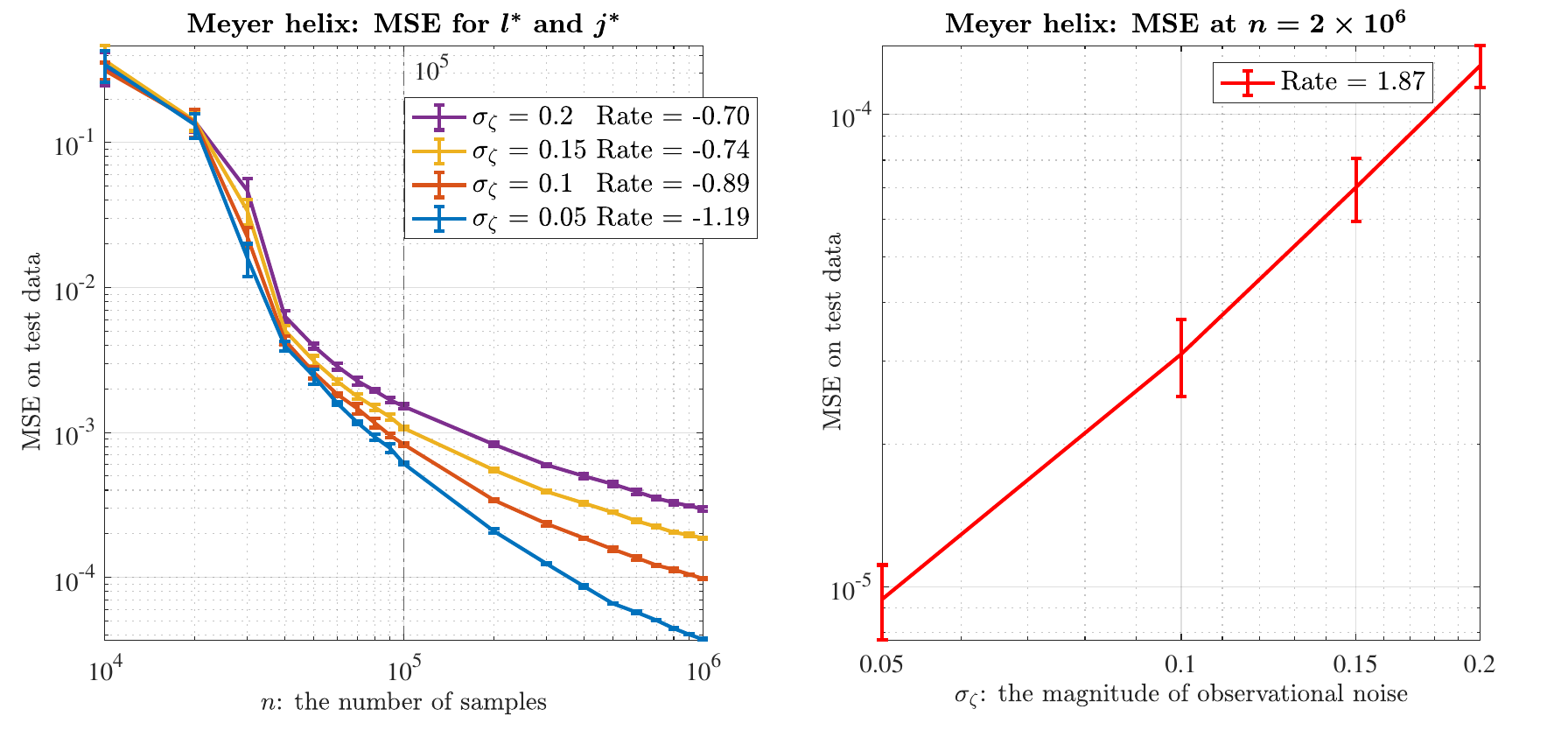}
\includegraphics[width=0.85\textwidth]{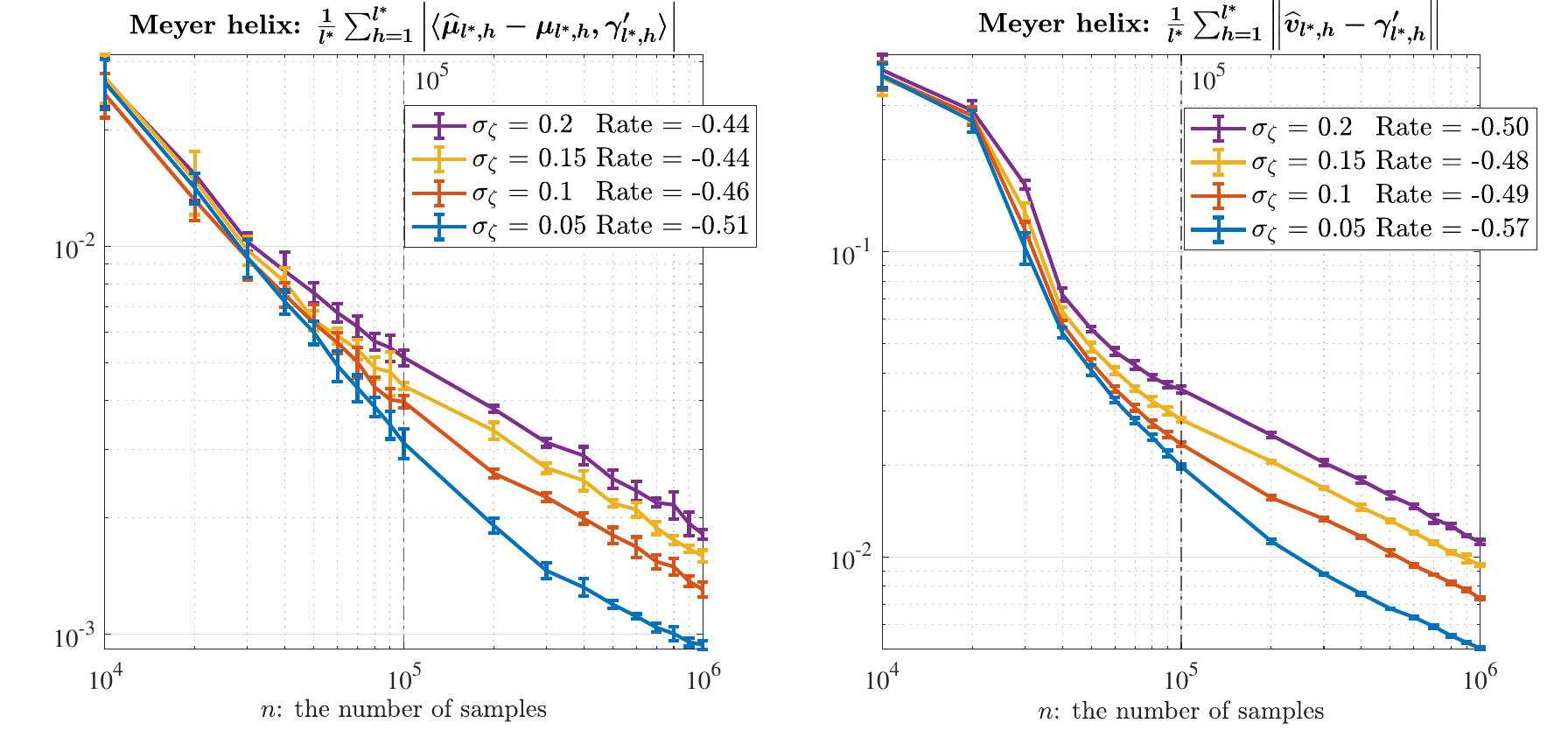}
\caption{Numerical tests in Section \ref{Example 2.1} for a Meyer helix in dimension $d=7$. The link is smooth and monotone; because it is Lipschitz on the compact parameter interval, we use the benchmark $s=1$. The noise level $\sigma_\zeta$ varies over $[0.05,0.2]$. Top row: MSE for $\hF$ (left) and MSE at $n=2\times10^6$ as a function of $\sigma_\zeta$ (right). Bottom row: tangential error of the estimated slice center (left) and error of the estimated significant vector relative to the tangent direction (right), averaged over five independent runs.}

\label{Figure: LearningInfo_MeyerHelixL1V1_D7}
\end{figure}
Figure \ref{Figure: LearningInfo_MeyerHelixL1V1_D7} illustrates the following comparisons:
(i) When $\sigma_\zeta>0$, the measured error levels off at large sample sizes.
(ii) At $s=1$ with $l^*=l_{\max}$ and $\omega_f=0$, the curve-approximation term is proportional to $\sigma_\zeta^{2}$.
(iii) The lower panels show how the estimated centers and directions vary with sample size and noise. The direction bound \eqref{eq:thetathin} also contains a geometric contribution at fixed slice width.

\subsubsection{Dependence on dimension}\label{Example 2.2}

 We consider a collection of Meyer helices with the following ambient dimensions and geometric parameters:
 \begin{table}[H]
 \small{
\begin{center}\begin{tabular}{ ||c|| c c c c c c c|| }
 \hline
 $d$ & 3 & 4 & 5 & 6 & 7 & 8 & 9 \\
 \hline
 $\len_\gamma$ & 20.86 & 33.39 & 35.35 & 43.89 & 53.78 & 90.20 & 96.65 \\
 \hline
 $\reach_\gamma$ & 1.73 & 2.00 & 2.24 & 2.45 & 2.65 & 2.83 & 3.00 \\
 \hline
\end{tabular}\end{center}}\end{table}
We use the smooth link $f(t)=\len_\gamma\exp(t/\len_\gamma)$ on $[0,\len_\gamma]$. It is Lipschitz on this compact interval, so we use the benchmark of Theorem~\ref{Thm: NLSIM} at $s=1$. The observational noise $\zeta$ follows the normal distribution $\calN(0,\sigma_\zeta^2)$ with $\sigma_\zeta = 0.2$.
\begin{figure}[!h]\centering
\includegraphics[width=\textwidth]{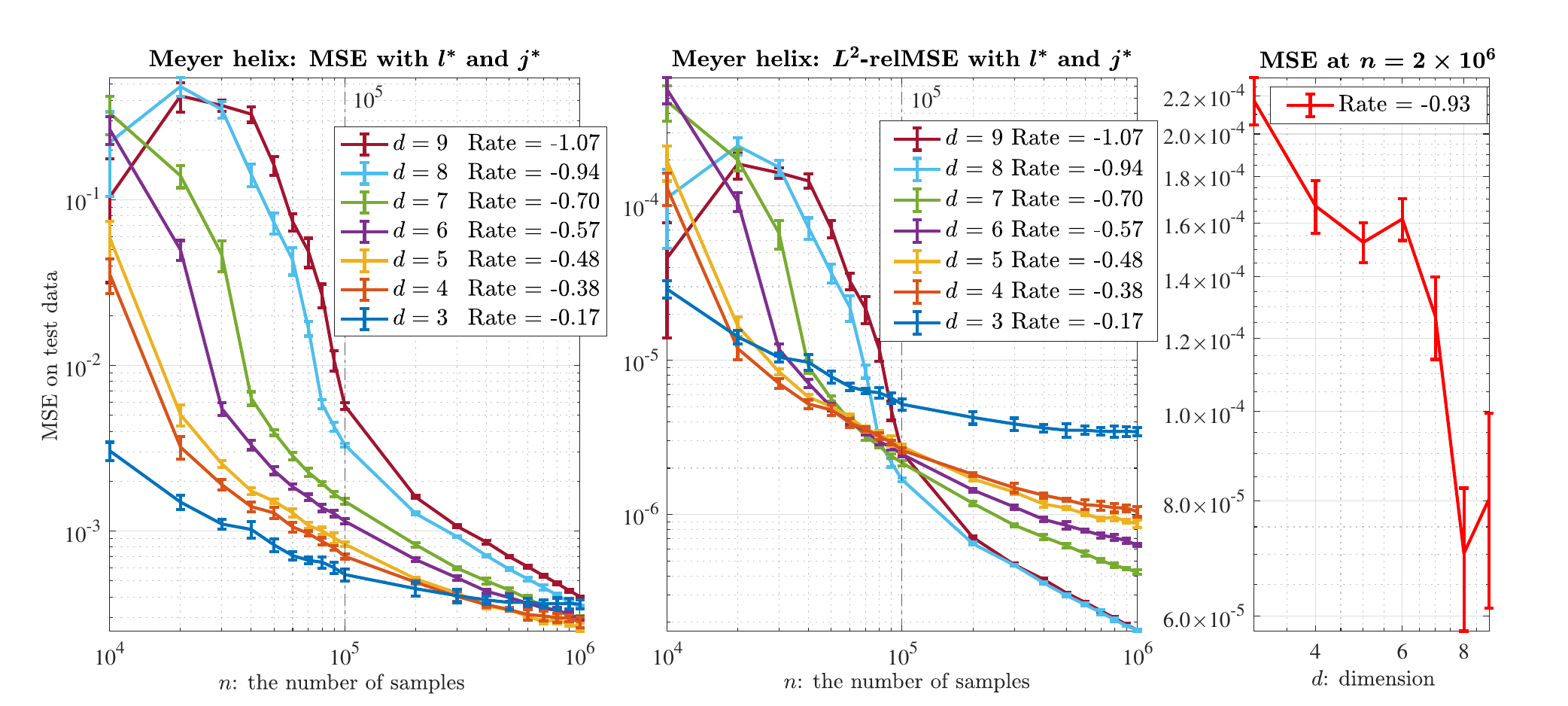}
\caption{Numerical tests in Section \ref{Example 2.2} for Meyer helices in $\mathbb{R}^d$, $d\in\{3,\dots,9\}$. The link is smooth and monotone and is analyzed at the Lipschitz endpoint $s=1$; the noise level is $\sigma_\zeta=0.2$. Left: MSE for $\hF$. Middle: relative MSE, $\bbE[|\hF-F|^2]/\bbE[|F(X)-\bbE(F(X))|^2]$, averaged over five independent runs. Right: MSE of $\hF$ at $n=2\times10^6$ as a function of $d$.}
\label{Figure: LearningInfo_MeyerHelix_D3-9}
\end{figure}

In the left panel of Figure \ref{Figure: LearningInfo_MeyerHelix_D3-9}, we observe that, for fixed $n=10^5$, the mean squared error is larger for Meyer helices with increasing ambient dimension $d$.
Since the smooth link is analyzed as Lipschitz, the benchmark of Theorem~\ref{Thm: NLSIM} at $s=1$ has an estimation term that has rate $n^{-2/3}\log n$; its coefficient depends on $\len_\gamma$, $d$ and the other model parameters.
In the right panel, we observe that, in contrast, the mean squared error at $n=2\times10^6$ generally decreases with $d$, with increases from $d=5$ to $6$ and from $d=8$ to $9$.
At fixed slice width, link and normal variance, the geometric term in the upper bound at $s=1$ is proportional to $\reach_\gamma^{-2}$. Across these helices the length, reach and dimension all vary. In the left panel, the fitted slopes over $10^5\leq n\leq10^6$ are $-0.17$ for $d=3$ and $-0.70$ for $d=7$. These measurements describe the flattening of the observed curves. Theorem~\ref{Thm: NLSIM} gives no sample size at which the observed MSE must level off.
The numerical test also exhibits a transition for $d=8,9$: the error increases as $n$ grows from $10^4$ to $2\times10^4$ and decreases once $n\geq4\times10^4$. The reported average empirical shape score changes from negative to positive and is close to zero for $n\in[3\times10^4,4\times10^4]$. A score near zero is compatible with separated extremal eigenvalues, and scores of opposite signs can cancel across slices. Assessing isotropy requires the individual slice spectra.

\begin{figure}[tb]\centering
\hspace{-3.8mm}
\includegraphics[width=0.405\textwidth]{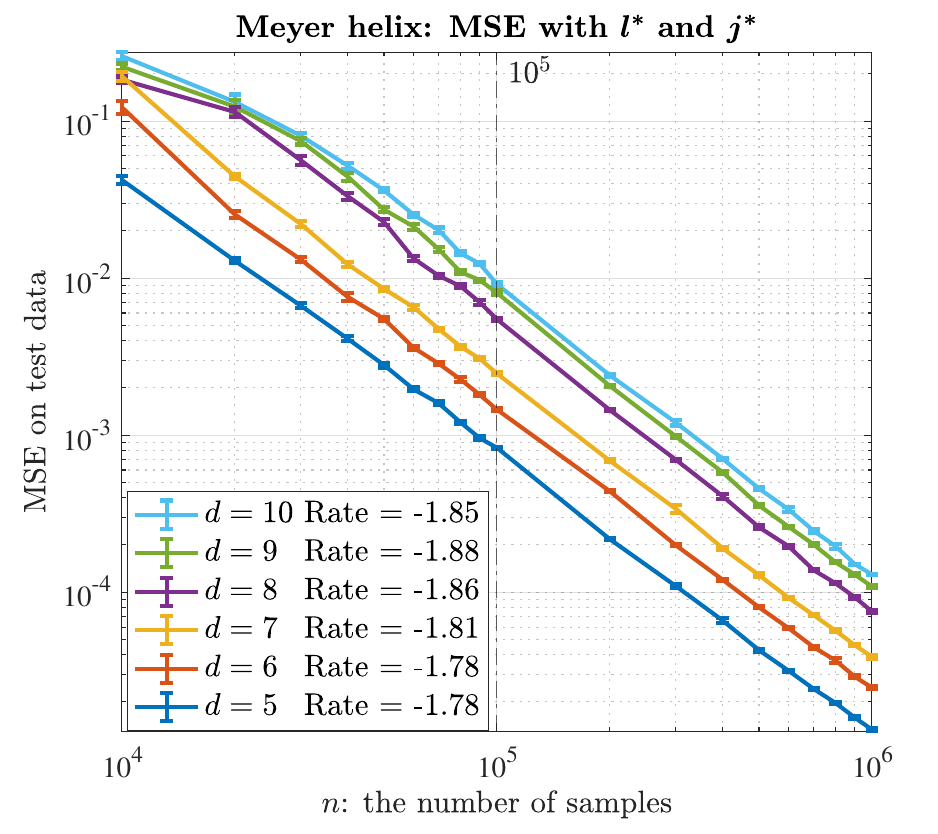} \hspace{0.5mm}
\includegraphics[width=0.405\textwidth]{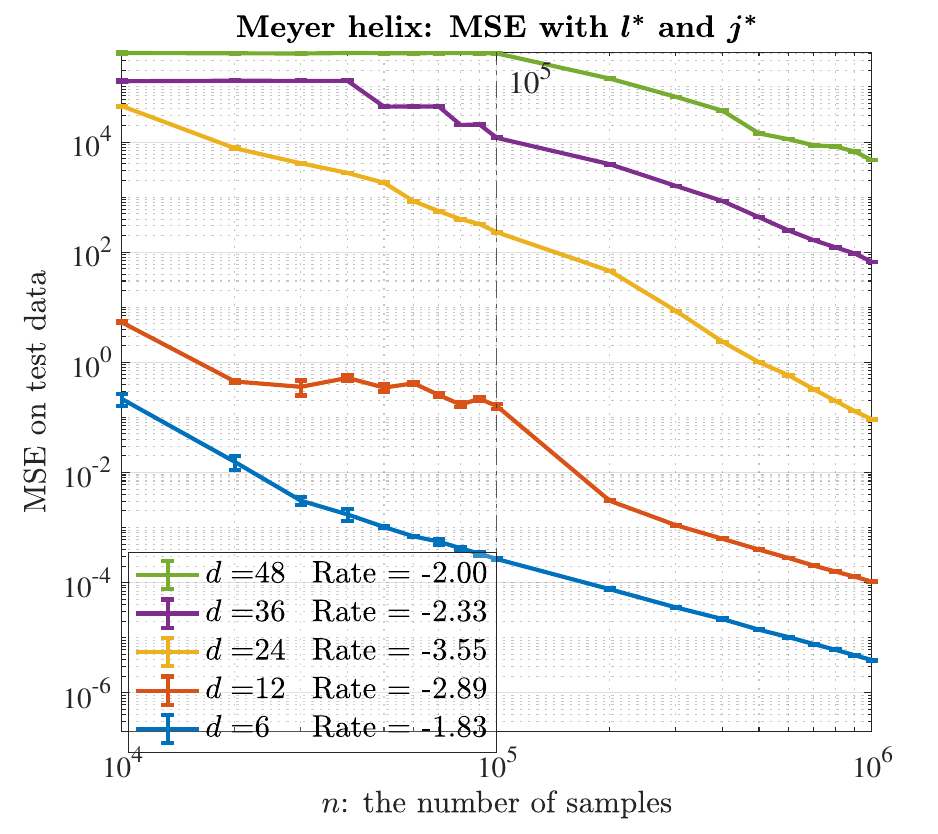}
\caption{Numerical tests in Section \ref{Example 2.3} for Meyer helices of varying ambient dimension. Left: $d=5,\dots,10$. Right: $d=6,12,24,36,48$. In all tests, the observational noise is zero, $\zeta\equiv0$, and the smooth link is analyzed at the Lipschitz endpoint $s=1$. The reported errors are averages over five independent repetitions.}
\label{Figure: LearningInfo_MeyerHelix_ZeroNoise}
\end{figure}
\subsubsection{Noiseless observations}\label{Example 2.3}
We consider a collection of Meyer helices in the following ambient dimensions and geometric parameters:
\begin{table}[H]
\small{
\begin{center}\begin{tabular}{ ||c|| c c c c c c c | c c c||}
 \hline
 $d$ & 5 & 6 & 7 & 8 & 9 & 10 & 12 & 24 & 36 & 48 \\
 \hline
 $\len_\gamma$ & 35.35 & 43.89 & 53.78 & 90.20 & 96.65 & 93.88 & 135.76 & 435.48 & 730.62 & 1306.78 \\
 \hline
 $\reach_\gamma$ & 2.24 & 2.45 & 2.65 & 2.83 & 3.00 & 3.16 & 3.46 & 4.90 & 6.00 & 6.93 \\
 \hline
\end{tabular}\end{center}}\end{table}
In the right panel the absolute errors for $d=24,36,48$ are larger than for $d=6$ at $n=10^4$. The sample-size condition of Theorem~\ref{Thm: noiselessNLSIM} depends on $C_{\gamma,f}$ and the density and normal-variance scales.
Figure \ref{Figure: LearningInfo_MeyerHelix_ZeroNoise} illustrates the following comparisons:
(i) under a uniform positive lower bound on arclength density, the noiseless upper bound has no positive asymptotic floor;
(ii) the Lipschitz, monotone noiseless benchmark under a uniform positive arclength density is $\mathcal O(n^{-2}\log^3 n)$;
(iii) the sample-size condition includes a factor $d^4$ in $C_{\gamma,f}$ at fixed remaining scales.

\subsection{Learning reaction paths and committor functions in overdamped Langevin dynamics}
\label{s:ex_reaction_coordinates}
Let $X_t$ denote the configuration at time $t$, $U$ the potential, $W_t$ a standard $d$-dimensional Brownian motion, and $\sigma_{BM}$ its noise amplitude. The overdamped Langevin dynamics satisfy
\begin{equation}
\label{Equation: SDE}
dX_t=-\nabla U(X_t)\,dt+\sigma_{BM}\,dW_t,
\end{equation}
For two metastable sets $A$ and $B$, let $\tau_A$ and $\tau_B$ be their first hitting times. Writing $\mathbb P_x$ for the law of the dynamics started at $x$, define the committor by
\begingroup
\[
F_A(x):=\mathbb{P}_x(\tau_A<\tau_B).
\]
\endgroup
In the interior outside $A\cup B$, $F_A$ is harmonic with respect to the backward generator
\[
\mathcal{L}=\frac{\sigma_{BM}^2}{2}\Delta-\nabla U\cdot\nabla,
\qquad \mathcal{L}F_A=0,
\]
with boundary values $F_A|_A=1$ and $F_A|_B=0$. Computing this function directly is difficult in high dimension; see \citep{freidlin1998random,e2006towards,metzner2006illustration,metzner2009transition} and the references therein.

Assume that the dominant transition paths lie near a smooth curve $\gamma$. Write $U_{tan}$ for the potential along $\gamma$ and $U_{nor}$ for the confining potential as a function of the distance $\dist_\gamma(x)$ from the curve. We assume the approximate separation
\begin{equation}\label{Equation: CRC_potential}
U(x)=U_{tan}(\Pi_\gamma x)+U_{nor}(\dist_\gamma(x)).
\end{equation}
When the normal confinement is strong, trajectories rapidly relax toward $\gamma$ and evolve primarily through the coordinate $s=\Pi_\gamma X_t$. In the limiting regime the committor therefore factors as
\[
F_A(x)=f(\Pi_\gamma x),\qquad f(s):=F_A(\gamma(s)),
\]
and for large but finite confinement this relation remains an approximation. This is precisely the NSVM structure: $\Pi_\gamma$ is the nonlinear reaction coordinate and $f$ describes progress along the transition path.
As shown in Figure~\ref{Figure: CRC_view}, the latent circular reaction path is not readily visible in either a random three-dimensional projection or the projection onto the first three principal components of the predictor cloud, illustrating the need to learn a nonlinear reaction coordinate.

Training responses can be obtained by Monte Carlo. From an initial condition $x$, simulate $K$ independent trajectories. Denote the hitting times for trajectory $k$ by $\tau_A^{(k)}(x)$ and $\tau_B^{(k)}(x)$. With $\ind\{E\}$ the indicator of the event $E$, set
\begin{equation}\label{Equation: reaction_coordinate_Y}
Y(x):=\frac1K\sum_{k=1}^K\ind\cbracket{\tau_A^{(k)}(x)<\tau_B^{(k)}(x)}.
\end{equation}
Then $\mathbb{E}[Y(x)\mid x]=F_A(x)$ and $\operatorname{Var}(Y(x)\mid x)=K^{-1}F_A(x)(1-F_A(x))$, so the labels have the required conditional mean. Their conditional noise variance varies with the committor, whereas the risk theorems assume observational noise independent of $X$.

\begin{figure}[tb]\centering
\includegraphics[width=0.49\textwidth]{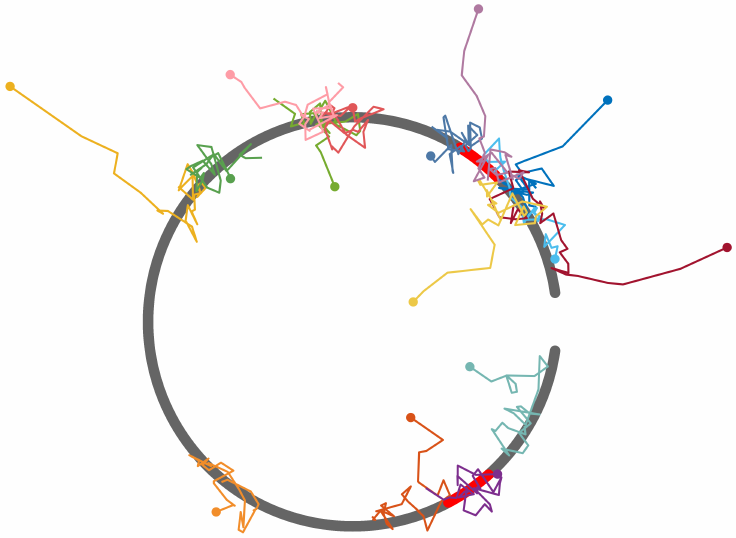}
\includegraphics[width=0.49\textwidth]{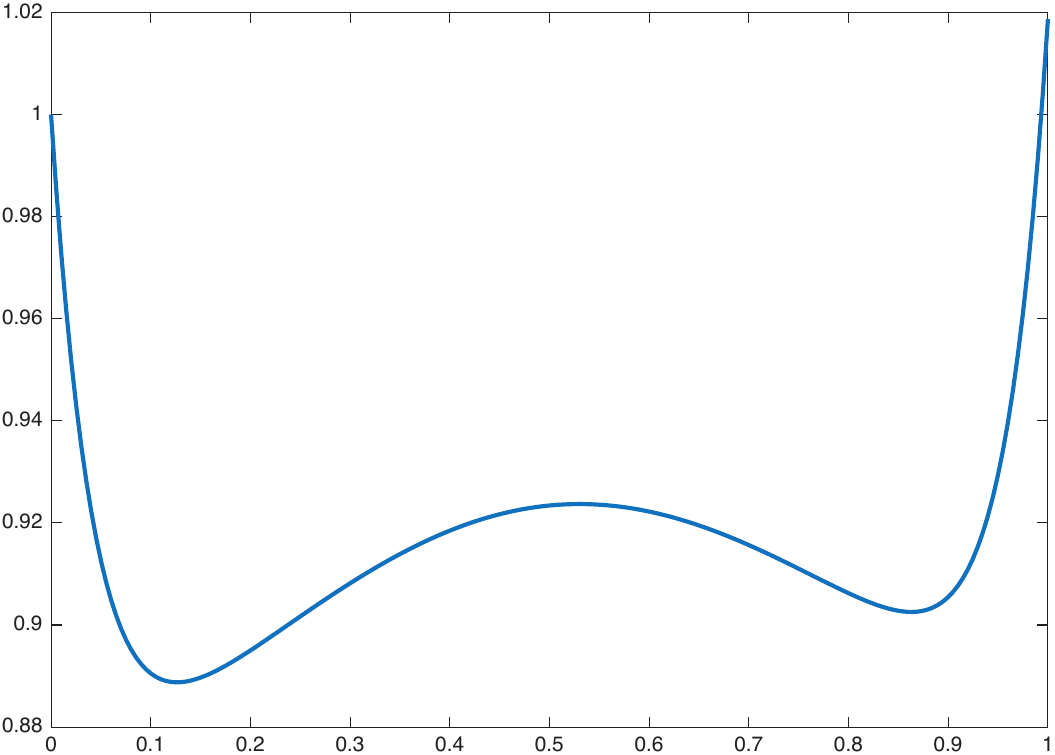}
\caption{Left: short trajectories for \eqref{Equation: SDE} around a circular reaction path; the two metastable basins are highlighted in red. Right: the tangential double-well potential $U_{tan}$. Low-dimensional projections of the sampled configurations are shown in Figure~\ref{Figure: CRC_view} in Appendix~\ref{Appendix: committor_details}.}
\label{Figure: CRC_trajectories}
\end{figure}

For the numerical example, we take $d=40$, a unit-length circular reaction path, a double-well tangential potential, and harmonic normal confinement. We generate $n=4\times10^4$ configurations and estimate each response from $K=4\times10^4$ independent trajectories. The full potential coefficients, basin tolerances, sampling variances, and time-discretization parameters are recorded in Appendix \ref{Appendix: committor_details} and in the public repository.

We apply the pooled numerical estimator for several choices of $l$ and $j$. Because the true committor is unavailable, performance is measured by ten-fold cross-validation against the Monte Carlo labels.
\begin{table}[t]
\centering
\caption{Ten-fold cross-validation MSE $\mathbb{E}[|\hF(X)-Y|^2]$ for the committor experiment.}
\small{
\begin{tabular}{|c | c c c c c c|}
\hline
\diagbox[dir=NW, width=4.5em, height=2.7 em]{$l$}{MSE}{ $\ \ \ j$}
& 1 & 2 & 4 & 8 & 16 & 32 \\
\hline
10 & 0.0022 & 0.0021 & 0.0021 & 0.0021 & 0.0021 & 0.0021 \\
12 & 0.0020 & 0.0019 & 0.0020 & 0.0019 & 0.0019 & 0.0019 \\
14 & 0.0018 & 0.0018 & 0.0018 & 0.0018 & 0.0018 & 0.0018 \\
16 & 0.0018 & 0.0018 & 0.0018 & 0.0017 & 0.0017 & 0.0018 \\
18 & 0.0018 & 0.0017 & 0.0017 & 0.0017 & 0.0017 & 0.0017 \\
20 & 0.0018 & 0.0017 & 0.0017 & 0.0017 & 0.0017 & 0.0017 \\
24 & 0.0021 & 0.0021 & 0.0021 & 0.0021 & 0.0021 & 0.0021 \\
\hline
\end{tabular}}
\end{table}
Several nearby choices of $(l,j)$ attain a cross-validation MSE of $1.7\times10^{-3}$, indicating that the practical performance is stable across a range of resolutions.

\section{Robustness: Performance of Algorithm \ref{Alg: NVM} in a General Setting}
Assumption~\ref{SC} describes the wide-slice regime. It bounds the normal variation relative to the slice scale and the slice scale relative to the reach. Under these conditions the largest principal component estimates the local tangent.

In the thin-slice regime, the second condition follows automatically from \eqref{eqn: SCfree}, with a stronger dimension-dependent bound. The thin-slice condition \ref{LCV} compares the lower normal standard deviation $\sigma_\gamma$ with $C_f\max(\sigma_\zeta,\omega_f)$, whereas the wide-slice condition \ref{SC} requires the upper normal standard deviation $\bar\sigma_\gamma$ to be small relative to this scale. When the normal variance varies along the curve, both conditions can fail.

\begin{proposition}\label{Proposition: wide slice}
Suppose $d\geq2$ and that \ref{Xsub}, \ref{Ysub}, \ref{zetasub}, \ref{gamma1}, \ref{Omega} and \ref{rho} hold, together with \ref{SC}, with $c_1,c_2$ sufficiently small. Then, for every $l$ such that $|\Rlh|\asymp\max(\sigma_\zeta,\omega_f)$, $h\in\calH_l$ with $\bbP(Y\in\Rlh)\geq c_\rho C_f'|\Rlh|/600$, we have
\[ \lambda_1(\Siglh)-\lambda_2(\Siglh)\gtrsim C_f^2|\Rlh|^2\qquad\text{and}\qquad\norm{v_{l,h}-\gamma'(\bar t_h)}\lesssim\norm{\gamma''}_\infty C_f|\Rlh|. \]
\end{proposition}
Here $\vlh$ is the largest population principal component. For this regime we prescribe the largest empirical component and the second distance in \eqref{Equation: distance function} on every retained slice. The eigengap controls the direction error in \eqref{eq:thetawide}, and the assignment bound of Lemma~\ref{lem:assign} applies to this choice.

\subsection{Wide slices on a Meyer helix}\label{Example 2.4}
We examine how the empirical saturation level depends on observational noise and normal variation in the wide-slice regime.
Let the underlying curve $\gamma$ be the Meyer helix in dimension $d=21$, with $\len_\gamma=370.63$ and $\reach_\gamma=4.58$.
The link function $f(t)=\len_\gamma\exp(t/\len_\gamma)$ on $[0,\len_\gamma]$ is smooth and hence Lipschitz, so we use the benchmark $s=1$. The observational noise is distributed as $\zeta\sim\calN(0,\sigma_\zeta^2)$, with $\sigma_\zeta$ varying from $2$ to $6$. Assumption \ref{LCV} is not satisfied in this case. We also vary $\sigma_\gamma$, which controls the magnitude of the deviation of $X$ from the hidden curve $\gamma$. Smaller $\sigma_\gamma$ means that the distribution of $X$ is more concentrated near $\gamma$.

\begin{figure}[tb]\centering
\includegraphics[width=0.85\textwidth]{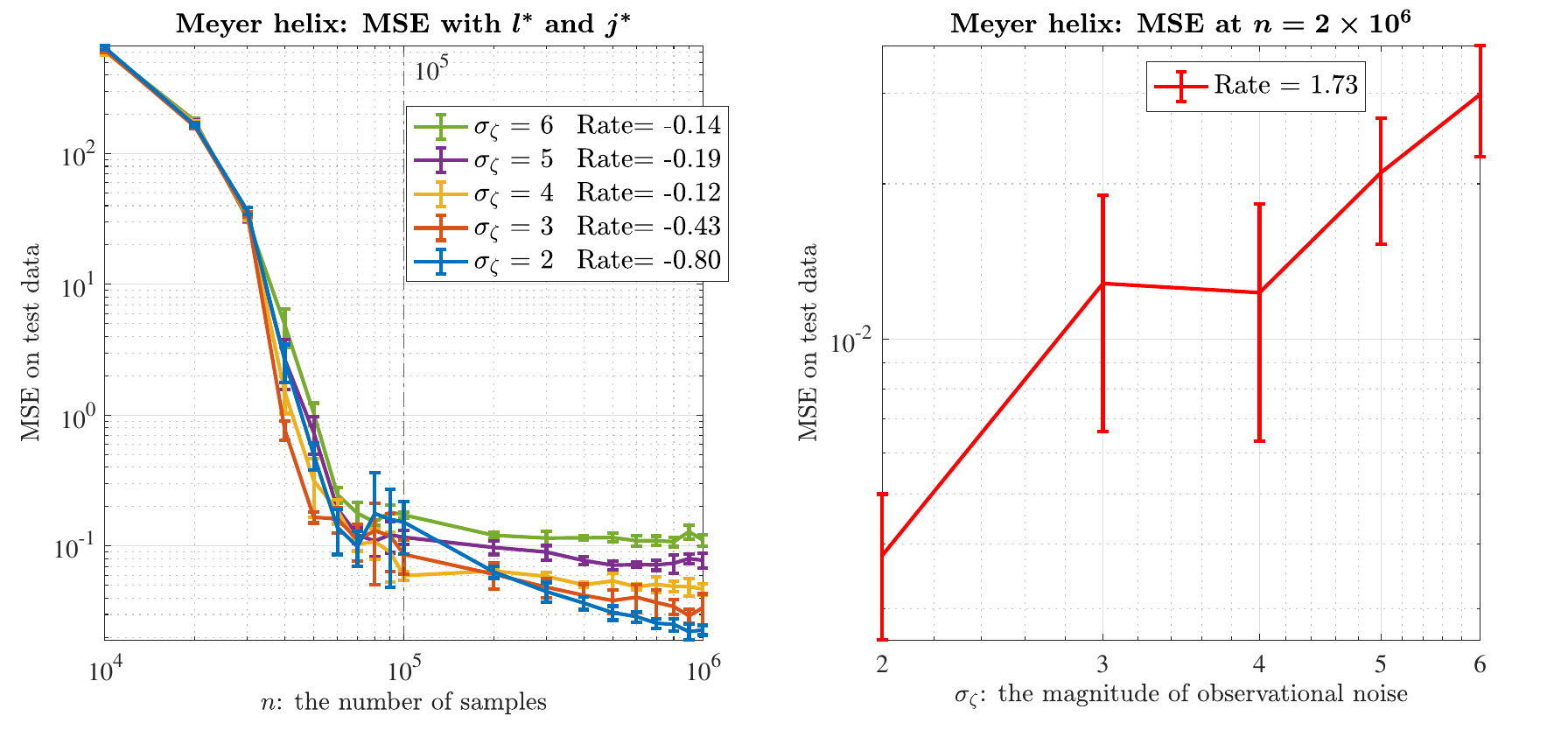}
\caption{Numerical tests in Section \ref{Example 2.4}: the Meyer helix in dimension $d=21$ with $\sigma_\gamma=0.5$. The link is smooth and hence Lipschitz, so the comparison uses $s=1$; the noise level $\sigma_\zeta$ varies from $2$ to $6$. Left: mean squared error; right: mean squared error at $n=2\times10^6$. The reported errors are averages over five independent repetitions.}
\label{Figure: LearningInfo01_wide_Meyer-Helix_D21}
\end{figure}

\begin{figure}[tb]\centering
\includegraphics[width=0.85\textwidth]{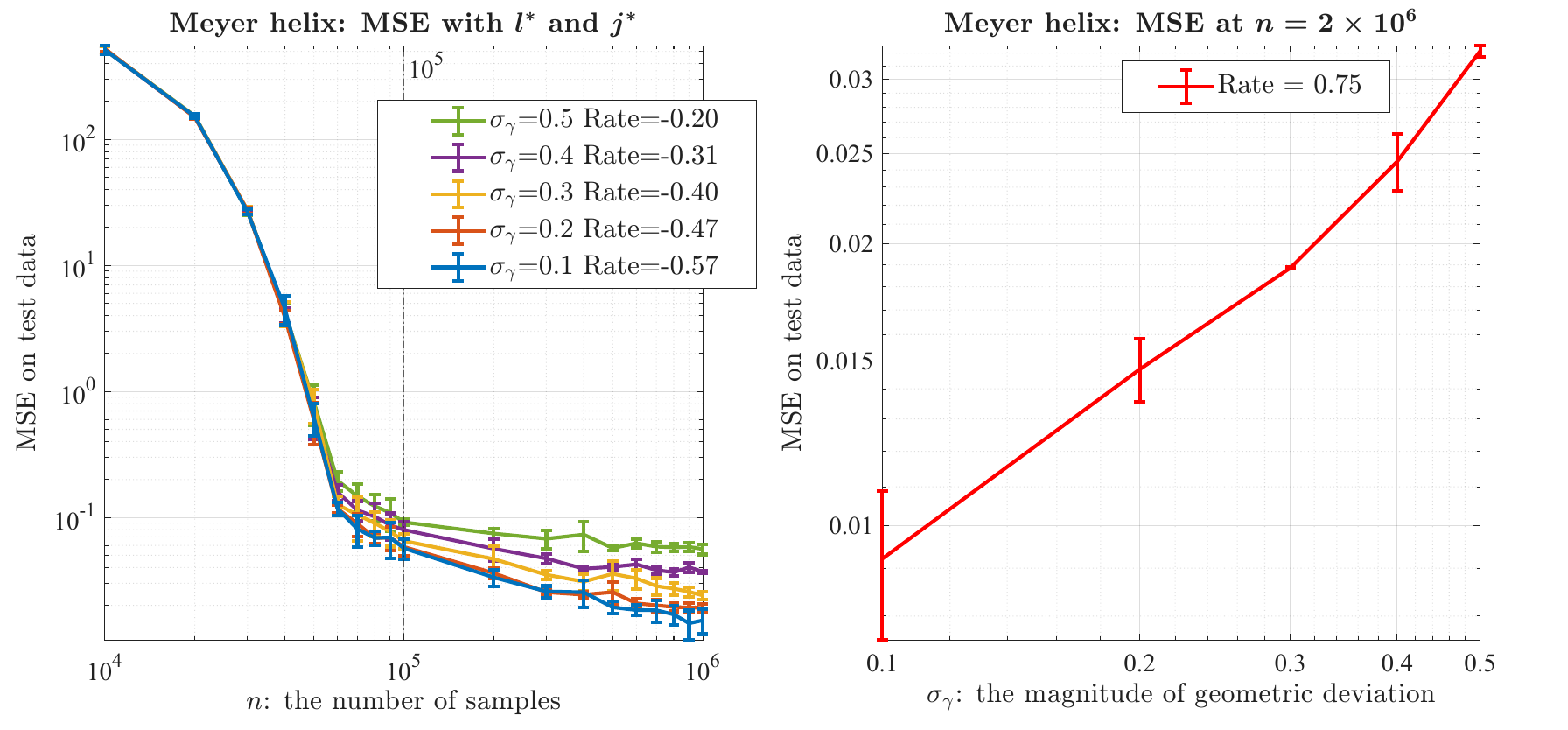}
\caption{Numerical tests in Section \ref{Example 2.4}: the Meyer helix in dimension $d=21$ with $\sigma_\gamma=0.1,0.2,\dots,0.5$. The link is smooth and hence Lipschitz, so the comparison uses $s=1$; the noise level is $\sigma_\zeta=5$. Left: mean squared error. Right: mean squared error at $n=2\times10^6$. The reported errors are averages over five independent repetitions.}
\label{Figure: LearningInfo02_wide_Meyer-Helix_D21}

\end{figure}

Figures \ref{Figure: LearningInfo01_wide_Meyer-Helix_D21} and \ref{Figure: LearningInfo02_wide_Meyer-Helix_D21} show the empirical MSE leveling off in these wide-slice experiments, at a scale comparable to the estimated geometric error.

\section{Conclusion}
We studied the Nonlinear Single-Variable Model, a compositional model $F=f\circ g$ for functions in high dimensions where both $f$ and $g$ are nonlinear, but $g$ has a one-dimensional range and $f$ is therefore a function of one variable.
Using the geometric structure inherent in $g$, and under the coarse-monotonicity, density and lower conditional variance conditions stated above, inverse regression allows us to estimate local level-set geometry, construct local coordinates, and then estimate $F$ by one-dimensional regression, with a one-dimensional rate exponent and polynomial dependence on the ambient dimension for fixed model parameters.

The bounds use a common thin-slice weight and piecewise constant regression. For wide slices the largest principal component supplies the tangent estimate under the small-curvature condition.
The formal risk guarantees are stated for the H\"older--Lipschitz range $s\in[\frac12,1]$, in which the finite-slicing curve-approximation term is compatible with the one-dimensional nonparametric rate. The present estimator uses a first-order local approximation of the curve and its closest-point coordinate; exploiting regularity beyond the Lipschitz level is a natural direction for higher-order geometric estimators.

Extending the method to functions $f$ that are not roughly monotone presents challenges for inverse regression. The pre-image of a set of response values can consist of several disconnected slices, which are difficult to cluster in high dimensions. Related obstructions appear, for example, in work on stochastic-gradient methods for single- and multi-index models \citep{JMLR:v22:20-1288,bietti2023learninggaussianmultiindexmodels}. The sharp statistical and computational tradeoffs in these settings remain incompletely understood.
Another extension of interest is a Nonlinear Multi-Variable Model in which the curve $\gamma$ is replaced by a higher-dimensional manifold $\mathcal{M}$. This setting presents additional challenges because the geometry of the level sets of $F$ becomes substantially more complicated.
Both extensions are under investigation and are left to future work.

An important open direction is to understand how compositional structure affects estimator design and the behavior of general-purpose construction algorithms, such as stochastic gradient descent applied to a suitable loss. This question is especially compelling for compositions with more than two levels and for identifying when such structure can avoid the curse of dimensionality.

\acks{The authors are grateful to A. Lanteri, S. Vigogna and T. Klock (during their stays at Johns Hopkins) for early discussions and experiments on an initial version of this model, and to Fei Lu and Xiong Wang for their feedback on an early version of this manuscript.
We thank the reviewers of the initial version of this manuscript for comments and feedback that improved the presentation of the work.
This research was partially supported by AFOSR FA9550-21-1-0317 and FA9550-23-1-0445.
Prisma Analytics Inc. provided and managed the computing resources for the numerical experiments.}

\appendix

\section{Proofs of the risk bounds}\label{app:proofs}
\subsection{Population slice estimates}\label{app:geometry}
For $x=\gamma(t)+W$ and $t'\in[0,\len_\gamma]$ we use
\begin{equation}\label{eq:reach}
|\ip{W}{\gamma(t')-\gamma(t)}|\leq\frac{\norm{W}}{2\,\reach_\gamma}\norm{\gamma(t')-\gamma(t)}^2,\qquad\norm{\gamma(t')-\gamma(t)}\geq\frac2\pi\min\bigl(|t'-t|,\pi\,\reach_\gamma\bigr),
\end{equation}
by \citet[Theorem 4.8(7),(12)]{federer1959curvature} and \citet[Theorem 1]{BLW19} respectively.
\paragraph{Proof of Proposition~\ref{Prop: Pi_Gamma_X}.}
Let $p=\bbP(Y\in T)$ and let $\mathcal R_f$ be the range of $f$. For any bounded interval $I$, \ref{Omega} gives $\rmdiam(f^{-1}(I))\leq C_f\max(|I|,\omega_f)$. Indeed, if $|I\cap\mathcal R_f|<\omega_f$, enlarge $I$ until the intersection has measure $\omega_f$, using continuity of this measure under enlargement and $|\mathcal R_f|\geq\omega_f$. Apply this estimate with $I=T-z$ for each noise value $z\in\bbR$ and use $\rho_t\leq C_\rho$ to bound $\bbP(f(t)\in T-z)\leq C_\rho C_f|T|$. By independence, for any Borel set $D$,
\[
 \bbP(\zeta\in D,Y\in T)\leq C_\rho C_f|T|\,\bbP(\zeta\in D).
\]
Divide by $p\geq c_\rho C_f'|T|/600$ to obtain the conditional noise tail in (c), with the zero-noise case interpreted directly. Take an independent copy $(t',\zeta')$ drawn from the conditional law given $Y\in T$. Since both responses lie in $T$, the diameter bound in \ref{Omega} gives
$|t-t'|\leq C_f\max(\omega_f,|f(t)-f(t')|)\leq C_f(|T|+|\zeta|+|\zeta'|)$.
Averaging in the copy gives the pointwise assertion in (c). The conditional noise tail now gives (a) and (b), and for $1\leq k\leq4$,
\[
 \bbE[|t-\bar t_h|^k\mid Y\in\Rlh]\lesssim w^k.
\]

\paragraph{Proof of Proposition~\ref{Prop: Thin1}, Corollary~\ref{Coro: Thin2} and Proposition~\ref{Proposition: wide slice}.}
We write the proof for $T=\Rlh$. For a general interval in Proposition~\ref{Prop: Thin1}, use $w=C_f|T|$ and condition on $Y\in T$ throughout. All expectations in this paragraph are with respect to the conditional law. Given $t$, conditioning on $Y$ leaves the law of $W$ unchanged. Put $u=\bar\tau_h$, $D=t-\bar t_h$, and $L=\norm{\gamma''}_\infty$. Taylor's formula gives $\gamma(t)=\gamma(\bar t_h)+Du+E$, with $\norm E\leq LD^2/2$. In particular,
\begin{equation}\label{eq:mean}
 \norm{\mulh-\gamma(\bar t_h)}\leq\tfrac12 L\Var(t\mid Y\in\Rlh)\lesssim Lw^2 .
\end{equation}
The slice covariance is the sum of $\Cov(\gamma(t))$ and $\bbE[\sigma_\gamma(t)^2(I-\tau(t)\tau(t)^\intercal)]$. The curve variance in any direction is at most $Cw^2$ by the unit speed of $\gamma$. Since $\ip{W}{\tau(t)}=0$, $L\norm W\leq c$ (also for a segment, where $L=0$), and $\norm{u-\tau(t)}\leq L|D|$, the Rayleigh quotient in direction $u$ is at most $Cw^2$. This proves Proposition~\ref{Prop: Thin1}(b). On $u^\perp$ the normal covariance is at least $(\sigma_\gamma^2-Cw^2)I$, by the same tangent estimate. The min--max principle gives the thin-slice gap when $w\leq\sigma_\gamma/C$. For $d\geq3$, $\lamlh{\mathrm{mid}}\geq\lambda_{d-1}(\Siglh)\geq\sigma_\gamma^2-Cw^2$, while $\lamlh{1}\leq\bar\sigma_\gamma^2+Cw^2$ and $\lamlh{d}\leq Cw^2$. Taking $w/\sigma_\gamma$ sufficiently small in terms of $\bar\sigma_\gamma/\sigma_\gamma$ gives $\lamlh{\mathrm{mid}}^2>\lamlh{1}\lamlh{d}$, so the population rule selects the smallest eigenvector.

For the direction bound, the tangent--normal block of the curve covariance has norm at most $CLw^3$ by Cauchy--Schwarz and the fourth moment bound. The corresponding block of the normal covariance has norm at most $C\bar\sigma_\gamma^2 Lw$. Dividing their sum by the thin-slice separation gives $\norm{\vlh-u}\lesssim Lw$. The curve projection is Lipschitz in $t$, and
$|\ip{W}{\vlh}|\leq \norm W(\norm{\vlh-u}+L|D|)\lesssim w+|D|$. Proposition~\ref{Prop: Pi_Gamma_X}(c) now gives the conditional tail in Proposition~\ref{Prop: Thin1}(a).

For wide slices the conditional density of $t$ is bounded by $B=C_\rho/\bbP(Y\in\Rlh)$. A density on the line bounded by $B$ has variance at least $1/(12B^2)$, as follows by integrating $\bbP(|t-\bbE t|>r)\geq(1-2Br)_+$. Thus $\Var(t)\gtrsim w^2$. In direction $u$, the standard deviation of the curve projection is at least $\sqrt{\Var(t)}-CLw^2$. On $u^\perp$ the curve variance is at most $CL^2w^4$ and the normal variance at most $\bar\sigma_\gamma^2$. Assumption~\ref{SC}, with $c_1,c_2$ sufficiently small, gives a gap of order $w^2$. Dividing the same tangent--normal block estimate by this separation gives $\norm{\vlh-u}\lesssim L(w+\bar\sigma_\gamma^2/w)\lesssim Lw$ for the largest principal component.

Let $\Omega_0'$ be the set defined as $\Omega_0$ with $w$ in place of $2w$; thus $\Omega_0\subseteq\Omega_0'$, and $x'\in\Omega_0'$ whenever $x\in\Omega_0$ and $|\Pi_\gamma x'-\Pi_\gamma x|\leq w$. The next lemma shows that near every point of $\Omega_0'$ there is a slice satisfying this hypothesis.
\begin{lemma}\label{lem:near}
Assume \ref{zetasub}, \ref{Omega} and \ref{rho}, and suppose that $f$ is continuous on $[0,\len_\gamma]$. For all $t,t'\in[0,\len_\gamma]$:
\begin{enumerate}[label=\textup{(\roman*)}]
\item $|t-t'|\leq C_f\max(|f(t)-f(t')|,\omega_f)$;
\item $|f(t)-f(t')|\leq|t-t'|/C_f'+2(C_f/C_f')\,\omega_f$;
\item for every $x\in\Omega_0'$ there is an index $h_0$ with $\bbP(Y\in R_{l,h_0})\geq c_\rho C_f'|\Rlh|/21$ and $|\Pi_\gamma x-\bar t_{h_0}|\leq Cw$.
\end{enumerate}
\end{lemma}
\begin{proof}
(i) Since $|\mathcal R_f|\geq\omega_f$, the values $f(t),f(t')$ lie in an interval $T_0\subseteq\mathcal R_f$ of length $\max(|f(t)-f(t')|,\omega_f)$, and the upper bound in \ref{Omega} applies. (ii) Let $f(t')<f(t)$ and $T_0:=[f(t'),f(t)]$, and assume $|T_0|\geq\omega_f$. If $u$ is not between $t$ and $t'$ and $f(u)\in T_0$, then $f(u)=f(p)$ for some $p$ between $t$ and $t'$, by continuity of $f$, and $|u-p|\leq C_f\omega_f$ by (i). Hence the smallest interval containing $f^{-1}(T_0)$ has length at most $|t-t'|+2C_f\omega_f$, and the lower bound in \ref{Omega} gives (ii). (iii) Let $t=\Pi_\gamma x$. Since $|\mathcal R_f|\geq|\Rlh|\geq\omega_f$, there is an interval $T\subseteq\mathcal R_f$ of length $|\Rlh|$ containing $f(t)$. By (i), $f^{-1}(T)\subseteq[t-w,t+w]$, and $|f^{-1}(T)|\geq C_f'|\Rlh|$ by \ref{Omega}, so that $\bbP(f(\Pi_\gamma X)\in T)\geq c_\rho C_f'|\Rlh|$ because $x\in\Omega_0'$. Since $\bbP(|\zeta|\leq2\sigma_\zeta)\geq2/3$ and $\sigma_\zeta\leq|\Rlh|$, with probability at least $2c_\rho C_f'|\Rlh|/3$ the response $Y$ lies in $T+[-2\sigma_\zeta,2\sigma_\zeta]\subseteq R$. This interval has length at most $5|\Rlh|$ and meets at most seven response bins, one of which therefore has the stated probability. Every point of this slice has $|f(\Pi_\gamma X)-f(t)|\leq4|\Rlh|+|\zeta|$, and (i) and Proposition~\ref{Prop: Pi_Gamma_X}(c) give the bound on $|t-\bar t_{h_0}|$.
\end{proof}

\subsection{Concentration and estimation of significant vectors}
For a failure-probability parameter $\delta\in(0,1)$ let $\Lambda:=\log(8dl/\delta)$ and
\begin{gather*}
T:=C\bigl(w+C_f\sigma_\zeta\sqrt{\log(n/\delta)}\bigr),\qquad e_\mu:=B_X\sqrt{\frac{2d\Lambda}{n_{loc}}},\qquad e_\Sigma:=11B_X^2\sqrt{\frac{d\Lambda}{n_{loc}}},\\
e_\tau:=3T\sqrt{\frac{\Lambda}{n_{loc}}},\qquad e_{\Sigma\tau}:=10B_XT\sqrt{\frac{d\Lambda}{n_{loc}}}.
\end{gather*}
\begin{lemma}[concentration]\label{lem:conc}
Assume \ref{Xsub}, \ref{zetasub}, \ref{gamma1}, \ref{Omega}, \ref{rho}, $\kappa w\leq1$, $n\geq C$ and $n_{loc}=\lceil c_\rho C_f'|\Rlh|n/400\rceil\geq2d\Lambda$. There is an event $\cS$, measurable with respect to $\{(X_i,Y_i)\}_{i\leq n/4}$, with $\bbP(\cS^c)\leq2\delta+2le^{-n_{loc}/2}$, on which:
\begin{enumerate}[label=\textup{(\roman*)}]
\item every $h$ with $\bbP(Y\in\Rlh)\geq c_\rho C_f'|\Rlh|/21$ is in $\calH_l$, and every $h\in\calH_l$ has $\bbP(Y\in\Rlh)\geq c_\rho C_f'|\Rlh|/600$;
\item for every $h\in\calH_l$, $\norm{\hmulh-\mulh}\leq e_\mu$ and $\norm{\hSiglh-\Siglh}\leq e_\Sigma$;
\item for every $h\in\calH_l$, $|\ip{\hmulh-\mulh}{\bar\tau_h}|\leq e_\tau$ and $\norm{(\hSiglh-\Siglh)\bar\tau_h}\leq e_{\Sigma\tau}$.
\end{enumerate}
\end{lemma}
\begin{proof}
The number of $i\leq n/4$ with $Y_i\in\Rlh$ is binomial with parameters $n/4$ and $\bbP(Y\in\Rlh)$, and (i) follows from the multiplicative Chernoff bounds, with probability at least $1-2le^{-n_{loc}/2}$. Since $R$ is fixed, we label each $Y_i$, $i\leq n/4$, by its bin index and use label zero for $Y_i\notin R$. Conditional on this full vector of labels, the $X_i$ with $Y_i\in\Rlh$ are independent with the law of $X$ given $Y\in\Rlh$, and they are bounded by $B_X$. Hoeffding's inequality for each coordinate of $X_i$, and the matrix Hoeffding inequality \citep[Theorem 1.3]{Tropp12} for $X_iX_i^\intercal$, together with $\hSiglh=\frac{1}{n_{l,h}}\sum_iX_iX_i^\intercal-\hmulh\hmulh^\intercal$, give (ii) with probability at least $1-\delta/2$.

For (iii) let $h$ satisfy $\bbP(Y\in\Rlh)\geq c_\rho C_f'|\Rlh|/600$, and for a point of the slice let $\xi:=\ip{X-\mulh}{\bar\tau_h}$. By Proposition~\ref{Prop: Pi_Gamma_X}(c), \eqref{eq:mean} and $|\ip{W}{\bar\tau_h}|=|\ip{W}{\bar\tau_h-\tau(\Pi_\gamma X)}|\leq|\Pi_\gamma X-\bar t_h|$, we have $|\xi|\leq C(w+C_f|\zeta|)$. Let $z:=\sigma_\zeta\sqrt{2\log(n/\delta)}$, so that $|\zeta_i|\leq z$ for all $i\leq n/4$ with probability at least $1-\delta/2$, and let $\xi^\circ:=\xi\,\ind\{|\zeta|\leq z\}$, which is bounded by $T$. By Proposition~\ref{Prop: Pi_Gamma_X}(c), $\bbE[|\xi|\ind\{|\zeta|>z\}\mid Y\in\Rlh]\leq CT\delta/n\leq T\sqrt{\Lambda/n_{loc}}$ for $n\geq C$. Hoeffding's inequality applied to the variables $\xi_i^\circ$, and to each coordinate of the vectors $(X_i-\mulh)\xi^\circ_i$, which is bounded by $2B_XT$, gives with probability at least $1-\delta/2$, for every such $h$ in $\calH_l$, deviations from the means at most $T\sqrt{2\Lambda/n_{loc}}$ and $B_XT\sqrt{8d\Lambda/n_{loc}}$ respectively. On the event that $|\zeta_i|\leq z$ for all $i\leq n/4$ we have $\xi_i^\circ=\xi_i$, and (iii) follows from $\bbE[(X-\mulh)\xi\mid Y\in\Rlh]=\Siglh\bar\tau_h$ and $(\hSiglh-\Siglh)\bar\tau_h=\frac{1}{n_{l,h}}\sum_i(X_i-\mulh)\xi_i-\Siglh\bar\tau_h-(\hmulh-\mulh)\ip{\hmulh-\mulh}{\bar\tau_h}$.
\end{proof}

\paragraph{Significant vectors.} On $\cS$, Lemma~\ref{lem:conc}(i) guarantees the population mass required for the eigengap bound on each retained slice $h$. Let $k=d$ in Corollary~\ref{Coro: Thin2} and $k=1$ in Proposition~\ref{Proposition: wide slice}. Write $g$ for the corresponding eigengap of $\Siglh$, $\vlh$ for its population eigenvector of index $k$, and $\widehat v_{l,h,k}$ for the unit eigenvector of $\hSiglh$ with the same index. If $e_\Sigma\leq g/2$, the residual form of the Davis--Kahan bound for an extremal eigenvector \citep[Theorem VII.3.1]{Bhatia97} gives $\min_{\varepsilon\in\{-1,1\}}\norm{\widehat v_{l,h,k}-\varepsilon v_{l,h}}\lesssim\norm{(\hSiglh-\Siglh)v_{l,h}}/g$, and $\norm{(\hSiglh-\Siglh)v_{l,h}}\leq\min\{e_\Sigma,\ e_{\Sigma\tau}+e_\Sigma\norm{v_{l,h}-\bar\tau_h}\}$. We write $\angle(u,v)$ for the acute or right angle between the lines spanned by nonzero vectors $u,v$, taking it to be zero for parallel lines.
\begin{itemize}
\item \emph{Thin slices.} Under the hypotheses of Corollary~\ref{Coro: Thin2}, take $w\leq\sigma_\gamma/C$ and $e_\Sigma\leq\sigma_\gamma^2/C$. $\lambda_d(\Siglh)\lesssim w^2$ by Proposition~\ref{Prop: Thin1}(b), $\lambda_{d-1}(\Siglh)\geq\sigma_\gamma^2-Cw^2$ by the proof of Corollary~\ref{Coro: Thin2}, and $\lambda_1(\Siglh)\leq\bar\sigma_\gamma^2+Cw^2$. By Weyl's inequality, $\lambda_d(\hSiglh)\leq\sigma_\gamma^2/C$, $\lambda_{d-1}(\hSiglh)\geq\sigma_\gamma^2/2$ and $\lambda_1(\hSiglh)\leq2\bar\sigma_\gamma^2$, so that $\widehat G_{l,h}>0$ for $d\geq3$, and Algorithm~\ref{Alg: NVM} selects $\hvlh=\widehat v_{l,h,d}$ and the first line of \eqref{Equation: distance function}. Applying the residual bound with $k=d$ and using Corollary~\ref{Coro: Thin2} gives
\begin{equation}\label{eq:thetathin}
\angle(\hvlh,\bar\tau_h)\leq\theta:=C\bigl(\kappa w+e_{\Sigma\tau}/\sigma_\gamma^2\bigr).
\end{equation}
\item \emph{Wide slices.} Under the hypotheses of Proposition~\ref{Proposition: wide slice}, the wide-slice rule prescribes $\hvlh=\widehat v_{l,h,1}$. If $e_\Sigma\leq w^2/C$, the residual bound with $k=1$ and the population direction estimate of the proposition give
\begin{equation}\label{eq:thetawide}
\angle(\hvlh,\bar\tau_h)\leq\theta:=C\bigl(\kappa w+e_\Sigma/w^2\bigr).
\end{equation}
\end{itemize}

\subsection{Assignment to a nearby retained slice}
Let $W_{\max}:=\min(c\,\reach_\gamma,2B_X)$, which bounds $\norm{W}$.
\begin{lemma}[slice assignment]\label{lem:assign}
Under the hypotheses of Lemma~\ref{lem:conc}, assume also that $f$ is continuous on $[0,\len_\gamma]$. Let the event $\cS$ of Lemma \ref{lem:conc} hold, let $\angle(\hvlh,\bar\tau_h)\leq\theta\leq(1-c)/8$ for every $h\in\calH_l$, and let $q:=q_l$ if the first line of \eqref{Equation: distance function} is used for every $h\in\calH_l$, and $q:=1$ if the second line is used for every $h\in\calH_l$. Assume $q_l\leq1$ and
\begin{equation}\label{eq:H}
\kappa w\leq1/C,\qquad \min(e_\tau,e_\mu)\leq w,\qquad\theta\,(W_{\max}+e_\mu)\lesssim w,\qquad q\,e_\mu(e_\mu+W_{\max})\leq w^2 .
\end{equation}
Then every $x\in\Omega_0'$ satisfies: \textup{(i)} $|\Pi_\gamma x-\bar t_{\hhx}|\leq Cw$; \textup{(ii)} $|\hhx-\hxp|\leq C$; \textup{(iii)} $\ip{\widehat v_{l,\hhx}}{x}\in I^{(l,\hhx)}$.
\end{lemma}
\begin{proof}
Let $x=\gamma(t)+W\in\Omega_0'$. By Lemma \ref{lem:near}(iii) and Lemma \ref{lem:conc}(i) there is $h_0\in\calH_l$ with $|t-\bar t_{h_0}|\leq Cw$. For $h\in\calH_l$ let $D_h:=|t-\bar t_h|$, $y_h:=\norm{\gamma(t)-\mulh}$, $a_h:=\ip{x-\hmulh}{\hvlh}$ and $b_h:=\norm{x-\hmulh}^2-\norm{W}^2$. We claim that, with $E:=e_\mu(e_\mu+W_{\max})$,
\begin{gather}
\tfrac2\pi\min(D_h,\pi\,\reach_\gamma)-w\leq y_h\leq D_h+w,\label{eq:y}\\
\tfrac12(1-c)y_h^2-C(w^2+E)\leq b_h\leq3y_h^2+C(w^2+E),\label{eq:b}\\
(1-c-\kappa D_h/2-\theta)D_h-Cw\leq|a_h|\leq3D_h+Cw.\label{eq:a}
\end{gather}
Indeed, \eqref{eq:y} follows from \eqref{eq:mean} and \eqref{eq:reach}. For \eqref{eq:b} write $b_h=\norm{\gamma(t)-\hmulh}^2+2\ip{W}{\gamma(t)-\mulh}+2\ip{W}{\mulh-\hmulh}$; the first term lies between $y_h^2-2y_he_\mu$ and $(y_h+e_\mu)^2$, the third is at most $2W_{\max}e_\mu$ in absolute value, and, since $\mulh=\bbE[\gamma(\Pi_\gamma X)\mid Y\in\Rlh]$, \eqref{eq:reach} and Proposition~\ref{Prop: Pi_Gamma_X}(b) bound the second by $c\,(y_h^2+Cw^2)$. For \eqref{eq:a}, orient $\hvlh$ so that $\norm{\hvlh-\bar\tau_h}\leq\theta$ and write
\[ a_h=\ip{\gamma(t)-\gamma(\bar t_h)}{\bar\tau_h}+\ip{W}{\bar\tau_h-\tau(t)}+\ip{\gamma(\bar t_h)-\mulh}{\bar\tau_h}+\ip{\mulh-\hmulh}{\bar\tau_h}+\ip{x-\hmulh}{\hvlh-\bar\tau_h}. \]
The first term has absolute value between $D_h-\kappa D_h^2/2$ and $D_h$; the others are at most $cD_h$, $w$, $\min(e_\tau,e_\mu)\leq w$ by Lemma~\ref{lem:conc}(ii)--(iii) and $\theta(D_h+W_{\max}+w+e_\mu)\leq\theta D_h+Cw$ respectively.

Let $h\in\calH_l$ with $D_h\geq C_*w$, where $C_*$ is a sufficiently large constant. By \eqref{eq:y}--\eqref{eq:a}, $a_{h_0}^2\leq Cw^2$ and $b_{h_0}\leq C(w^2+E)$, and $qE\leq w^2$. If the first line of \eqref{Equation: distance function} is used, then $\hdist(x,h)-\hdist(x,h_0)=a_h^2-a_{h_0}^2+q(b_h-b_{h_0})$. When $D_h\leq(1-c)\reach_\gamma/4$, \eqref{eq:a} gives $|a_h|\geq\frac34(1-c)D_h-Cw$ and \eqref{eq:b} gives $b_h\geq-C(w^2+E)$, so the difference is at least $(\frac34(1-c)C_*-C)^2w^2-Cw^2>0$. When $D_h>(1-c)\reach_\gamma/4$, \eqref{eq:y} and \eqref{eq:b} give $b_h\geq(1-c)^3\reach_\gamma^2/128-C(w^2+E)$, so the difference is at least $q_l(1-c)^3\reach_\gamma^2/128-Cw^2>0$, because $q_l=C(w/\reach_\gamma)^2$ with $C$ sufficiently large. If the second line of \eqref{Equation: distance function} is used, its weights lie in $[0,1]$, so $\hdist(x,h)-\hdist(x,h_0)\geq b_h-b_{h_0}-a_{h_0}^2\geq\frac12(1-c)y_h^2-Cw^2>0$ by \eqref{eq:y}. In all cases $h\neq\hhx$, which proves (i), and (iii) follows from (i), \eqref{eq:a} and the choice of $r_l$. For (ii), by Proposition~\ref{Prop: Pi_Gamma_X}(b),(c) and Markov's inequality the slice $\hhx$ contains a point with arclength $t'$ and noise $\zeta'$ such that $|t'-\bar t_{\hhx}|\leq Cw$ and $|\zeta'|\leq C\sigma_\zeta$; then $f(t')+\zeta'\in R_{l,\hhx}$ and $|f(t)-f(t')|\leq C|\Rlh|$ by Lemma \ref{lem:near}(ii).
\end{proof}

\subsection{Risk of local regression}
Let $\mathcal G$ be the $\sigma$-algebra generated by the observations $\{(X_i,Y_i)\}_{i\leq n/4}$. Condition on $\mathcal G$, which determines $\calH_l$, the map $x\mapsto\hhx$, the vectors $\hvlh$ and the bins $I^{(l,h)}_{j,k}$. Let $N:=n/2$ be the number of regression observations, $A_{l,h}=\{x:|\hhx-h|\leq1\}$, and
\[ \mathcal D_{h,k}:=\{x\in A_{l,h}:\ip{\hvlh}{x}\in I^{(l,h)}_{j,k}\}\quad(h\in\calH_l,\ k\leq j),\qquad\mathcal Q:=\{x:\ip{\widehat v_{l,\hhx}}{x}\in I^{(l,\hhx)}\}. \]
When $\calH_l=\emptyset$, set $\mathcal Q=\emptyset$, in accordance with the zero estimate in Algorithm~\ref{Alg: NVM}.
\begin{lemma}[accepted-region domination]\label{Lem: accepted_region_domination}
Condition on $\mathcal G$. For every collection of nonnegative measurable functions $G_h$, $h\in\calH_l$,
\begingroup
\[
\sum_{h\in\calH_l}\bbE\sbracket{G_h(X)\II\{\widehat h(X)=h\}}
\leq
\sum_{h\in\calH_l}\bbE\sbracket{G_h(X)\II\{X\in A_{l,h}\}}.
\]
\endgroup
Moreover, $\sum_{h\in\calH_l}\II\{X\in A_{l,h}\}\leq3$ pointwise.
\end{lemma}
Indeed, if $\widehat h(x)=h$, then $|\widehat h(x)-h|=0$, hence $x\in A_{l,h}$. The displayed inequality follows pointwise from $\II\{\widehat h(X)=h\}\leq\II\{X\in A_{l,h}\}$. For fixed $x$, the condition $x\in A_{l,h}$ is equivalent to $h\in\{\widehat h(x)-1,\widehat h(x),\widehat h(x)+1\}$, so at most three indices contribute.

\begin{lemma}[local regression]\label{lem:risk}
Let $m=0$ and $\mathcal K^{(l,h)}_j=\{k:n^{(l,h)}_{j,k}\geq1\}$.
\begin{enumerate}[label=\textup{(\alph*)}]
\item With $V:=\sum_{h\in\calH_l}\sum_{k\leq j}\bbP(X\in\mathcal D_{h,k}\mid\mathcal G)\Var(F(X)\mid X\in\mathcal D_{h,k},\mathcal G)$,
\[ \bbE\bigl[|\hF(X)-F(X)|^2\mid\mathcal G\bigr]\leq4V+(2\sigma_\zeta^2+4M^2)\frac{lj}{N+1}+4M^2\,\bbP(X\notin\mathcal Q\mid\mathcal G). \]
\item Assume \ref{gamma1}. Suppose that for every $h\in\calH_l$ there is an interval $J_h\subseteq[0,\len_\gamma]$ such that $X\in A_{l,h}\cap\Omega_0$ implies $\Pi_\gamma X\in J_h$ and such that $\angle(\hvlh,\tau(t))\leq\beta\leq\frac18$ for $t\in J_h$, and that each $t$ belongs to at most $K_0$ of these intervals. Then
\[ V\leq24\Holder{f}{s}^2\Bigl(\frac{2r_l}{j}\Bigr)^{2s}+8K_0\Holder{f}{s}^2(2\bar\sigma_\gamma\beta)^{2s}+12M^2\,\bbP(X\notin\Omega_0). \]
When $\beta=0$, the projected normal displacement vanishes on each $J_h$, and we take the second term to be zero even if $\bar\sigma_\gamma$ is infinite.
\end{enumerate}
\end{lemma}
\begin{proof}
(a) Let $x\in\mathcal Q$ with $\hhx=h$ and $\ip{\hvlh}{x}\in I^{(l,h)}_{j,k}$, so that $x\in\mathcal D_{h,k}$, and let $N_{h,k}$ be the number of regression observations in $\mathcal D_{h,k}$, which is binomial with parameters $N$ and $p:=\bbP(X\in\mathcal D_{h,k}\mid\mathcal G)$. Cells of zero probability have zero contribution. If $N_{h,k}\geq1$, then $\hF(x)$ is the truncation of the average of the $Y_i=F(X_i)+\zeta_i$ over these observations, and truncation does not increase the error. With $\bar F:=\bbE[F(X)\mid X\in\mathcal D_{h,k},\mathcal G]$, and writing $\bbE_\zeta$ for expectation conditional on $\mathcal G$ and the regression predictors,
\[ \bbE_\zeta|\hF(x)-F(x)|^2\leq2(F(x)-\bar F)^2+\frac{2}{N_{h,k}}\sum_{i:X_i\in\mathcal D_{h,k}}(F(X_i)-\bar F)^2+\frac{\sigma_\zeta^2}{N_{h,k}} . \]
Conditional on $\mathcal G$ and $N_{h,k}=r\geq1$, the $r$ predictors in the cell have the product law of $r$ draws from $X\mid X\in\mathcal D_{h,k},\mathcal G$. The queries assigned to this cell form a subset of $\mathcal D_{h,k}$ of probability at most $p$. Integration over this subset under the independent test law and over the $r$ predictors bounds the first two terms by $4p\Var(F(X)\mid X\in\mathcal D_{h,k},\mathcal G)$ and the noise term by $p\sigma_\zeta^2/r$. Averaging the latter bound over positive cell counts gives
\[
p\sigma_\zeta^2\sum_{r=1}^N\frac{\bbP(N_{h,k}=r\mid\mathcal G)}{r}
\leq 2p\sigma_\zeta^2\bbE\!\left[\frac{1}{N_{h,k}+1}\,\middle|\,\mathcal G\right]
=\frac{2\sigma_\zeta^2}{N+1}\bigl(1-(1-p)^{N+1}\bigr).
\]
If $N_{h,k}=0$ or $x\notin\mathcal Q$, the error is at most $4M^2$, and $p\,(1-p)^N\leq1/(N+1)$. Sum over the at most $lj$ pairs $(h,k)$.

(b) Fix $h$. If $J_h$ is empty, then $\mathcal D_{h,k}\cap\Omega_0$ is empty for every $k$, and we use $c_k=0$ below. Otherwise let $\phi(u):=\ip{\hvlh}{\gamma(u)}$. On $J_h$ its continuous derivative satisfies $|\phi'(u)|=|\ip{\hvlh}{\tau(u)}|\geq\cos\beta\geq\frac12$ and hence has constant sign. The inverse on $\phi(J_h)$ is therefore $2$-Lipschitz for either sign of the fitted vector. Extend the inverse to $\bbR$ by its endpoint limits outside $\phi(J_h)$, retaining the same Lipschitz constant and taking values in $\overline{J_h}\subseteq[0,\len_\gamma]$. Let $z_k$ be the centre of $I^{(l,h)}_{j,k}$. If $X=\gamma(t)+W\in\mathcal D_{h,k}\cap\Omega_0$, then $t\in J_h$ and $\ip{\hvlh}{X}=\phi(t)+\ip{\hvlh}{W}$ is within $r_l/j$ of $z_k$, so $|t-\phi^{-1}(z_k)|\leq2(r_l/j+|\ip{\hvlh}{W}|)$ and, with $c_k:=(f(\phi^{-1}(z_k))\vee(-M))\wedge M$ and since $|F(X)|\leq M$, $|F(X)-c_k|^2\leq|F(X)-f(\phi^{-1}(z_k))|^2\leq8\Holder{f}{s}^2((r_l/j)^{2s}+|\ip{\hvlh}{W}|^{2s})$. On $\mathcal D_{h,k}\setminus\Omega_0$, $|F(X)-c_k|^2\leq4M^2$. Since the variance is the smallest mean squared deviation from a constant, since the sets $\mathcal D_{h,k}$, $k\leq j$, are disjoint subsets of $A_{l,h}$, and since $\sum_h\ind_{A_{l,h}}\leq3$ (Lemma~\ref{Lem: accepted_region_domination}), the first and third terms of the bound follow. For the second, $\hvlh$ is determined by $\mathcal G$ and $X$ is independent of $\mathcal G$, so by \ref{gamma1} and Jensen's inequality $\bbE[|\ip{\hvlh}{W}|^{2s}\mid t,\mathcal G]\leq(\bar\sigma_\gamma\beta)^{2s}$ for almost every $t\in J_h$, and each $t$ belongs to at most $K_0$ intervals $J_h$.
\end{proof}

\subsection{Proofs of the main theorems}
Since $|\hF|,|F|\leq M\leq2\Linfty{f}$, the three risk bounds hold if $\Omega_0=\emptyset$. Assume $\Omega_0\neq\emptyset$ and set $l=l^*$. By \ref{Omega}, $C_f'|\mathcal R_f|\leq\len_\gamma\leq C_f|\mathcal R_f|$ and $|\mathcal R_f|\leq2C_YR_0\lesssim|\mathcal R_f|$. Together with $1/C_\rho\leq\len_\gamma\leq1/c_\rho$, these give $\len_\gamma\leq lw$ and $n_{loc}=\lceil c_\rho C_f'|\Rlh|n/400\rceil\asymp 1+n/l$.

In Theorems~\ref{Thm: NLSIM} and~\ref{Thm: noiselessNLSIM}, $dR_0^2\geq\operatorname{tr}\operatorname{Cov}(X)\geq(d-1)\sigma_\gamma^2$ by \ref{Xsub} and \ref{gamma1}, so $R_0\gtrsim\sigma_\gamma$ and \eqref{e:constantsinThm} gives $C_{\gamma,f}\geq d^4(R_0/\sigma_\gamma)^8\gtrsim1$. The prescribed choices of $l$ then give $l\lesssim n$, using $c_\rho\len_\gamma\leq1$ in the noiseless case. The lower bound on $l$ in the noisy case and the sample-size condition in the noiseless case give $w\lesssim\sigma_\gamma\lesssim B_X$.

In Theorem~\ref{Thm: NLSIM_wide}, write $S=\max(\sigma_\zeta,\omega_f)$. The choice $l=\lceil C_YR_0/S\rceil$ and $|\Rlh|\geq S$ give $C_fS\leq w\leq2C_fS$, so $\kappa w\leq2c_2$ by \ref{SC}. Since $\Omega_0\neq\emptyset$, $\rho_t\geq c_\rho$ on an arc of length $\gtrsim w$. On this arc $\gamma(u)=\bbE[X\mid t=u]$ lies in the closed ball of radius $B_X$ almost everywhere and hence everywhere by continuity. Applying \eqref{eq:reach} to a subarc of length comparable to $w$ gives $B_X\gtrsim w$. The sample-size condition now yields $n/\log(dn)\gtrsim l d(B_X/w)^6\gtrsim l$. Thus $n_{loc}\asymp n/l$ in all three regimes.

\paragraph{Theorems~\ref{Thm: NLSIM} and~\ref{Thm: NLSIM_wide}.} Let $\delta:=1/n$, so that $\Lambda\lesssim\log(dn)$ and, since $C_f\sigma_\zeta\leq w$, $T\lesssim w\sqrt{\log n}$. In Theorem~\ref{Thm: NLSIM}, $l^*\gtrsim C_f\len_\gamma/(C_f'\sigma_\gamma)$ gives $w\leq\sigma_\gamma/C$, hence $\kappa w\leq1/C$ and $q_l\leq1$, because $\sigma_\gamma\leq\reach_\gamma$ by \ref{gamma1}; and $l^*\leq C^{-1}C_{\gamma,f}^{-1}n\log^{-2}n$, together with $C_{\gamma,f}\geq d^4(R_0/\sigma_\gamma)^8$, $R_0\gtrsim\sigma_\gamma$ and $\log(dn)\leq(1+\log d)\log n$, gives
\begin{equation}\label{eq:nthin}
n_{loc}\geq Cd\,(B_X/\sigma_\gamma)^4\log^2(dn).
\end{equation}
In Theorem~\ref{Thm: NLSIM_wide}, $w\asymp C_f S$, \ref{SC} gives $\bar\sigma_\gamma\leq w/C$ and $\kappa w\leq1/C$, and the condition on $n$ gives
\begin{equation}\label{eq:nwide}
n_{loc}\geq Cd\,(B_X/w)^6\log(dn).
\end{equation}
By the definitions of $e_\mu,e_\Sigma,e_\tau,e_{\Sigma\tau}$ and the bounds above, \eqref{eq:nthin} gives $e_\mu\leq\sigma_\gamma/C$, $e_\Sigma\leq\sigma_\gamma^2/C$, $e_\tau\leq w$ and $e_{\Sigma\tau}\leq w\sigma_\gamma^2/(CB_X)$, and \eqref{eq:nwide} gives $\min(e_\tau,e_\mu)\leq e_\mu\leq w^3/(CB_X^2)\leq w$ and $e_\Sigma\leq w^3/(CB_X)\leq w^2/C$. In both cases the conditions of the paragraph on significant vectors hold, with $\theta\lesssim\kappa w+w/(CB_X)$ in \eqref{eq:thetathin} and \eqref{eq:thetawide} respectively, and so do $\theta\leq(1-c)/8$ and \eqref{eq:H}, since $W_{\max}\leq\min(c\,\reach_\gamma,2B_X)$. In Theorem~\ref{Thm: NLSIM} the first line of \eqref{Equation: distance function} is used for every $h\in\calH_l$, as shown in the same paragraph. Lemma \ref{lem:assign} therefore applies on $\cS$.

For $h\in\calH_l$ enlarge $\Rlh$ by $C|\Rlh|$ on each side to obtain $R_h^+$, and let $J_h$ be the smallest interval containing $f^{-1}(R_h^+)$. By Lemma \ref{lem:assign}(ii) and the definition of $A_{l,h}$, $X\in A_{l,h}\cap\Omega_0$ implies $F(X)\in R_h^+$, hence $\Pi_\gamma X\in J_h$. By \ref{Omega} and Lemma \ref{lem:near}(ii), $|J_h|\leq Cw$ and each $t$ belongs to at most $C$ of the intervals $J_h$; by Proposition~\ref{Prop: Pi_Gamma_X}(b),(c), $J_h$ contains a point within $Cw$ of $\bar t_h$, so that $\angle(\hvlh,\tau(t))\leq\theta+\kappa|t-\bar t_h|\leq\theta+C\kappa w\leq\frac18$ on $J_h$. By Lemma \ref{lem:assign}(iii), $\bbP(X\notin\mathcal Q\mid\mathcal G)\leq\bbP(X\notin\Omega_0)$. Write $\theta_e:=e_{\Sigma\tau}/\sigma_\gamma^2$ in Theorem~\ref{Thm: NLSIM} and $\theta_e:=e_\Sigma/w^2$ in Theorem~\ref{Thm: NLSIM_wide}. Lemma \ref{lem:risk}, the bound $\bbP(\cS^c)\leq2\delta+2le^{-n_{loc}/2}\lesssim1/n$ and $r_l=Cw$ give, for every $j\geq1$,
\begin{equation}\label{eq:master}\begin{aligned}
\bbE|\hF(X)-F(X)|^2 &\lesssim\Holder{f}{s}^2\Bigl(\frac wj\Bigr)^{2s}+\Holder{f}{s}^2\bigl(\bar\sigma_\gamma\kappa w\bigr)^{2s}+\Holder{f}{s}^2\bigl(\bar\sigma_\gamma\theta_e\bigr)^{2s}\\
&\quad+\bigl(\sigma_\zeta^2+\Linfty{f}^2\bigr)\frac{lj}{n}+\Linfty{f}^2\,\bbP(X\notin\Omega_0),
\end{aligned}\end{equation}

In Theorem~\ref{Thm: NLSIM}, take $l=l^*$ and $j=j^*$. The second and last terms of \eqref{eq:master} give $C_2$ up to a constant. Since $w=2C_fC_YR_0/l$ and $lj\asymp M^*n^{1/(2s+1)}$, the choice of $M^*$ gives $\Holder{f}{s}^2(w/j)^{2s}\asymp\sigma_\zeta^2lj/n$. Using $C_fC_YR_0\lesssim (C_f/C_f')\len_\gamma$, we bound the sum of the first and fourth terms by a constant times the first term of $C_1$ times $n^{-2s/(2s+1)}$. The third term is at most the second term of $C_1$ times $n^{-2s/(2s+1)}\log n$, because $\log^{2s}(dn)\,n^{-s}\leq2(\log ed)^2n^{-2s/(2s+1)}\log n$ for $s\geq\frac12$ and
\[ \bar\sigma_\gamma\theta_e\lesssim\frac{B_Xw}{\sigma_\gamma}\log(dn)\sqrt{\frac{d}{n_{loc}}}\lesssim\frac{C_XR_0\,C_fC_YR_0\,d}{\sigma_\gamma}\,\frac{\log(dn)}{\sqrt n}. \] In Theorem~\ref{Thm: NLSIM_wide}, the first term of \eqref{eq:master} is at most the second by the choice of $j^*$, the second is the saturation term, and the third and fourth are bounded by it by the additional condition, since $\bar\sigma_\gamma\theta_e\lesssim\bar\sigma_\gamma B_X^2w^{-2}\sqrt{d\,l^*\log(dn)/n}$.

\paragraph{Theorem~\ref{Thm: noiselessNLSIM}.} Let $\delta:=n^{-2}$. Here $T=Cw$, the condition on $n$ gives $w\leq\sigma_\gamma/C$, and the choice of $l^*$ gives, as for \eqref{eq:nthin}, $n_{loc}\geq Cd\,(B_X/\sigma_\gamma)^4\log(dn)$ and $\bbP(\cS^c)\lesssim n^{-2}$. As for Theorem~\ref{Thm: NLSIM}, \eqref{eq:thetathin} holds, the first line of \eqref{Equation: distance function} is used for every $h\in\calH_l$, and Lemma \ref{lem:assign} applies on $\cS$; let $K$ be the constant in its part (ii). In Step 3.b the condition is $|\widehat h(X_i)-h|\leq2K+1$, so that $A_{l,h}=\{x:|\hhx-h|\leq2K+1\}$. Since $j^*=1$, for each $h\in\calH_l$ there is one cell $\mathcal D_h:=\mathcal D_{h,1}$, and $\hF(x)$ is the truncated average of the $Y_i=F(X_i)$ over the regression observations in $\mathcal D_{\hhx}$, if there are any.

Work on $\cS$ and let $x\in\Omega_0$ with $\hhx=h$ and $t=\Pi_\gamma x$. As in the proof of Lemma \ref{lem:near}(iii), there is an interval $T\subseteq\mathcal R_f$ of length $|\Rlh|$ containing $f(t)$, with $f^{-1}(T)\subseteq[t-w,t+w]$ and $\bbP(\Pi_\gamma X\in f^{-1}(T))\geq c_\rho C_f'|\Rlh|$. Since $\omega_f=0$, \ref{Omega} makes $f$ injective, and the density of $\Pi_\gamma X$ gives zero probability to the preimage of each response-bin endpoint. Thus almost every $x'$ under the law of $X$ with $\Pi_\gamma x'\in f^{-1}(T)$ is in $\Omega_0'$ and has $|h'_{x'}-\hxp|\leq1$, hence $x'\in A_{l,h}$ by Lemma \ref{lem:assign}(ii), and $\ip{\hvlh}{x'}\in I^{(l,h)}$ by Lemma \ref{lem:assign}(i), \eqref{eq:a} and the choice of $r_l$. Thus $\bbP(X\in\mathcal D_h\mid\mathcal G)\geq c_\rho C_f'|\Rlh|$, and the number $N_h$ of regression observations in $\mathcal D_h$ is zero with probability at most $e^{-n_{loc}}\lesssim n^{-2}$; moreover $x\in\mathcal D_h$ by Lemma \ref{lem:assign}(iii). If $X_i\in\mathcal D_h\cap\Omega_0'$, then $|h'_{X_i}-\hxp|\leq4K+1$ by Lemma \ref{lem:assign}(ii), hence $|\Pi_\gamma X_i-t|\leq Cw$ by Lemma \ref{lem:near}(i) and $|F(X_i)-F(x)|\leq\Holder{f}{s}(Cw)^s$; otherwise $|F(X_i)-F(x)|\leq2\Linfty{f}$. Let $B_h$ be the number of regression observations in $\mathcal D_h\setminus\Omega_0'$. For $N_h\geq1$,
\begin{align*}
|\hF(x)-F(x)|^2&\leq2\Holder{f}{s}^2(Cw)^{2s}+8\Linfty{f}^2\frac{B_h}{N_h},\\
\bbE\Bigl[\ind\{N_h\geq1\}\frac{B_h}{N_h}\Bigm|\mathcal G\Bigr]&\leq\frac{\bbP(X\in\mathcal D_h\setminus\Omega_0'\mid\mathcal G)}{\bbP(X\in\mathcal D_h\mid\mathcal G)} .
\end{align*}
The queries $x\in\Omega_0$ with $\hhx=h$ have probability at most $\bbP(X\in\mathcal D_h\mid\mathcal G)$; summing over $h$ and using $\sum_h\ind_{A_{l,h}}\leq4K+3$ and $\Omega_0\subseteq\Omega_0'$ gives
\[ \bbE|\hF(X)-F(X)|^2\lesssim\Holder{f}{s}^2w^{2s}+\Linfty{f}^2\bigl(\bbP(X\notin\Omega_0)+n^{-2}\bigr). \]
Finally $w\lesssim\frac{C_fC_{\gamma,f}}{c_\rho C_f'}\frac{\log^{3/2}n}{n}$ by the choice of $l^*$, and $\Linfty{f}^2n^{-2}\lesssim\Holder{f}{s}^2w^{2s}$ because $\Linfty{f}\lesssim|\mathcal R_f|\leq\Holder{f}{s}\len_\gamma^s$ and $\len_\gamma\leq l^*w\leq Cnw$.
With $\omega_f=0$, Lemma~\ref{lem:near}(i)--(ii) gives $|t-u|/C_f\leq |f(t)-f(u)|\leq |t-u|/C_f'$ on $[0,\len_\gamma]$. Extending $f$ by its endpoint values outside this interval gives a bounded Lipschitz link with seminorm at most $1/C_f'$ and the same regression function. Since $l^*$, $j^*$ and the window $a$ are independent of $s$, applying the bound with $s=1$ gives the second conclusion for the same estimator.

For completeness, a direct implementation has construction cost
\begingroup
\[
O\!\left(d^2n+n\log n+dn|\calH_l|+d^3|\calH_l|{}+j|\calH_l|\right).
\]
\endgroup
In the noisy regime, $|\calH_l|\leq l^*\leq l_{\max}$ and $l_{\max}$ is a fixed model-class parameter, so the assignment term is absorbed into the model-dependent constant. Moreover, $n_{loc}\geq2d+1$ and $|\calH_l|n_{loc}\leq n/4$ imply $d^3|\calH_l|=O(d^2n)$. Tabulating the nearest occupied bins costs $O(j^*|\calH_{l^*}|)$, with $j^*|\calH_{l^*}|\lesssim M^*n^{1/(2s+1)}$. For fixed $l_{\max}$ and $M^*$ this yields the stated $O(d^2n\log n)$ construction bound.
For a query, the empirical distances cost $O(d|\calH_l|)$ operations. The fit in the nearest occupied bin can be read in constant time after tabulating the nearest occupied index on each side of each bin.
\section{Parameters for the committor experiment}\label{Appendix: committor_details}
The experiment in Section \ref{s:ex_reaction_coordinates} uses ambient dimension $d=40$ and a unit-length circular arc of radius $1/6$, with angular parameter ranging from $0$ to $6$. The tangential potential is
\begingroup
\[
\begin{aligned}
U_{tan}(s)={}&512s^{12}-3072s^{11}+8448s^{10}-14080s^9+15840s^8-12672s^7\\
&+7392s^6-3168s^5+990.625s^4-221.25s^3+33.625s^2-2.9813s+1.
\end{aligned}
\]
\endgroup
with minima at $s_A=0.126$ and $s_B=0.863$. The basin tolerances are $\delta_{tan}=0.02$ and $\delta_{nor}=0.05$, and the normal confinement is $U_{nor}(r)=5r^2$. The initial distribution has normal standard deviation $0.1$ in the first two coordinates, $0.12$ in coordinates $3,\dots,13$, $0.15$ in coordinates $14,\dots,26$, and $0.08$ in coordinates $27,\dots,40$. We use $\sigma_{BM}=0.2$, Euler--Maruyama time step $\Delta t=2\times10^{-3}$, $n=4\times10^4$ initial configurations, and $K=4\times10^4$ trajectories per configuration. The public repository contains the complete simulation and reproduction workflow.

\begin{figure}[tb]\centering
\includegraphics[width=0.9\textwidth]{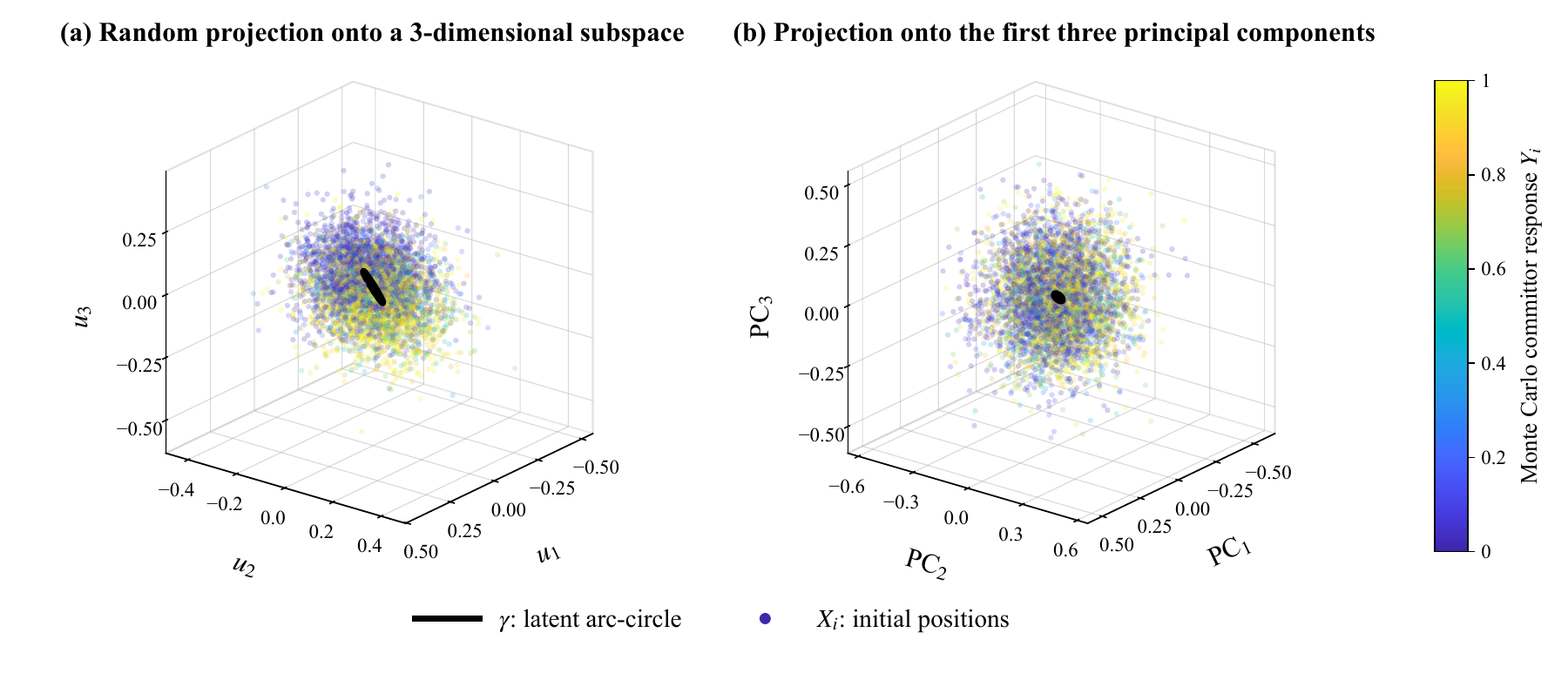}
\caption{Low-dimensional projections of the predictors $X_i$ in the committor experiment, colored by the Monte Carlo responses $Y_i$ defined in \eqref{Equation: reaction_coordinate_Y}. Left: a random projection onto a three-dimensional subspace. Right: projection onto the first three principal components. The latent circular reaction path is shown in black. In this example, the latent geometric structure is not readily revealed by either standard projection.}
\label{Figure: CRC_view}
\end{figure}

\section{Case Analysis: Meyer helix}\label{Appendix: Meyer-Helix}
\subsubsection{Background: standard \& modified Meyer staircase}
We begin with the standard Meyer staircase. Fix constant $\delta\geq 1$. Consider the unit interval $I = [0,1]$ and the set of Gaussians $\calN(t; \mu,\delta^2)$ where the mean $\mu$ takes values in $I$, and the density function is truncated to accept arguments $t\in I$ only. Varying $\mu \in I$ in this manner induces a smooth embedding of the interval $I$ into the infinite-dimensional Hilbert space $L^2(I)$, i.e., a curve. Explicitly, we take the square root of the Gaussian density centered at $\mu \in I$ and truncate it to $t\in I$:
\begin{align} \label{Eqn: Meyer1} \begin{split}
I   \to L^2(I) \quad:\quad
\mu \mapsto g_\mu(t) := \frac{1}{\sqrt[4]{2\pi}\sqrt{\delta}}\exp\rbracket{-\frac{|t-\mu|^2}{4\delta^2}}
\end{split} \ \, .\end{align}
 By discretizing $I$, we may sample this manifold and project it into a finite-dimensional space. In particular, for any $d\in\bbN$, a grid $\Gamma_d \subseteq I$ of $d$ points may be generated. It is obtained by subdividing $I$ in $d$ equal parts and thus $\Gamma_d(k) = k / d$ for $k=1,\dots,d$. Let $C(I)$ denote the space of continuous functions on $I$. We evaluate the Gaussian functions on the grid
 \begin{align}\label{Eqn: Meyer2}\begin{split}
C(I) \to \bbR^d \quad:\quad
g_\mu(t) \mapsto (g_\mu(1/d),\dots, g_\mu(1) )^\intercal
\end{split} \ \, . \end{align}
 Thus, combining the preceding two maps \eqref{Eqn: Meyer1} and \eqref{Eqn: Meyer2} produces an embedding of interval $I$ into $\bbR^d$ (which is equivalent to a curve in $\bbR^d$). We write it explicitly as $x(t) = (x_1(t),\dots,x_d(t))^\intercal$ where $t\in I$ and for each $k=1,\dots,d$. For the standard Meyer staircase, the expression for $x_k(t)$ is
 \begin{equation}\label{Eqn: Meyer}
 x_k(t) = \frac{1}{\sqrt[4]{2\pi}\sqrt{\delta}} \exp\rbracket{-\frac{|k/d - t|^2}{4\delta^2}} \ \, .\end{equation}
 This expression differs from Definition \ref{def:NSVM} because it does not use the unit-speed parameterization. It has two advantages: it is defined uniformly across dimensions, avoiding dimension-dependent parameter intervals, and it emphasizes that these curves are finite-dimensional approximations of $g_\mu(t)\in L^2(I)$. For the numerical experiments, we reparameterize the curve by arclength.

Notice that the map \eqref{Eqn: Meyer1} describes a curve in $L^2(I)$. Here $\mu\in I$ parameterizes this curve while $t\in I$ is merely the argument for the function $g_\mu$. On the other hand, Equation \eqref{Eqn: Meyer} describes a curve in $\bbR^d$. Here $t\in I$ parameterizes the curve while $\mu$ is replaced by a discrete grid $\Gamma_d$.

The standard Meyer staircase provides a curve in high-dimensional Euclidean space $\bbR^d$.
For fixed $\delta$, the piecewise constant interpolant of equal to $x_k(t)$ on $((k-1)/d,k/d]$ converges in $L^2(I)$ to $g_t$ by continuity. Its squared $L^2$ norm is $d^{-1}\norm{x(t)}^2$, so the corresponding Euclidean normalization is $d^{-1/2}x(t)$. Because we are focusing on the regression problem for general curve classes, we need to consider various curves with different complexity and different ambient dimensions. Our strategy is to consider analogies of \eqref{Eqn: Meyer}.

A direct modification of the standard Meyer staircase is to let $\delta=\frac1d$ in equation \ref{Eqn: Meyer}. This modified Meyer staircase is an example of a collection of curves whose complexity grows with dimension $d$. Besides parameters such as length, diameter, curvature, and reach, we also consider the effective linear dimension. One way to measure the effective linear dimension is to study the singular values of the curve. Suppose we perform singular value decomposition on a curve in $\bbR^d$ and obtain its singular value $\lambda_\gamma(k), k = 1,\dots,d$ in descending order. Then, we can consider the sum of the singular values divided by the largest singular value,
$ \normS{\lambda_\gamma}{1} := \sum_{k=1}^d \lambda_\gamma(k) / \lambda_\gamma(1)$,
or count the number of singular values that are greater than $0.05$ times the largest singular value,
$ \normS{\lambda_\gamma}{0} := \#\{ \lambda_\gamma(k): \lambda_\gamma(k)>0.05 \lambda_\gamma(1) \} $.
Both quantities are invariant under rescaling and summarize the spread of the singular values. For a relative mean squared reconstruction error at most $\alpha$, the minimum PCA dimension is the smallest $r$ for which $\sum_{k>r}\lambda_\gamma(k)^2\leq\alpha\sum_{k=1}^d\lambda_\gamma(k)^2$.

\begin{figure}[tb]\centering
\includegraphics[width=0.8\textwidth]{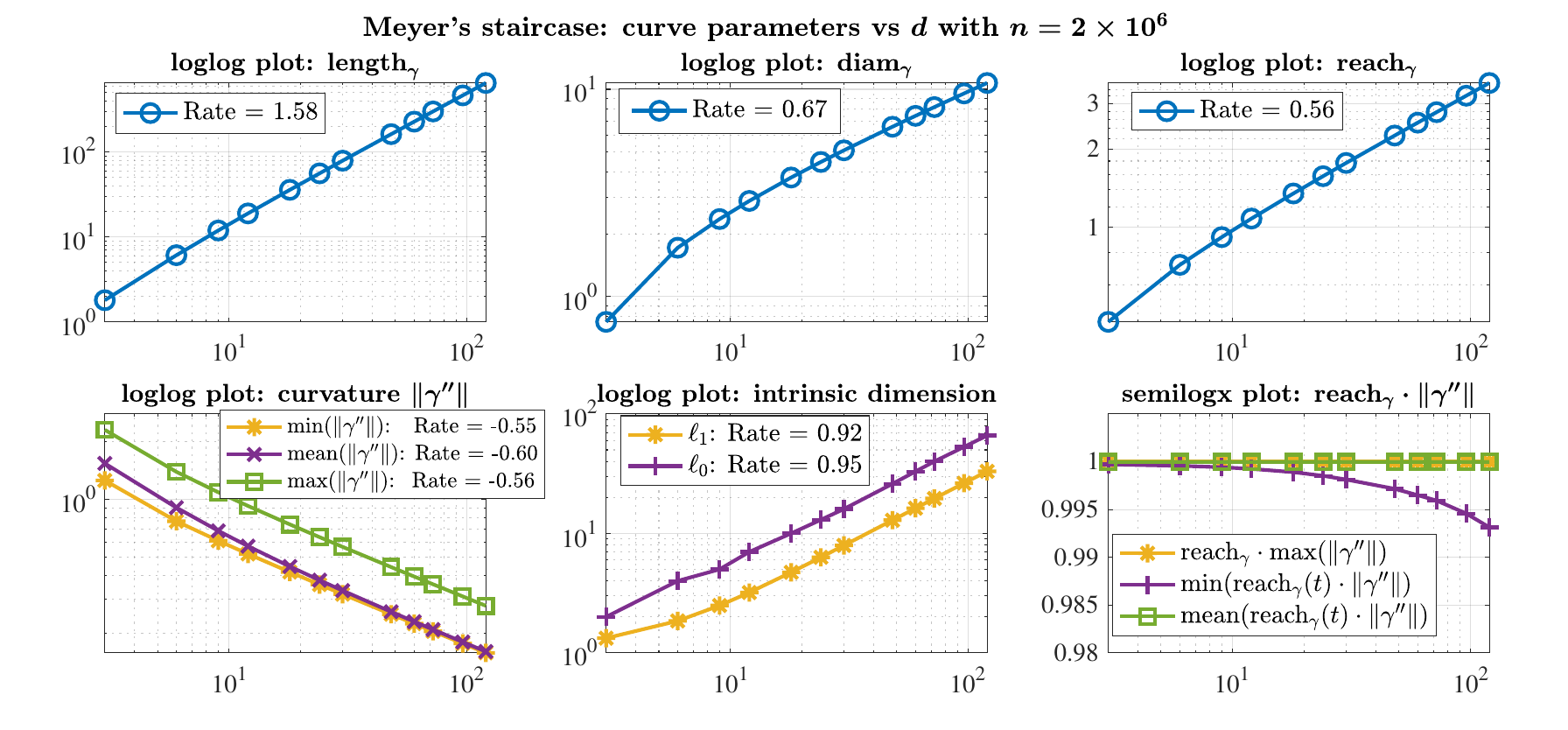}
\caption{Behavior of geometric features of the modified Meyer staircase in $\mathbb{R}^d$ as a function of $d$.}
\label{Fig_Meyer}
\end{figure}

Figure \ref{Fig_Meyer} illustrates the relationship between the modified Meyer staircase parameters and the dimension $d$.
We observe that the length is roughly proportional to $d^{1.5}$, the diameter to $d^{0.5}$, the curvature to $d^{-0.5}$, and the reach to $d^{0.5}$. Both $\normS{\lambda_\gamma}{1}$ and $\normS{\lambda_\gamma}{0}$ are roughly proportional to $d^1$. Thus, the complexity of the modified Meyer staircase grows with $d$.

However, the modified Meyer staircase is still special in the following two respects:
(i) It remains approximately on the sphere $\sqrt{d} \bbS^{d-1}$: For $t\in(0,1)$,
$$ \frac{1}{d} \norm{x(t)}^2 := \frac{1}{d}\sum_{k=1}^d x_k(t)^2 \approx \int_0^1 \frac{1}{\sqrt{2\pi} \delta} \exp\rbracket{-\frac{|s-t|^2}{2\delta^2}} ds \asymp 1 \ \ \text{ when } d \text{ is large }\,;$$
(ii) the local reach of the modified Meyer staircase is almost the reciprocal of the magnitude of local curvature.
 These two properties indicate that the curve traverses the space with weak self-entanglement. This suggests the possibility of finding a linear projection $P:\bbR^d\to\bbR^{d'}$ with $d'$ much smaller than $d$ such that the projected image $P\gamma$ is a much simpler curve. For example, suppose we project the modified Meyer staircase linearly onto its first few principal components: the projected image is a simple curve, and the learning problem can be significantly simplified if we study the regression problem on the projected curve. Hence, to test the performance of the regression algorithm, we aim to test curves that are so complex that there is no trivial dimension reduction, e.g., via standard techniques such as principal component analysis.

\subsubsection{Meyer helix}
For the numerical experiments, we seek sufficiently complex curves and use the following indicators of complexity:
(i) geometric quantities---length, diameter, reach, and the effective linear dimensions $\normS{\lambda_\gamma}{1}$ and $\normS{\lambda_\gamma}{0}$---that grow with $d$;
(ii) the curve $\gamma$ admits no trivial dimension reduction. In particular, consider a linear projection $P_{d'}:\bbR^d\to\bbR^{d'}$ with $2\leq d'\leq d$, such as projection onto the first $d'\leq d$ principal components. Define the ``regression complexity''
\begin{equation}\label{Eqn: regression complexity}
C_\gamma:=\frac{\len_\gamma}{\reach_\gamma}
\end{equation}
 of a curve as its length divided by its reach. We seek curves sufficiently complex that, whenever the projected curve $P_{d'}\gamma$ has \emph{regression complexity} $C_{P_{d'}\gamma}\lesssim C_\gamma$, the target dimension cannot be small; for example, $d'\gtrsim d$.

Because the regression problem is scale invariant, we can freely rescale the curve, and we choose the normalization such that the reach equals $\sqrt{d}$. This is consistent with Assumption \ref{gamma1} and allows us to take $\sigma_\gamma$, the deviation of data $X$ away from the curve, of order one. In particular, we do not let curvature grow with $d$ here.

For each coordinate $k=1,\ldots,d$ set $a=10$, $\delta_d=(1+0.3\cos(ak))/d$, and $\delta_d'=(1+0.3\sin(ak))/d$. With the envelope $G(z)=\exp(-z^2/(1+z))$ for $z\geq0$, we define the Meyer helix on $0\leq t\leq1$ by
\begin{equation}\label{Eqn: Meyer-Helix-variant}
 x_k(t) = \frac{1}{\sqrt[4]{2\pi} \sqrt{\delta_d}} \cos\rbracket{a k+\frac{t-k/d}{\delta_d '}} G\rbracket{ \frac{\abs{k/d-t}}{\delta_d} } \ \, ,\end{equation}
Compared with the standard and modified Meyer staircases, this curve contains an additional cosine factor that makes $x(t)$ traverse $\bbR^d$ more broadly and introduces greater self-entanglement. Moreover, the following quantities vary across coordinates: the frequency $1/\delta_d$ in function $G$, the frequency $1/\delta_d'$ in the cosine term, and the phase $ak$ in the cosine term. These variations make the curve less structured while preserving the desired complexity.

\begin{figure}[tb]\centering
\includegraphics[width=0.8\textwidth]{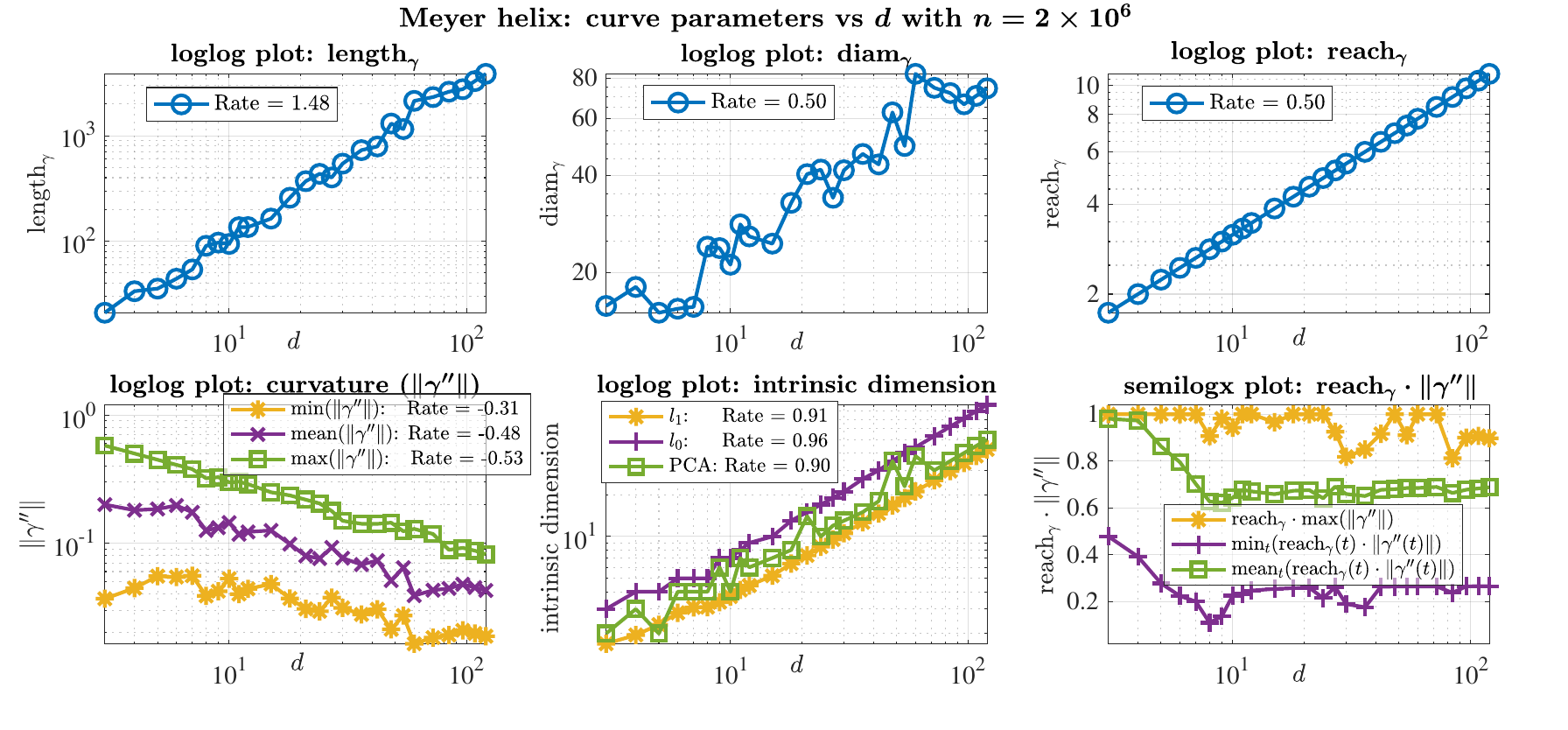}
\caption{Behavior of geometric features of the Meyer helix in $\mathbb{R}^d$ as a function of $d$.}
\label{Fig_Meyer-Helix}
\end{figure}

In Figure \ref{Fig_Meyer-Helix}, we plot the geometric parameters of the Meyer helix after scaling its reach to $\sqrt{d}$. As for the modified Meyer staircase, the length is roughly proportional to $d^{1.5}$, the diameter is roughly proportional to $d^{0.5}$, $\norm{\gamma''}_\infty$ is roughly proportional to $d^{-1/2}$, the reach is roughly proportional to $d^{1/2}$, and the effective linear dimensions $\normS{\lambda_\gamma}{1}$ and $\normS{\lambda_\gamma}{0}$ are roughly proportional to $d$. For projection onto the first $d'$ principal components, we define $d_{SVD}$ as the smallest $d'\in\{2,\ldots,d\}$ with $C_{P_{d'}\gamma}\leq 1.2 C_\gamma$. A projection onto a line has an interval as its image, hence infinite reach and zero length-to-reach ratio. In the dimensions tested, $d_{SVD}$ grows approximately linearly with $d$.

These complexity properties are robust: we can also choose Gaussian-type decay $G(z)=\exp(-z^2)$ or choose different constants in the construction (e.g., $a,\delta_d$, and $\delta_d'$) and these properties persist. This indicates that the complexity of this collection of Meyer helices grows with the ambient dimension $d$.
Numerical tests suggest that this collection of curves does not admit a simple random-projection-based dimension reduction that preserves geometric properties such as length and reach. Recall the following form of the Johnson--Lindenstrauss random-projection lemma for manifolds:
\begin{lemma}[{\cite[proof of Theorem~3.1]{RM09JL}}]\label{lem:manifold-projection} Let $\calM$ be a compact smooth $K$-dimensional submanifold without boundary in $\bbR^N$, with volume $V$ and reach $\tau>0$. Let $R>0$ be a geodesic covering regularity constant for $\calM$, meaning that for every $0<u\leq\tau$ the manifold can be covered by at most $V(R\sqrt K/u)^K$ geodesic balls of radius $u$. Let $\Phi$ be orthogonal projection onto a uniformly random $M$-dimensional subspace, expressed in orthonormal coordinates. For relative distortion $0<\epsilon<1$, failure probability $0<\rho<1$ and absolute constants $C,C_0$, a sufficient condition is
\[C\epsilon^{-2}\log(e/\rho)\log\!\left(e+\frac{V}{\tau^K}
       \left(\frac{C_0 R\sqrt K\,N^{3/2}}{\epsilon^3}\right)^K\right)\leq M\leq N\,.\]
Then, with probability at least $1-\rho$, for all distinct $x,y\in\calM$,
$$ (1-\epsilon)\sqrt{\frac{M}{N}} \leq \frac{\norm{ \Phi x - \Phi y}_2}{\norm{ x - y}_2} \leq (1+\epsilon)\sqrt{\frac{M}{N}} \ \, . $$
\end{lemma}

For the embedded arcs considered here the endpoints can be included directly in the covering argument. Let $\Phi$ be the random projection of Lemma~\ref{lem:manifold-projection} with ambient dimension $N=d$, target dimension $M$, distortion $\epsilon$ and failure probability $\rho$. Write $L=\len_\gamma$ and fix $0<r\leq\reach_\gamma$ with $r$ finite. For $\eta=\epsilon/(32\sqrt d)$ put $b=\min(L,r)$ and take an arclength grid, including both endpoints, of mesh at most $b\eta^2/100$ and size $O(1+L/(b\eta^2))$. The grid tangents and the normalized chords between distinct grid points, with both signs, form a $2\eta$-net of the unit secant directions. Indeed, for an arclength separation $D\leq b\eta$, averaging the tangent along the arc and using $\norm{\gamma''}_\infty\leq1/r$ bounds the distance from the unit secant to the initial tangent by $D/r$. Moving to the nearest grid tangent costs at most $\eta^2/100$. For $D>b\eta$, \eqref{eq:reach} gives a chord length of at least $2b\eta/\pi$. Rounding both endpoints to the grid and normalizing therefore changes its direction by at most $2\eta$.
Apply the finite-set Johnson--Lindenstrauss lemma to these directions and the origin with distortion $\epsilon/2$. Since $\norm{\Phi}\leq1$, passage from a grid direction to a unit secant adds at most $2\eta\sqrt{d/M}\leq\epsilon/2$ to the rescaled norm error. Thus the same distortion conclusion holds for the arc whenever
\[C\epsilon^{-2}\log(e/\rho)\log\!\left(\frac{ed}{\epsilon}\left(1+\frac{L}{r}\right)\right)\leq M\leq d\,.\]
For finite reach one may take $r=\reach_\gamma$.
In our setting, however, the more relevant question is whether the projection reduces geometric complexity, for example $\len_\gamma / \reach_\gamma$; furthermore, our samples are not distributed on the curve, but in a tube around the curve of radius as large as a fraction of the reach.
We perform numerical tests in the setting of Figure \ref{Figure: 36 dimensional Meyer-Helix-L1-C1-variant1}. For the Meyer helix in dimension $d=36$, we project onto a 12-dimensional subspace using PCA or a random projection and compute the length and reach of the image. We consider ten independent repetitions and, for comparison, we also include the original curve as well as the projection onto the first 12 principal components. To be consistent with the scaling in the Johnson-Lindenstrauss lemma, for the PCA and random projection, we rescale points on the projected image of the curve by the factor $\sqrt{36/12}=\sqrt{3}$.

 \begin{table}[H]
 \small{
\begin{center}\begin{tabular}{ ||c|| c |c | c c c c c c c c c c|| }
 \hline
 projection $P$ & original $\gamma$ & PCA & 1 & 2 & 3 & 4 & 5 & 6 & 7 & 8 & 9 & 10 \\
 \hline
 $ \len_{P\gamma}$ & 731 & 1136 & 703 & 720 & 709 & 717 & 639 & 748 & 766 & 709 & 737 & 706 \\
 \hline
 $ \reach_{P\gamma}$ & 6.0 & 5.0 & 2.2 & 4.4 & 2.4 & 3.9 & 2.9 & 4.4 & 2.0 & 4.3 & 2.5 & 2.3 \\
 \hline
 $\frac{\len_{P\gamma}}{ \reach_{P\gamma}}$ & 122 & 226 & 322 & 164 & 295 & 187 & 223 & 171 & 383 & 166 & 291 & 313\\
 \hline
\end{tabular}\end{center}}\end{table}
For the tested projections, the measured length-to-reach ratios exceed that of the original helix. The distance guarantee above controls secant lengths, whereas this comparison also measures the reach of the projected image.

\vskip 0.2in
\bibliography{RefDatabase}

@book{freidlin1998random,
  author    = {Freidlin, Mark I. and Wentzell, Alexander D.},
  title     = {Random Perturbations of Dynamical Systems},
  series    = {Grundlehren der mathematischen Wissenschaften},
  volume    = {260},
  edition   = {Second},
  publisher = {Springer},
  address   = {New York},
  year      = {1998},
  doi       = {10.1007/978-1-4612-0611-8},
  url       = {https://doi.org/10.1007/978-1-4612-0611-8},
  note      = {Translated from the 1979 Russian original by Joseph Sz{\"u}cs}
}

@article{e2006towards,
  title={Towards a theory of transition paths},
  author={E, Weinan and Vanden-Eijnden, Eric},
  journal={Journal of Statistical Physics},
  volume={123},
  pages={503--523},
  year={2006},
  publisher={Springer}
}

@article{metzner2006illustration,
  title={Illustration of transition path theory on a collection of simple examples},
  author={Metzner, Philipp and Sch{\"u}tte, Christof and Vanden-Eijnden, Eric},
  journal={The Journal of Chemical Physics},
  volume={125},
  number={8},
  pages={084110},
  year={2006},
  publisher={AIP Publishing}
}

@article{metzner2009transition,
  title={Transition path theory for Markov jump processes},
  author={Metzner, Philipp and Sch{\"u}tte, Christof and Vanden-Eijnden, Eric},
  journal={Multiscale Modeling \& Simulation},
  volume={7},
  number={3},
  pages={1192--1219},
  year={2009},
  publisher={SIAM}
}

@article{Xia_TwoEstimatorsSIM,
 ISSN = {02664666, 14694360},
 URL = {http://www.jstor.org/stable/4093215},
 author = {Yingcun Xia},
 journal = {Econometric Theory},
 number = {6},
 pages = {1112--1137},
 publisher = {Cambridge University Press},
 title = {Asymptotic Distributions for Two Estimators of the Single-Index Model},
 urldate = {2025-07-10},
 volume = {22},
 year = {2006}
}

@inproceedings{pmlr-v247-damian24a,
  title = 	 {Computational-Statistical Gaps in Gaussian Single-Index Models (Extended Abstract)},
  author =       {Damian, Alex and Pillaud-Vivien, Loucas and Lee, Jason and Bruna, Joan},
  booktitle = 	 {Proceedings of Thirty Seventh Conference on Learning Theory},
  pages = 	 {1262--1262},
  year = 	 {2024},
  editor = 	 {Agrawal, Shipra and Roth, Aaron},
  volume = 	 {247},
  series = 	 {Proceedings of Machine Learning Research},
  month = 	 {30 Jun--03 Jul},
  publisher =    {PMLR},
  url = 	 {https://proceedings.mlr.press/v247/damian24a.html},
}

@inproceedings{wang2024learning,
  title     = {Learning Hierarchical Polynomials with Three-Layer Neural Networks},
  author    = {Wang, Zihao and Nichani, Eshaan and Lee, Jason D.},
  booktitle = {International Conference on Learning Representations (ICLR)},
  year      = {2024},
  url       = {https://openreview.net/forum?id=QgwAYFrh9t}
}

@article{BGGIST07,
  title={Average decay estimates for Fourier transforms of measures supported on curves},
  author={Brandolini, Luca and Gigante, Giacomo and Greenleaf, Allan and Iosevich, Alexander and Seeger, Andreas and Travaglini, Giancarlo},
  journal={The Journal of Geometric Analysis},
  volume={17},
  number={1},
  pages={15--40},
  year={2007},
  doi={10.1007/BF02922080}
}

@misc{Lai2021TheKS,
  title         = {The Kolmogorov Superposition Theorem can Break the Curse of Dimensionality When Approximating High Dimensional Functions},
  author        = {Lai, Ming{-}Jun and Shen, Zhaiming},
  year          = {2021},
  eprint        = {2112.09963},
  archivePrefix = {arXiv},
  primaryClass  = {math.NA},
  doi           = {10.48550/arXiv.2112.09963},
  url           = {https://arxiv.org/abs/2112.09963}
}

@inproceedings{liu2024kankolmogorovarnoldnetworks,
  title     = {{KAN}: Kolmogorov--Arnold Networks},
  author    = {Liu, Ziming and Wang, Yixuan and Vaidya, Sachin and Ruehle, Fabian and Halverson, James and Solja{\v{c}}i{\'c}, Marin and Hou, Thomas Y. and Tegmark, Max},
  booktitle = {International Conference on Learning Representations (ICLR)},
  year      = {2025},
  url       = {https://openreview.net/forum?id=Ozo7qJ5vZi}
}

@article{bietti2022learning,
  title={Learning single-index models with shallow neural networks},
  author={Bietti, Alberto and Bruna, Joan and Sanford, Clayton and Song, Min Jae},
  journal={Advances in Neural Information Processing Systems},
  volume={35},
  pages={9768--9783},
  year={2022}
}

@article{doi:10.1126/science.adi5639,
author = {Adityanarayanan Radhakrishnan  and Daniel Beaglehole  and Parthe Pandit  and Mikhail Belkin },
title = {Mechanism for feature learning in neural networks and backpropagation-free machine learning models},
journal = {Science},
volume = {383},
number = {6690},
pages = {1461-1467},
year = {2024},
doi = {10.1126/science.adi5639},
URL = {https://www.science.org/doi/abs/10.1126/science.adi5639},
eprint = {https://www.science.org/doi/pdf/10.1126/science.adi5639},
}

@inproceedings{lee2024neuralnetworklearnslowdimensional,
 author = {Lee, Jason D. and Oko, Kazusato and Suzuki, Taiji and Wu, Denny},
 booktitle = {Advances in Neural Information Processing Systems},
 editor = {A. Globerson and L. Mackey and D. Belgrave and A. Fan and U. Paquet and J. Tomczak and C. Zhang},
 pages = {58716--58756},
 publisher = {Curran Associates, Inc.},
 title = {Neural Network Learns Low-Dimensional Polynomials with {SGD} near the Information-Theoretic Limit},
 url = {https://proceedings.neurips.cc/paper_files/paper/2024/file/6bd5fca2074dcd9ede9de50f71f7ec28-Paper-Conference.pdf},
 volume = {37},
 year = {2024}
}

@article{liu2024deepneuralnetworksadaptive,
  title   = {Deep Neural Networks are Adaptive to Function Regularity and Data Distribution in Approximation and Estimation},
  author  = {Liu, Hao and Cheng, Jiahui and Liao, Wenjing},
  journal = {Journal of Machine Learning Research},
  volume  = {26},
  number  = {213},
  pages   = {1--56},
  year    = {2025},
  url     = {https://www.jmlr.org/papers/v26/24-1148.html}
}

@article{liu2024learningcentralsubspace,
  title   = {Learning Functions Varying along a Central Subspace},
  author  = {Liu, Hao and Liao, Wenjing},
  journal = {SIAM Journal on Mathematics of Data Science},
  volume  = {6},
  number  = {2},
  pages   = {343--371},
  year    = {2024},
  doi     = {10.1137/23M1557751},
  url     = {https://doi.org/10.1137/23M1557751},
  eprint  = {2001.07883},
  archivePrefix = {arXiv},
  primaryClass = {stat.ML}
}

@article{PoggioRosascoHyperkernels,
  title   = {Learning Multi-Index Models with Hyper-Kernel Ridge Regression},
  author  = {Huang, Shuo and Labarri{\`e}re, Hippolyte and De Vito, Ernesto and Poggio, Tomaso and Rosasco, Lorenzo},
  journal = {arXiv preprint arXiv:2510.02532},
  year    = {2025},
  url     = {https://arxiv.org/abs/2510.02532}
}

@article{liao2024taskmanifoldneuralnet,
  title   = {Transformers for Learning on Noisy and Task-Level Manifolds: Approximation and Generalization Insights},
  author  = {Shen, Zhaiming and Havrilla, Alex and Lai, Rongjie and Cloninger, Alexander and Liao, Wenjing},
  journal = {arXiv preprint arXiv:2505.03205},
  year    = {2025},
  url     = {https://arxiv.org/abs/2505.03205}
}

@article{bietti2023learninggaussianmultiindexmodels,
  title        = {On Learning Gaussian Multi-Index Models with Gradient Flow Part I: General Properties and Two-Timescale Learning},
  author       = {Bietti, Alberto and Bruna, Joan and Pillaud-Vivien, Loucas},
  journal      = {Communications on Pure and Applied Mathematics},
  volume       = {78},
  number       = {12},
  pages        = {2354--2435},
  year         = {2025},
  doi          = {10.1002/cpa.70006}
}

@article{JMLR:v22:20-1288,
  author  = {Gerard Ben Arous and Reza Gheissari and Aukosh Jagannath},
  title   = {Online stochastic gradient descent on non-convex losses from high-dimensional inference},
  journal = {Journal of Machine Learning Research},
  year    = {2021},
  volume  = {22},
  number  = {106},
  pages   = {1--51},
  url     = {https://www.jmlr.org/papers/v22/20-1288.html}
}

@article{RM09JL,
  author  = {Baraniuk, Richard G. and Wakin, Michael B.},
  title   = {Random Projections of Smooth Manifolds},
  journal = {Foundations of Computational Mathematics},
  volume  = {9},
  number  = {1},
  pages   = {51--77},
  year    = {2009},
  doi     = {10.1007/s10208-007-9011-z},
  url     = {https://doi.org/10.1007/s10208-007-9011-z}
}

@article{Shamir_discussion,
title={Discussion of: ``Nonparametric regression using deep neural networks with {R}e{L}{U} activation function''},
author={Ohad Shamir},
journal = {The Annals of Statistics},
year=2020, 
volume=48,
number=4,
pages={1911--1915},
url={https://doi.org/10.1214/19-AOS1915}
}

@article{10.1214/19-AOS1875,
author = {Johannes Schmidt-Hieber},
title = {{Nonparametric Regression Using Deep Neural Networks with {ReLU} Activation Function}},
volume = {48},
journal = {The Annals of Statistics},
number = {4},
publisher = {Institute of Mathematical Statistics},
pages = {1875 -- 1897},
year = {2020},
doi = {10.1214/19-AOS1875},
URL = {https://doi.org/10.1214/19-AOS1875}
}

@article{Horowitz_generalclassnonparametric,
title={Rate-Optimal Estimation for a General Class of Nonparametric Regression Models with Unknown Link Functions},
author={Joel L. Horowitz and Enno Mammen},
journal={The Annals of Statistics},
volume={35},
number={6},
year={2007},
pages={2589--2619}
}

@article{doi:10.1080/01621459.1985.10478157,
author = {Leo Breiman and Jerome H. Friedman},
title = {Estimating Optimal Transformations for Multiple Regression and Correlation},
journal = {Journal of the American Statistical Association},
volume = {80},
number = {391},
pages = {580--598},
year = {1985},
publisher = {ASA Website},
doi = {10.1080/01621459.1985.10478157},
URL = {https://www.tandfonline.com/doi/abs/10.1080/01621459.1985.10478157}
}

@article{shen2021deepquantileregressionmitigating,
  title   = {Deep Quantile Regression: Mitigating the Curse of Dimensionality Through Composition},
  author  = {Shen, Guohao and Jiao, Yuling and Lin, Yuanyuan and Horowitz, Joel L. and Huang, Jian},
  journal = {arXiv preprint arXiv:2107.04907},
  year    = {2021},
  eprint  = {2107.04907},
  archivePrefix = {arXiv},
  primaryClass  = {stat.ML},
  url     = {https://arxiv.org/abs/2107.04907}
}

@inproceedings{suzuki2021deep,
  title={Deep learning is adaptive to intrinsic dimensionality of model smoothness in anisotropic {B}esov space},
  author={Taiji Suzuki and Atsushi Nitanda},
  booktitle={Advances in Neural Information Processing Systems},
  volume={34},
  pages={3609--3621},
  year={2021},
  url={https://proceedings.neurips.cc/paper_files/paper/2021/hash/1dacb10f0623c67cb7dbb37587d8b38a-Abstract.html}
}

@article{federer1959curvature,
	Author = {Federer, H.},
	Journal = {Transactions of the American Mathematical Society},
	Number = {3},
	Pages = {418--491},
	Publisher = {JSTOR},
	Title = {Curvature measures},
	Volume = {93},
	Year = {1959}}

@inproceedings{stromberg,
title={Computation with wavelets in higher dimensions},
author= {Jan-Olov Str\"omberg},
year = {1998},
booktitle={Proceedings of the International Congress of Mathematicians},
volume	= {3},
pages	= {523--532}
}

@book{Hastie:GeneralizedAdditiveModels,
author = {Hastie, Trevor J. and Tibshirani, Robert J.},
title = {Generalized Additive Models},
publisher = {Chapman and Hall},
address = {London},
year = {1990},
isbn = {9780412343902}
}

@article{DeVore:UniversalAlgorithmsLearningTheoryII,
	Author = {P. Binev and A. Cohen and W. Dahmen and R.A. DeVore and V. Temlyakov},
	Journal = {Constr. Approx.},
	Number = {2},
	Pages = {127--152},
	Title = {Universal Algorithms for Learning Theory Part II: piecewise polynomial functions},
	Volume = {26},
	Year = {2007}}

@article{DeVore:UniversalAlgorithmsLearningTheoryI,
	Author = {P. Binev and A. Cohen and W. Dahmen and R.A. DeVore and V. Temlyakov},
	Journal = {J. Mach. Learn. Res.},
	Pages = {1297--1321},
	Title = {Universal Algorithms for Learning Theory Part I: piecewise constant functions},
	Volume = {6},
	Year = {2005}}

@article{LMV22B,
author = {Alessandro Lanteri and Mauro Maggioni and Stefano Vigogna},
title = {{Conditional regression for single-index models}},
volume = {28},
journal = {Bernoulli},
number = {4},
publisher = {Bernoulli Society for Mathematical Statistics and Probability},
pages = {3051--3078},
year = {2022},
doi = {10.3150/22-BEJ1482},
URL = {https://doi.org/10.3150/22-BEJ1482}
}

@book{GKKW02,
  author    = {Gy{\"o}rfi, L{\'a}szl{\'o} and Kohler, Michael and Krzy{\.z}ak, Adam and Walk, Harro},
  title     = {A Distribution-Free Theory of Nonparametric Regression},
  publisher = {Springer},
  address   = {New York},
  year      = {2002},
  doi       = {10.1007/b97848},
  url       = {https://doi.org/10.1007/b97848}
}

@book{Bhatia97,
  author    = {Bhatia, Rajendra},
  title     = {Matrix Analysis},
  series    = {Graduate Texts in Mathematics},
  volume    = {169},
  publisher = {Springer},
  address   = {New York},
  year      = {1997},
  doi       = {10.1007/978-1-4612-0653-8},
  url       = {https://doi.org/10.1007/978-1-4612-0653-8}
}

@article{Barron93,
  author  = {Barron, Andrew R.},
  title   = {Universal Approximation Bounds for Superpositions of a Sigmoidal Function},
  journal = {IEEE Transactions on Information Theory},
  volume  = {39},
  number  = {3},
  pages   = {930--945},
  year    = {1993},
  doi     = {10.1109/18.256500}
}

@article{JLT09,
 ISSN = {00905364, 21688966},
 URL = {http://www.jstor.org/stable/30243671},
 author = {Anatoli B. Juditsky and Oleg V. Lepski and Alexandre B. Tsybakov},
 journal = {The Annals of Statistics},
 number = {3},
 pages = {1360--1404},
 publisher = {Institute of Mathematical Statistics},
 title = {Nonparametric Estimation of Composite Functions},
 volume = {37},
 year = {2009}
}

@article{BL07,
 ISSN = {07492170},
 URL = {http://www.jstor.org/stable/20461468},
 author = {Peter J. Bickel and Bo Li},
 journal = {Lecture Notes-Monograph Series},
 pages = {177--186},
 publisher = {Institute of Mathematical Statistics},
 title = {Local Polynomial Regression on Unknown Manifolds},
 volume = {54},
 year = {2007}
}

@inproceedings{Kpotufe11,
  author    = {Kpotufe, Samory},
  title     = {{k-NN} Regression Adapts to Local Intrinsic Dimension},
  booktitle = {Advances in Neural Information Processing Systems 24},
  pages     = {729--737},
  publisher = {Curran Associates, Inc.},
  year      = {2011},
  isbn      = {9781618395993},
  url       = {https://proceedings.neurips.cc/paper/2011/hash/05f971b5ec196b8c65b75d2ef8267331-Abstract.html}
}

@inproceedings{KG13,
  author    = {Kpotufe, Samory and Garg, Vikas},
  title     = {Adaptivity to Local Smoothness and Dimension in Kernel Regression},
  booktitle = {Advances in Neural Information Processing Systems 26},
  pages     = {3075--3083},
  publisher = {Curran Associates, Inc.},
  year      = {2013},
  url       = {https://proceedings.neurips.cc/paper/2013/hash/28fc2782ea7ef51c1104ccf7b9bea13d-Abstract.html}
}

@inproceedings{LMV16,
  author    = {Liao, Wenjing and Maggioni, Mauro and Vigogna, Stefano},
  title     = {Learning Adaptive Multiscale Approximations to Data and Functions near Low-Dimensional Sets},
  booktitle = {2016 IEEE Information Theory Workshop (ITW)},
  pages     = {226--230},
  publisher = {IEEE},
  address   = {Cambridge, United Kingdom},
  year      = {2016},
  doi       = {10.1109/ITW.2016.7606829},
  url       = {https://doi.org/10.1109/ITW.2016.7606829}
}

@article{LMV22ME,
  author  = {Liao, Wenjing and Maggioni, Mauro and Vigogna, Stefano},
  title   = {Multiscale Regression on Unknown Manifolds},
  journal = {Mathematics in Engineering},
  volume  = {4},
  number  = {4},
  pages   = {1--25},
  year    = {2022},
  issn    = {2640-3501},
  doi     = {10.3934/mine.2022028},
  url     = {https://doi.org/10.3934/mine.2022028}
}

@article{Stone82,
 ISSN = {00905364},
 URL = {http://www.jstor.org/stable/2240707},
 author = {Charles J. Stone},
 journal = {The Annals of Statistics},
 number = {4},
 pages = {1040--1053},
 publisher = {Institute of Mathematical Statistics},
 title = {Optimal Global Rates of Convergence for Nonparametric Regression},
 volume = {10},
 year = {1982}
}

@article{HS89,
  author  = {H{\"a}rdle, Wolfgang and Stoker, Thomas M.},
  title   = {Investigating Smooth Multiple Regression by the Method of Average Derivatives},
  journal = {Journal of the American Statistical Association},
  volume  = {84},
  number  = {408},
  pages   = {986--995},
  year    = {1989},
  issn    = {0162-1459},
  url     = {https://www.jstor.org/stable/2290074}
}

@book{Horowitz98,
  author    = {Horowitz, Joel L.},
  title     = {Semiparametric Methods in Econometrics},
  publisher = {Springer},
  address   = {New York},
  year      = {1998},
  doi       = {10.1007/978-1-4612-0621-7},
  url       = {https://doi.org/10.1007/978-1-4612-0621-7}
}

@article{GL07,
  author  = {Ga{\"i}ffas, St{\'e}phane and Lecu{\'e}, Guillaume},
  title   = {Optimal Rates and Adaptation in the Single-Index Model Using Aggregation},
  journal = {Electronic Journal of Statistics},
  volume  = {1},
  pages   = {538--573},
  year    = {2007},
  doi     = {10.1214/07-EJS077},
  url     = {https://doi.org/10.1214/07-EJS077}
}

@article{Ichimura93,
  author  = {Ichimura, Hidehiko},
  title   = {Semiparametric Least Squares ({SLS}) and Weighted {SLS} Estimation of Single-Index Models},
  journal = {Journal of Econometrics},
  volume  = {58},
  number  = {1--2},
  pages   = {71--120},
  year    = {1993},
  issn    = {0304-4076},
  doi     = {10.1016/0304-4076(93)90114-K},
  url     = {https://doi.org/10.1016/0304-4076(93)90114-K}
}

@article{HHI93,
  author  = {H{\"a}rdle, Wolfgang and Hall, Peter and Ichimura, Hidehiko},
  title   = {Optimal Smoothing in Single-Index Models},
  journal = {The Annals of Statistics},
  volume  = {21},
  number  = {1},
  pages   = {157--178},
  year    = {1993},
  issn    = {0090-5364},
  url     = {https://www.jstor.org/stable/3035585}
}

@article{DHH97,
  author  = {Delecroix, Michel and H{\"a}rdle, Wolfgang and Hristache, Marian},
  title   = {Efficient Estimation in Conditional Single-Index Regression},
  journal = {Journal of Multivariate Analysis},
  volume  = {86},
  number  = {2},
  pages   = {213--226},
  year    = {2003},
  doi     = {10.1016/S0047-259X(02)00046-5},
  url     = {https://doi.org/10.1016/S0047-259X(02)00046-5}
}

@article{DH99,
  title={M-estimateurs semi-param{\'e}triques dans les mod{\`e}les {\`a} direction r{\'e}v{\'e}latrice unique},
  author={Delecroix, Michel and Hristache, Marian},
  journal={Bulletin of the Belgian Mathematical Society Simon Stevin},
  volume={6},
  number={2},
  pages={161--186},
  year={1999},
  publisher={Brussels: Societe Mathematique de Belgique, c1994-}
}

@article{DHP06,
  author  = {Delecroix, Michel and Hristache, Marian and Patilea, Valentin},
  title   = {On Semiparametric M-Estimation in Single-Index Regression},
  journal = {Journal of Statistical Planning and Inference},
  volume  = {136},
  number  = {3},
  pages   = {730--769},
  year    = {2006},
  doi     = {10.1016/j.jspi.2004.09.006},
  url     = {https://doi.org/10.1016/j.jspi.2004.09.006}
}

@article{CFGW97,
  title={Generalized partially linear single-index models},
  author={Carroll, Raymond J and Fan, Jianqing and Gijbels, Ir{\`e}ne and Wand, Matt P},
  journal={Journal of the American Statistical Association},
  volume={92},
  number={438},
  pages={477--489},
  year={1997},
  publisher={Taylor \& Francis}
}

@article{Stoker86,
 ISSN = {00129682, 14680262},
 URL = {http://www.jstor.org/stable/1914309},
 author = {Thomas M. Stoker},
 journal = {Econometrica},
 number = {6},
 pages = {1461--1481},
 publisher = {[Wiley, Econometric Society]},
 title = {Consistent Estimation of Scaled Coefficients},
 volume = {54},
 year = {1986}
}

@article{HJS01,
 ISSN = {00905364},
 URL = {http://www.jstor.org/stable/2673964},
 author = {Marian Hristache and Anatoli Juditsky and Vladimir Spokoiny},
 journal = {The Annals of Statistics},
 number = {3},
 pages = {595--623},
 publisher = {Institute of Mathematical Statistics},
 title = {Direct Estimation of the Index Coefficient in a Single-Index Model},
 urldate = {2023-10-03},
 volume = {29},
 year = {2001}
}

@article{DL91,
 ISSN = {00905364},
 URL = {http://www.jstor.org/stable/2242072},
 author = {Naihua Duan and Ker-Chau Li},
 journal = {The Annals of Statistics},
 number = {2},
 pages = {505--530},
 publisher = {Institute of Mathematical Statistics},
 title = {Slicing Regression: A Link-Free Regression Method},
 volume = {19},
 year = {1991}
}

@article{Li91,
 ISSN = {01621459},
 URL = {http://www.jstor.org/stable/2290563},
 author = {Ker-Chau Li},
 journal = {Journal of the American Statistical Association},
 number = {414},
 pages = {316--327},
 publisher = {[American Statistical Association, Taylor & Francis, Ltd.]},
 title = {Sliced Inverse Regression for Dimension Reduction},
 volume = {86},
 year = {1991}
}

@article{Cook00,
  author    = {Cook, R. Dennis},
  title     = {{SAVE}: A Method for Dimension Reduction and Graphics in Regression},
  journal   = {Communications in Statistics - Theory and Methods},
  volume    = {29},
  number    = {9--10},
  pages     = {2109--2121},
  year      = {2000},
  publisher = {Taylor \& Francis},
  doi       = {10.1080/03610920008832598},
  url       = {https://doi.org/10.1080/03610920008832598}
}

@article{LZC05,
 ISSN = {00905364},
 URL = {http://www.jstor.org/stable/3448618},
 author = {Bing Li and Hongyuan Zha and Francesca Chiaromonte},
 journal = {The Annals of Statistics},
 number = {4},
 pages = {1580--1616},
 publisher = {Institute of Mathematical Statistics},
 title = {Contour Regression: A General Approach to Dimension Reduction},
 volume = {33},
 year = {2005}
}

@article{LW07,
 ISSN = {01621459},
 URL = {http://www.jstor.org/stable/27639941},
 author = {Bing Li and Shaoli Wang},
 journal = {Journal of the American Statistical Association},
 number = {479},
 pages = {997--1008},
 publisher = {[American Statistical Association, Taylor & Francis, Ltd.]},
 title = {On Directional Regression for Dimension Reduction},
 volume = {102},
 year = {2007}
}

@article{CLS13,
  author  = {Coudret, Rapha{\"e}l and Liquet, Benoit and Saracco, J{\'e}r{\^o}me},
  title   = {Comparison of Sliced Inverse Regression Approaches for Underdetermined Cases},
  journal = {Journal de la Soci{\'e}t{\'e} Fran{\c c}aise de Statistique},
  volume  = {155},
  number  = {2},
  pages   = {72--96},
  year    = {2014},
  url     = {https://www.numdam.org/item/JSFS_2014__155_2_72_0/}
}

@book{ACK87,
  title={Trigonometric sums in number theory and analysis},
  author={Arkhipov, Gennady I. and Chubarikov, Vladimir N. and Karatsuba, Anatoly A.},
  series={De Gruyter Expositions in Mathematics},
  volume={39},
  publisher={Walter de Gruyter},
  address={Berlin},
  year={2004},
  note={Translated from the 1987 Russian original},
  isbn={3-11-016266-0}
}

@article{BAUER201793,
  author  = {Bauer, Benedikt and Devroye, Luc and Kohler, Michael and Krzy{\.z}ak, Adam and Walk, Harro},
  title   = {Nonparametric Estimation of a Function from Noiseless Observations at Random Points},
  journal = {Journal of Multivariate Analysis},
  volume  = {160},
  pages   = {93--104},
  year    = {2017},
  issn    = {0047-259X},
  doi     = {10.1016/j.jmva.2017.05.010},
  url     = {https://doi.org/10.1016/j.jmva.2017.05.010}
}

@unpublished{manuscriptNSIM2018,
  author = {Lanteri, Alessandro and Kereta, {\v Z}eljko and Klock, Timo and Maggioni, Mauro and Naumova, Valeriya and Vigogna, Stefano},
  title  = {Thinking Geometrically in Regression with Structured Models: Linear and Nonlinear Single-Index Models},
  note   = {Unpublished manuscript},
  year   = {2018}
}

@article{KeretaKlockNaumova2021,
author = {Kereta, {\v Z}eljko and Klock, Timo and Naumova, Valeriya},
title = {{Nonlinear Generalization of the Monotone Single Index Model}},
journal = {Information and Inference: A Journal of the IMA},
volume = {10},
number = {3},
pages = {987--1029},
year = {2021},
doi = {10.1093/imaiai/iaaa013},
url = {https://doi.org/10.1093/imaiai/iaaa013}
}

@article{alberti2021gradient,
   title={Gradient flow of the population risk in deep learning of single-index models},
   author={Alberti, Giacomo and Ingrosso, Alessandro and Toader, Radu},
   journal={Journal of Statistical Physics},
   volume={182},
   number={3},
   pages={1--33},
   year={2021},
   publisher={Springer}
 }

@article{BLW19,
  author = {Boissonnat, Jean-Daniel and Lieutier, Andr{\'e} and Wintraecken, Mathijs},
  title = {The reach, metric distortion, geodesic convexity and the variation of tangent spaces},
  journal = {Journal of Applied and Computational Topology},
  volume = {3},
  pages = {29--58},
  year = {2019}
}

@article{Tropp12,
  author = {Tropp, Joel A.},
  title = {User-friendly tail bounds for sums of random matrices},
  journal = {Foundations of Computational Mathematics},
  volume = {12},
  pages = {389--434},
  year = {2012}
}

\end{document}